%% file: sample-sigconf.tex
\documentclass[sigconf,nonacm,balance]{acmart}
\usepackage{courier}
\usepackage{colortbl}
\usepackage[nointegrals]{wasysym}
\usepackage{amsthm}
\usepackage{color}
\usepackage{mathtools}
\usepackage{bm}
\usepackage{float}
\usepackage{natbib}
\usepackage{caption}
\usepackage{algorithm}
\usepackage{algorithmic}
\usepackage{tcolorbox}
\usepackage{hhline}
\usepackage{subfiles}
\usepackage{booktabs}
\usepackage{multirow}
\usepackage{stfloats}
\usepackage{lipsum}
\usepackage{arydshln}
\usepackage{graphicx}
\usepackage{subcaption}
\usepackage[normalem]{ulem}

\usepackage{amssymb}
\usepackage{tikz}
\usepackage{amsmath}
\usepackage{enumitem}
\usepackage{xcolor}
\usepackage{makecell}

 \definecolor{agreen}{rgb}{0.55, 0.71, 0.0}
 \definecolor{lightcoral}{rgb}{0.94, 0.5, 0.5}

 \definecolor{mygreen}{rgb}{0.1, 0.6, 0.1}
\definecolor{myblue}{RGB}{0, 112, 192}
\definecolor{myred}{RGB}{255, 0, 0}

\definecolor{mycolor1}{RGB}{248, 233, 225}
\definecolor{mycolor2}{RGB}{192, 223, 231}

\newcommand{\vi}[1]{\textcolor{black}{#1}}
\newcommand{\vid}[1]{\textcolor{black}{#1}}
\newcommand{\vic}[1]{\textcolor{black}{#1}}

\newcommand{\zhiweihu}[1]{\textcolor{black}{#1}}

\newcommand{\zwh}[1]{\textcolor{black}{#1}}

\AtBeginDocument{%
  }

\setcopyright{acmlicensed}
\copyrightyear{2018}
\acmYear{2018}
\acmDOI{XXXXXXX.XXXXXXX}
\acmConference[Conference acronym 'XX]{Make sure to enter the correct
  conference title from your rights confirmation email}{June 03--05,
  2018}{Woodstock, NY}
\acmISBN{978-1-4503-XXXX-X/2018/06}

\begin{document}

%\pagenumbering{gobble}
%\twocolumn
%\section*{Summary of Changes}
%\subfile{sections/summary_of_changes}

%\clearpage
%\pagenumbering{arabic}

\title{\textsc{HyperMono}: A Monotonicity-aware Approach to Hyper-Relational Knowledge Representation}

\author{Zhiwei Hu}
\affiliation{
  \institution{School of Computer and Information Technology \\ Shanxi University}
  \city{Taiyuan}
  \country{China}
}
\email{zhiweihu@whu.edu.cn}

\author{Víctor Gutiérrez-Basulto}
\affiliation{
  \institution{School of Computer Science and Informatics\\ Cardiff University}
  \city{Cardiff}
  \country{UK}
}
\email{gutierrezbasultov@cardiff.ac.uk}

\author{Zhiliang Xiang}
\affiliation{
  \institution{School of Computer Science and Informatics\\ Cardiff University}
  \city{Cardiff}
  \country{UK}
}
\email{xiangz6@cardiff.ac.uk}

\author{Ru Li}
\authornotemark[1]
\affiliation{
  \institution{School of Computer and Information Technology \\ Shanxi University}
  \city{Taiyuan}
  \country{China}
}
\email{liru@sxu.edu.cn}

\author{Jeff Z. Pan}
\authornote{Contact Authors.}
\affiliation{
  \institution{ILCC, School of Informatics\\ University of Edinburgh}
  \city{Edinburgh}
  \country{UK}
}
\email{http://knowledge-representation.org/j.z.pan/}

\subfile{sections/abstract}

%    <concept>
%        <concept_id>10010147.10010178.10010187</concept_id>
%        <concept_desc>Computing methodologies~Knowledge representation and reasoning</concept_desc>
%        <concept_significance>500</concept_significance>
%        </concept>
%    <concept>
%        <concept_id>10002951.10003227.10003351</concept_id>
%        <concept_desc>Information systems~Data mining</concept_desc>
%        <concept_significance>500</concept_significance>
%        </concept>
%  </ccs2012>

% \ccsdesc[500]{Computing methodologies~Knowledge representation and reasoning}
% \ccsdesc[500]{Information systems~Data mining}



\maketitle

\section{Introduction}
\subfile{sections/introduction}
\label{introduction}

\section{Related Work}
\subfile{sections/related_work}

\section{Method}
\subfile{sections/method}

\section{Experiments}
\subfile{sections/experiment}

\section{Conclusions}
\subfile{sections/conclusion}

\clearpage
\bibliographystyle{ACM-Reference-Format}
\bibliography{sample-base}
\balance
\clearpage
\section*{Appendix}
\subfile{sections/appendix}

\end{document}

%% file: sections/abstract.tex
\begin{abstract}
\vic{In a hyper-relational knowledge graph (HKG), each fact is composed of a main triple associated with  attribute-value qualifiers, which  express  additional factual knowledge. The hyper-relational knowledge graph completion (HKGC) task aims at inferring plausible missing links in a HKG. Most existing approaches to HKGC  focus on enhancing the communication between qualifier pairs and main triples, while overlooking two important properties that emerge from the monotonicity of the \vid{hyper-relational graphs} representation regime. \textit{Stage Reasoning} 
allows for a two-step reasoning process, facilitating the integration of  coarse-grained inference results derived solely from main triples  and fine-grained inference results obtained from hyper-relational facts with qualifiers. In the initial stage, coarse-grained results provide an upper bound for correct predictions, which are subsequently  refined in the fine-grained step.}
%
%In a hyper-relational knowledge graph (HKG), each fact is composed of a primal triple associated with several auxiliary attribute-value qualifiers, which can effectively express more factually comprehensive semantics. Hyper-relational knowledge graph completion (HKGC) aims at inferring plausible missing links over HKG. Existing approaches to HKGC mainly focus on enhancing the communication between qualifier pairs and main triple, while overlooking two important properties: 
%\textit{reasoning diversity}, i.e., considering comprehensively of coarse-grained inference results without qualifiers and fine-grained inference results with qualifiers, and 
\vic{More generally, \textit{Qualifier Monotonicity} implies that by attaching more qualifier pairs to a  main triple, may only narrow down the answer set, but never enlarges it. This paper proposes the \textbf{\texttt{HyperMono}} model for hyper-relational knowledge graph completion, which  considers an initial triple-level reasoning phase to be subsequently refined by a   triple+qualifiers  reasoning step to infer missing entities. To realize qualifier monotonicity,   \texttt{HyperMono} represents main triples  as  cones and each qualifier is modeled  as a cone that shrinks  the cone of the main triple to which the qualifier has  been added. Moreover, \texttt{HyperMono} leverages the neighborhood context of an entity to semantically strengthen its  representation. Experiments on three real-world datasets
% (i.e., WD50K, WikiPeople, and JF17K) 
with three different scenario conditions 
% (i.e., mixed-percentage mixed-qualifier, fixed-percentage mixed-qualifier, and fixed-percentage fixed-qualifier) 
demonstrate the strong  performance of \texttt{HyperMono} when compared to the SoTA. %\textcolor{blue}{Datasets and code are available at the following anonymous website: \url{https://anonymous.4open.science/r/HyperMono-7800}}.
%\texttt{HyperMono}'s code is available at: \textcolor{blue}{\url{https://anonymous.4open.science/r/HKGC-HyperMono-AE61}}.
}

%Moreover, RDQM leverages the neighborhood contexts of the seen entity to semantically strengthen the representation of corresponding central entity, further enhancing confidence in missing entity prediction. Experiments on three real-world datasets (i.e., WD50K, WikiPeople, and JF17K) with three different scenario conditions (i.e., mixed-percentage mixed-qualifier, fixed-percentage mixed-qualifier, and fixed-percentage fixed-qualifier) demonstrate the superior performance of RDQM compared to the state-of-the-art.
\end{abstract}

%% file: sections/introduction.tex
\begin{figure}[t!]
    \centering
    \includegraphics[width=0.48\textwidth]{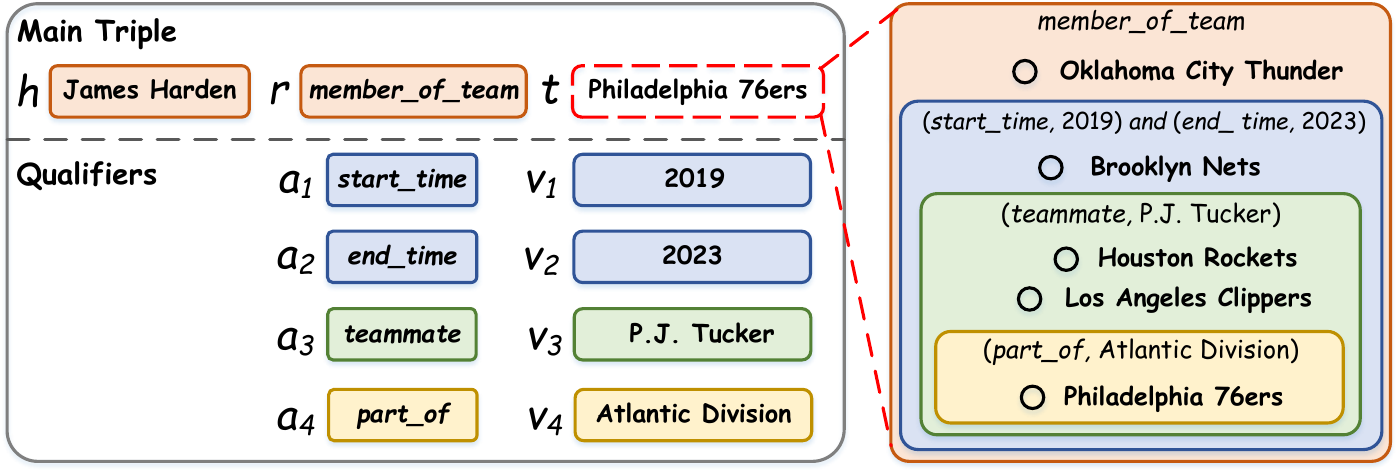}
    \caption{\vic{Qualifier Monotonicity: The box on the right presents all five answers to the query $q =$ (\textit{James Harden}, \textit{member\_of\_team}, \texttt{?}). If qualifiers are added to the query, the number of possible answers is reduced, e.g., if we add the qualifier pairs (\textit{start\_time}, \textit{2019}) and (\textit{end\_time}, \textit{2023}) to $q$,  \textit{Oklahoma City Thunder} is not a possible answer anymore.}}
    \label{figure_instance}
\end{figure}

\vic{Knowledge graphs (KGs) store graph-like  knowledge as a collection of factual triples~\citep{Antoine_2013, Zhiqing_2019}. Each triple $(h, r, t)$ in a KG represents a connection between the head  entity \textit{h} and the tail entity \textit{t} through the relation type $r$, e.g., (\textit{James Harden}, \textit{member\_of\_team}, \textit{Philadelphia 76ers}) in Figure~\ref{figure_instance}.Knowledge graphs provide the backbone of various applications like query answering~\citep{Zhanqiu_2021, Zhiwei_2022}, search~\citep{NguyenVNNP19}, and recommendation systems~\citep{Yuhao_2022}. In hyper-relational knowledge graphs (HKGs), hyper-relational facts enhance binary relational facts by augmenting triples with supplementary knowledge expressed in the form of attribute-value qualifier pairs $(a:v)$. For example, in Figure~\ref{figure_instance}, the triple (\textit{James Harden}, \textit{member\_of\_team}, \textit{Philadelphia 76ers}) is contextualized by the qualifier pairs \{(\textit{start\_time}: \textit{2019}), (\textit{end\_time}: \textit{2023}), (\textit{teammate}: \textit{P.J. Tucker}), (\textit{part\_of}: \textit{Atlantic Division})\}, describing teams of the \textit{Atlantic Division} in which \textit{James Harden} and \textit{P.J. Tucker} played together from \textit{2019} to \textit{2023}. Like standard KGs, HKGs also suffer from incompleteness. The hyper-relational knowledge graph completion (HKGC) task, which aims at inferring missing links in HKGs, has therefore become an important research topic.
}

%Knowledge graphs (KGs) store  real-world knowledge as a collection of binary factual triples~\citep{Antoine_2013, Zhiqing_2019, Rui_2022}, where each triple (\textit{h}, \textit{r}, \textit{t}) represents a relation \textit{r} between head entity \textit{h} and tail entity \textit{t}, e.g., (\textit{James Harden}, \textit{member\_of\_team}, \textit{Philadelphia 76ers}) as shown in Figure~\ref{figure_instance}. Nevertheless, besides binary relational facts, hyper-relational facts that extend triple by associating with additional knowledge in the form of attribute-value qualifier pairs, are also ubiquitous in reality, e.g., as shown in Figure~\ref{figure_instance}, a primary triple (\textit{James Harden}, \textit{member\_of\_team}, \textit{Philadelphia 76ers}) is contextualized by several auxiliary qualifier pairs \{(\textit{start\_time}:\textit{2019}), (\textit{end\_time}: \textit{2023}), (\textit{teammate}:\textit{P.J. Tucker}), (\textit{part\_of}:\textit{Atlantic Division})\}, depicts the teams \textit{James Harden} and \textit{P.J. Tucker} played together from \textit{2019} to \textit{2023}, and corresponding temas belong to the \textit{Atlantic Division}. KGs composed with hyper-relational facts are called hyper-relational knowledge graphs (HKGs), like normal KGs, HKGs also suffer from incompleteness. Hyper-relational knowledge graph completion (HKGC) which infers missing links in HKGs has therefore become an important research topic.

A wide variety of approaches to HKGC have been already proposed, including embedding-based~\citep{Jianfeng_2016,Richong_2018, Paolo_2020, Bo_2023}, transformer-based with graph neural networks (GNNs)~\citep{Mikhail_2020, Harry_2022} and transformer-based without GNNs methods~\citep{Donghan_2021, Quan_2021, Chanyoung_2023, Haoran_2023, Zhiwei_2023}. Despite the substantial progress achieved so far, each of these approaches has its own disadvantages: embedding-based methods can be categorized into n-ary, key-value pairs and main triple+key-value pairs  depending on how facts are defined. While hyper-relational knowledge is effectively represented, the modeling of distinct auxiliary information is done independently. This independence imposes constraints on the mutual interaction between key-value pairs within a given fact. Transformer-based with GNNs methods mainly rely on multi-layer GNNs operations to encode qualifier-pairs information into the main triple,  inevitably introducing noise that  propagates as the number of GNN layers increases. In addition, the use of GNNs  introduces a considerable computational burden. Transformer-based without GNNs methods capture the structure of HKGs within the transformer architecture,  thereby avoiding the increase in computational complexity. However, although these methods are in a dominant position on the HKGC task, they overlook  two important properties: \textit{Two-Stage Reasoning} and \textit{Qualifier Monotonicity}. When inferring missing entities, two-stage reasoning permits to first reason about  binary relational knowledge  and  then, guided by the results from the previous step, reason about hyper-relational knowledge, while qualifier monotonicity implies that adding qualifiers information to triples might allow for additional inferences, without invalidating those derived solely from considering main triples. Specifically, these  properties can be described as follows.

\begin{enumerate}[itemsep=0.5ex, leftmargin=5mm]
\item \vic{\textbf{Two-Stage Reasoning}} 
allows for a two-step reasoning process, facilitating the integration of  coarse-grained inference results derived solely from main triples  and fine-grained inference results obtained from hyper-relational facts with qualifiers. In the initial stage,  coarse-grained results provide an upper bound for correct predictions, which are subsequently  refined in the fine-grained step. For example,  in Figure~\ref{figure_instance},  the initial stage  derives  the following  coarse-grained answer set  from the main triple: $\textit{Ans} = $ \{\textit{Oklahoma City Thunder},$ $ \textit{Houston Rockets}, \textit{Brooklyn Nets}, \textit{Philadelphia 76ers}, \textit{Los Angeles Clippers}\}. $\textit{Ans}$ provides an upper bound to be refined in a  subsequent step:  after adding qualifiers to the main triple, $\textit{Ans}$ is narrowed down to \{\textit{Philadelphia 76ers}\} $\subseteq \textit{Ans}$. Hence, in practice, the outcomes of coarse-grained inference can serve to delimit the decision boundaries of fine-grained predictions, mitigating the occurrence of answer drift phenomena. 

% In HKG, based on the different utilization of knowledge, the inferred answer sets for missing entities can be categoried into two types: fine-grained and coarse-grained. Existing methods primarily focus on incorporating hyper-relational knowledge into the representation process of the main triples to infer fine-grained answers, overlooking the fact that the main triples themselves can directly predict missing entities and obtain coarse-grained answers. As a result, there is a lack of reasoning diversity. In practice, coarse-grained inference results can be used to constrain the decision boundaries of fine-grained predictions, preventing the occurrence of answer drift phenomena. For example, as shown in Figure~\ref{figure_instance}, the coarse-grained answer set for main triple is \{\textit{Oklahoma City Thunder}, \textit{Houston Rockets}, \textit{Brooklyn Nets}, \textit{Philadelphia 76ers}, \textit{Los Angeles Clippers}\}, excludes teams like \{\textit{Golden State Warriors}, \textit{Miami Heat}\}, therefore, after incorporating hyper-relational knowledge, the answer will not exceed the prediction range of the main triple.
\item \textbf{Qualifier Monotonicity} generally implies that for a given hyper-relational query $q$, as the number of qualifier pairs  in $q$ increases, the answer set of $q$ over a HKG might shrink but it never expands, while conversely removing qualifiers from $q$ could only lead to more possible answers. For example,  for the  query (\textit{James Harden},   \textit{member\_of\_team}, \texttt{?})  in Figure~\ref{figure_instance},  \texttt{?} has  five possible answers \{\textit{Oklahoma City Thunder}, \textit{Houston Rockets}, \textit{Brooklyn Nets}, \textit{Philadelphia 76ers}, \textit{Los Angeles Clippers}\}. If the qualifier pairs (\textit{start\_time}: \textit{2019}) and (\textit{end\_time}: \textit{2023}) are added, then \textit{Oklahoma City Thunder}  is not a possible answer anymore. The main challenge for qualifier monotonicity lies in devising a representation that accurately captures its essence. It should posses the capability to faithfully capture the shrinking-like dynamics induced by the incorporation of qualifiers. In the context of embedding-based methods, ShrinkE~\citep{Bo_2023} has explored qualifier monotonicity through box embeddings. However, the inherent limitations of the embedding-based methods lead to a large gap between ShrinkE's accuracy and  that of SoTA models, \zhiweihu{as shown in Table~\ref{table_mixed_percentage_mixed_qualifier} of \S~\ref{main_results}}.

 %ShrinkE~\citep{Bo_2023} adopts box embedding to initially explore the qualifier monotonicity phenomenon, however, the inherent shortcomings of the embedding-based method lead to a large gap between its accuracy and the state-of-arts.

%textbf{Qualifier Monotonicity} generally implies that for a given hyper-relational query, as the number of qualifier pairs increases, the corresponding answer set should shrink or at least not expand, while conversely removing the qualifiers from the query can only return more possible answers, this important properties is qualifier monotonicity. For example, as shown in Figure~\ref{figure_instance}, a query (\textit{James Harden}, \textit{member\_of\_team}, \texttt{?}) with a placeholder \texttt{?} corresponds to five answers \{\textit{Oklahoma City Thunder}, \textit{Houston Rockets}, \textit{Brooklyn Nets}, \textit{Philadelphia 76ers}, \textit{Los Angeles Clippers}\}, when adding qualifier pairs (\textit{start\_time}: \textit{2019}) and (\textit{end\_time}: \textit{2023}), the \textit{Oklahoma City Thunder} in the original answer set will be excluded. Although ShrinkE~\citep{Bo_2023} adopts box embedding to initially explore the qualifier monotonicity phenomenon, however, the inherent shortcomings of the embedding-based method lead to a large gap between its accuracy and the state-of-arts.

\end{enumerate}

To address the above limitations of existing methods, we introduce a monotonicity-aware framework for hyper-relational knowledge representation (\texttt{HyperMono}). To infer the missing entities in hyper-relational queries, \texttt{HyperMono} first introduces the \textbf{\textit{Head Neighborhood Encoder}} module to encode the content of the neighbors of the head entity. It considers coarse-grained and fine-grained neighborhood aggregators, characterized by the use of either main triples or complete hyper-relational facts to determine the neighbors of an entity. \texttt{HyperMono} \zhiweihu{further} implements two-stage reasoning through a   \textbf{\textit{Missing Entity Predictor}} module, which contains two types of predictors: \textit{Triple-based Predictor} and \textit{Qualifier-Monotonicity-aware Predictor}. They differ  on whether only main triples or triples+qualifers are used for prediction. Additionally, \texttt{HyperMono} employs cone embeddings to achieve qualifier monotonicity. We utilize cones instead of boxes (as done in ShrinkE) since the main objective is to obtain an answer set qualified by hyper-relational knowledge in a universal subset space. However, the offsets of boxes are unbounded, and how to find boxes to represent a universal set is thus unclear. Cone embeddings naturally represent any finite universal set and its subsets with a proper aperture, as the angle and aperture of cones are bounded to [-$\pi$, $\pi$] and [0, 2$\pi$]~\citep{Zhanqiu_2021, Chau_2023a}. Our contributions can be summarized as follows:
% propose a model named \textbf{RDQM}, which respects \textit{\textbf{R}easoning \textbf{D}iversity and \textbf{Q}ualifier \textbf{M}onotonicity} for hyper-relational knowledge graph completion. Specifically, to infer the missing tail entity of hyper-relational facts, RDQM first introduces the \textbf{\textit{Head Neighbor Encoder}} module to adequately encode the content of the neighbors of head entity, which considers two types of aggregators, i.e., \textit{Coarse-grained Neighbors Aggregator} without qualifier and \textit{Fine-grained Neighbors Aggregator} with qualifier, according to whether the neighbors contain qualifier information. Furthermore, to achieve reasoning diversity, RDQM designs the \textbf{\textit{Missing Entity Prediction}} module, which contains two types of predictors, i.e., \textit{Triple Direct Predictor} and \textit{Qualifier Monotonicity Predictor}, depending on whether the query to be predicted contains qualifier knowledge. Moreover, RDQM conducts the shrinking of cones operation in the \textit{Qualifier Monotonicity Predictor} to simulate the qualifier monotonicity, which ensures that attaching more qualifier pairs to a primal triple may only narrow down but never enlarges the answer space. Our contributions can be summarized as following three aspects:
\vic{
\begin{itemize}[itemsep=0.5ex, leftmargin=5mm]
\item We propose \texttt{HyperMono}, a competitive monotonicity-aware model  for  hyper-relational knowledge graph completion, which simultaneously considers the two-stage reasoning and qualifier monotonicity properties.
\item We model each qualifier as a cone space shrink transformation that  narrows down the answer set.
\item We conduct  experiments on three real-world datasets under three different conditions, and various ablation studies on parameter sensitivity, complexity analysis, and model  transferability. The obtained results demonstrate the strong performance when compared with SoTA models.
\end{itemize}}

% \zhiwei{Although many works try to use the intersection operation in box embedding to represent the relationship between different boxes~\citep{Hongyu_2020, Nurendra_2021, Zijie_2023, Bo_2023}. However, the main purpose is to obtain an answer set qualified by hyper-relational knowledge in a universal subset space, since the offsets of boxes are unbounded, how to find boxes to represent the universal set is unclear. Fortunately, compared with box, the cone embedding can naturally represent any finite universal set and its subset with a proper aperture (as the aperture of cones are bounded from 0 to 2$\pi$)~\citep{Zhanqiu_2021, Chau_2023a, Chau_2023b}. Based on this reason,}

%% file: sections/related_work.tex
\noindent \textbf{Embedding-based.} 
\vic{Depending on how the facts are represented, embedding-based methods can be classified as  n-ary, key-value pairs and main triple+key-value pairs. (\textit{i}) In n-ary methods, each fact consists of a pre-defined relation \textit{r} with \textit{n} corresponding values, denoted as $r(v_1,v_2,...,v_n)$. For instance, m-TransH~\citep{Jianfeng_2016} and RAE~\citep{Richong_2018} generalize TransH~\citep{Zhen_2014} by projecting all entities onto a relation-specific hyperplane, while m-CP~\citep{Bahare_2020} generalizes Canonical Polyadic (CP) decomposition~\citep{Trouillon_2017}, n-TuckER~\citep{Yu_2020} and GETD~\citep{Yu_2020} generalize  TuckER~\citep{Ivana_2019} to the higher-arity case. However, these methods convert n-ary facts using one abstract relation, ignoring  the semantic distinction of different values for each relation. (\textit{ii}) In key-value pairs methods, each fact  is viewed as a set of attribute-value pairs, denoted as $\{(a_i: v_i)\}_{i=1}^{m}$. For instance,  NaLP~\citep{Paolo_2020} employs a convolutional neural network followed by a multi-layer perceptron to measure the validity of a fact, RAM~\citep{Yu_2021} further encourages the relatedness among different attributes and between an attribute and all corresponding values. However, these methods treat all attribute-value pairs equally and do not distinguish the main triple from the complete fact. (\textit{iii}) Main triple+key-value pairs methods regard each fact as a main triple with attribute-value pairs, denoted as $((h, r, t),\{(a_i: v_i)\}_{i=1}^{m})$. For example, HINGE~\citep{Paolo_2020} and NeuInfer~\citep{Saiping_2020} independently consider each auxiliary pair  in the  main triple. However, as different auxiliary information is modeled independently, these methods do not properly encode the interactions between attribute-value pairs within fact.}

\noindent \textbf{Transformer-based with GNNs.} \vic{StarE~\citep{Mikhail_2020} adopts a message passing based graph encoder CompGCN~\citep{Shikhar_2020} to obtain the entity and relation embeddings, which are further passed through a transformer decoder to obtain a probability distribution over all entities. QUAD~\citep{Harry_2022} simultaneously takes into account the communication from qualifiers to the main triple and the flow of information from the main triple to qualifiers. It also introduces an auxiliary task to further improve performance. However, these methods primarily employ multi-layer GNNs to incorporate hyper-relational knowledge into the main triple, which inevitably introduces two problems. On the one hand, \zhiweihu{although} multi-layers GNNs operations can encode entity-neighborhood information, they may introduce noise, which can be propagated in a cascading fashion as the number of layers increases. On the other hand, GNNs operations incur in  an additional time and space overhead~\citep{Jie_2018, Felix_2019, Yuning_2020}. }

%StarE~\citep{Mikhail_2020} adopts a message passing based graph encoder CompGCN~\citep{Shikhar_2020} to obtain the entity and relation embeddings, which are further passed through a transformer decoder to obtain a probability distribution over all entities. QUAD~\citep{Harry_2022} simultaneously takes into account the communication from qualifiers to primary triple and the flow of information from main triple to qualifiers, besides, introduce an auxiliary task to further improve performance. However, these methods primarily employ multi-layers GNNs to incorporate hyper-relational knowledge into the main triple, which inevitably brings two effects. On the one hand, while multi-layers GNNs operations can encode entity neighbor structured information, they may introduce noise, and can further propagate in a cascading manner with the number of layers increases. On the other hand, GNNs operations incur additional time and space overhead, as previously mentioned in~\citep{Jie_2018, Felix_2019, Yuning_2020}. 

\noindent \textbf{Transformer-based without GNNs.} \vic{Many methods abandon the use of GNNs operations and instead  incorporate hyper-relational knowledge through  various transformer structures. For example, GRAN~\citep{Quan_2021} implements edge-biased fully-connected attention into transformers to encode the graph structure and its heterogeneity. HyNT~\citep{Chanyoung_2023} defines a specialized context transformer to exchange information between the main triple and \vid{its} qualifiers, HAHE~\citep{Haoran_2023} proposes a hierarchical attention structure with global-level and local-level attention to simultaneously model  graph-like  and sequential content. HyperFormer~\citep{Zhiwei_2023} introduces a mixture-of-experts strategy into the feed-forward layers of a transformer to strengthen its representation capabilities. However, these methods have two shortcomings: on the one hand, they only consider a fine-grained inference steps  with hyper-relational knowledge and overlook  coarse-grained  inferences solely based on the main  triples. On the other hand, they do not explicitly embody qualifier monotonicity. We note in passing that  although the embeeding-based method ShrinkE~\citep{Bo_2023} takes qualifier monotonicity into account, its embedding-based nature results in a large gap in performance when compared with the state-of-the-art, \zhiweihu{as shown in Table~\ref{table_mixed_percentage_mixed_qualifier} of \S~\ref{main_results}}.}

%Many methods endeavor to abandon GNNs operations and instead elegantly incorporate hyper-relational knowledge into various modified transformer structures. For example, GRAN~\citep{Quan_2021} adopts edge-biased fully-connected attention into transformer to encode the graph structure and its heterogeneity, HyNT~\citep{Chanyoung_2023} defines a specialized context transformer to exchange information between the primary triple and the qualifiers, HAHE~\citep{Haoran_2023} proposes a hierarchical attention structure with global-level and local-level attention simultaneously models the graphical and squential content, HyperFormer~\citep{Zhiwei_2023} introduces a mixture-of-expert strategy into the feed-forward layers of transformer to strengthen its representation capabilities. However, these methods have two shortcomings: on the one hand, they only consider fine-grained missing entity inference answer set with hyper-relational knowledge and overlook the coarse-grained answer set inferred solely based on the primary triple, so lacking reasoning diversity. On the other hand, they do not explicitly embody qualifier monotonicity, i.e., attaching qualifiers to a certain primary triple may only narrow down but never enlarges the answer set. Although ShrinkE~\citep{Bo_2023} take qualifier monotonicity into account, its embedding-based thought results in a large gap compared with state-of-art.

%% file: sections/method.tex
\subsection{Preliminaries}
\label{section_preliminaries}
%\subsubsection*{Hyper-relational Knowledge Graph Completion.}
A \textit{hyper-relational knowledge graph (HKG)} $ \mathcal G$ is defined as  $\{\mathcal{E}, \mathcal{R}, \mathcal{H}\}$, where $\mathcal{E}$ is a set of entities, $\mathcal{R}$ is a set of relation types, and $\mathcal{H}=\{(h, r, t, \mathcal{Q})~|~ h\in\mathcal{E}, r\in\mathcal{R},t\in\mathcal{E}\}$ is a set of \emph{hyper-relational facts}, such that $(h, r, t)$ denotes the \emph{main triple} and $\mathcal{Q}=\{(a_1 : v_1), \ldots, (a_n : v_n) ~|~ a_i\in\mathcal{R},v_i\in\mathcal{E}\}$ is a set of auxiliary attribute-value \emph{qualifier pairs}. We use $\mathcal G_{\textit{triple}}$ to denote the set of main triples occurring in $\mathcal G$.
Note that when $n=0$, $\mathcal G$ is a standard knowledge graph, \zhiweihu{i.e.,} only containing main triples. Under this representation schema, the main triple (\textit{James Harden}, \textit{member\_of\_team}, \textit{Philadelphia 76ers}) in Figure ~\ref{figure_instance} with its qualifiers can be represented as: \{(\textit{James Harden}: \textit{member\_of\_team}, \textit{Philadelphia 76ers}), (\textit{start\_time}: \textit{2019}), (\textit{end\_time}: \textit{2023}), (\textit{teammate}: \textit{P.J. Tucker}), (\textit{part\_of}: \textit{Atlantic Division})\}. {Building upon the definition of the knowledge graph completion task~\citep{Antoine_2013, Zhiqing_2019}, the \textit{hyper-relational knowledge graph completion (HKGC)} task aims at predicting entities within hyper-relational facts $\mathcal{H}$. Based on the entities to be predicted, HKGC can be classified into two distinct categories:
\begin{enumerate}[itemsep=0.5ex, leftmargin=5mm]
\item \textit{Head / Tail}: the entities to be predicted are exclusively located within the main triple, \textit{i.e.,} predict a missing entity \texttt{?} in  $(\texttt{?}, r, t, \mathcal{Q})$ or $(h, r, \texttt{?}, \mathcal{Q})$.
\item \textit{All Entities}: the entities to be predicted may not only reside within the main triple but could also be present in the qualifier pairs $\mathcal{Q}$, \textit{i.e.,} predict a missing entity \texttt{?} in $(\texttt{?}, r, t, \mathcal{Q})$ or $(h, r, \texttt{?}, \mathcal{Q})$ or $\{(a_1 : v_1), \ldots, (a_i : \texttt{?}), \ldots, (a_n : v_n)\}$.
\end{enumerate}
Without loss of generality, in what follows we concentrate on predicting a tail entity \texttt{?} in $(h, r, \texttt{?}, \mathcal{Q})$. The other settings  can be handled in a similar manner.
}

\subsubsection*{Qualifier Monotonicity.} Let $q_1=(h, r, \texttt{?}, \mathcal{Q}_1)$ and $q_2=(h, r, \texttt{?}, \mathcal{Q}_2)$ be two queries that share the same main triple $(h, r, \texttt{?})$ and \zhiweihu{$\mathcal{Q}_1$, $\mathcal{Q}_2$ satisfy} $\mathcal{Q}_1 \subseteq \mathcal{Q}_2$. %\zhiweihu{i.e., $\mathcal{Q}_1$ is a subset of $\mathcal{Q}_2$}. 
We say that the \emph{qualifier monotonicity property} holds over a hyper-relational knowledge graph $\mathcal G$ iff $\textit{Ans} (q_2, \mathcal G) \subseteq  \textit{Ans} (q_1, \mathcal G)$, where $\textit{Ans} (q_i, \mathcal G)$ denotes the set of answers of the query $q_i$ over $\mathcal G$. Intuitively,   this property implies that adding qualifier pairs to a query does not enlarge its answer set, and conversely, removing qualifier pairs  can only result in more answers. For example, consider the queries $q_1= $ (\textit{James Harden}, \textit{member\_of\_team}, \texttt{?}) \zhiweihu{with} $\mathcal{Q}_1=$ \{(\textit{start\_time}: \textit{2019}), (\textit{end\_time}: \textit{2023})\} and $q_2= $(\textit{James Harden}, \textit{member\_of\_team}, \texttt{?}) \zhiweihu{with} $\mathcal{Q}_2=$ \{(\textit{start\_time}: \textit{2019}), (\textit{end\_time}: \textit{2023}), (\textit{teammate}: \textit{P.J. Tucker})\} in Figure ~\ref{figure_instance}. It is clear that \zhiweihu{the main triple of $q_1$ and $q_2$ are the same and} $\mathcal{Q}_1 \subseteq \mathcal{Q}_2$, \vid{and the corresponding answer sets} $\textit{Ans}(q_1, \mathcal G)=$  \{\textit{Brooklyn Nets}, \textit{Houston Rockets}, \textit{Los Angeles Clippers}, \textit{Philadelphia 76ers}\} and $\textit{Ans}(q_2, \mathcal G)=$ \{\textit{Houston Rockets}, \textit{Los Angeles Clippers}, \textit{Philadelphia 76ers}\} satisfy $\textit{Ans}(q_2, \mathcal G)  \subseteq \textit{Ans}(q_1, \mathcal G)$.
\subsubsection*{Cone Embedding.} We present here only necessary background on cone embeddings, for a full discussion we refer the interested reader to~\citep{Zhanqiu_2021, Chau_2023a, Chau_2023b}. A \textit{d}-dimensional sector-cone can be parameterized by $\mathcal{C}^d(\alpha, \beta)=((\alpha^1, \beta^1),...,(\alpha^d, \beta^d))$, where $\alpha^i\in[-\pi,\pi)$ represents the angle between the symmetry axis of the \textit{i}-th-dimension sector-cone and the positive \textit{x} axis, and $\beta^i\in[0, 2\pi)$ represents the aperture of the \textit{i}-th-dimension sector-cone. We omit the superscript of $\mathcal{C}^d$  if it is clear from context. A variety of operations on  cone embeddings can be defined, here we introduce the  projection $\mathcal{P}$ and intersection $\mathcal{I}$ operations:
%We adopt the same definitions and propositions in~\citep{Zhanqiu_2021, Chau_2023a, Chau_2023b}, each \textit{d}-dimensional sector-cone can be parameterized by $\mathcal{C}^d(\alpha, \beta)=((\alpha^1, \beta^1),...,(\alpha^d, \beta^d))$, where $\alpha^i\in[-\pi,\pi)$ represents the angle between the symmetry axis of the \textit{i}-th-dimension sector-cone and the positive \textit{x} axis, and $\beta^i$ represents the aperture of the \textit{i}-th-dimension sector-cone, we leave the superscript of $\mathcal{C}^d$ away if it is clear from context. Figure~\ref{figure_shrink}(a) shows a 1-dimensional sector-cone, it can be expressed as $\mathcal{C}(\alpha, \beta)$. The cone embedding contains a variety of operations, here we introduce the simplified projection $\mathcal{P}$ and intersection $\mathcal{I}$ operations as follows. Regarding the details involved in different operations, considering that this is not our focus, readers can refer to the corresponding literature~\citep{Zhanqiu_2021}.
\begin{enumerate}[itemsep=0.5ex, leftmargin=5mm]
\item \textit{Projection Operator} $\mathcal{P}$.  $\mathcal{C}_2$=$\mathcal{P}$($\mathcal{C}_1$, $\mathcal{C}_r$) computes the \textit{projection} from an input cone $\mathcal{C}_1$ as a \textit{head} entity set to the set of \textit{tail} entities $\mathcal{C}_2$ via the specific relation cone $\mathcal{C}_r$, see Figure~\ref{figure_shrink}(b).
% via the relation cone $\mathcal{C}_2$, see Figure~\ref{figure_shrink}(b). }

\item \vic{\textit{Intersection Operator} $\mathcal{I}$. $\mathcal{I}$($\mathcal{C}_1$, $\mathcal{C}_2$) computes the \textit{intersection} of the set of entities $\mathcal{C}_1$ and the set of entities $\mathcal{C}_2$, see Figure~\ref{figure_shrink}(c).}
%of each element in one set $\mathcal{C}_1$ of entities and the corresponding element in the other entity set $\mathcal{C}_2$, see Figure~\ref{figure_shrink}(c).
\end{enumerate}

\subsection{Model Architecture}
%\textcolor{red}{TODO: We present bla bla}
We present the two main components of \texttt{HyperMono}'s architecture: Head Neighborhood Encoder (\S~\ref{entity_neighbor_encoder}) and Missing Entity Predictor (\S~\ref{sec:MEP}).

\subsubsection{HNE: Head Neighborhood Encoder}\label{entity_neighbor_encoder}
\begin{figure*}[!htp]
    \centering
    \includegraphics[width=1\textwidth]{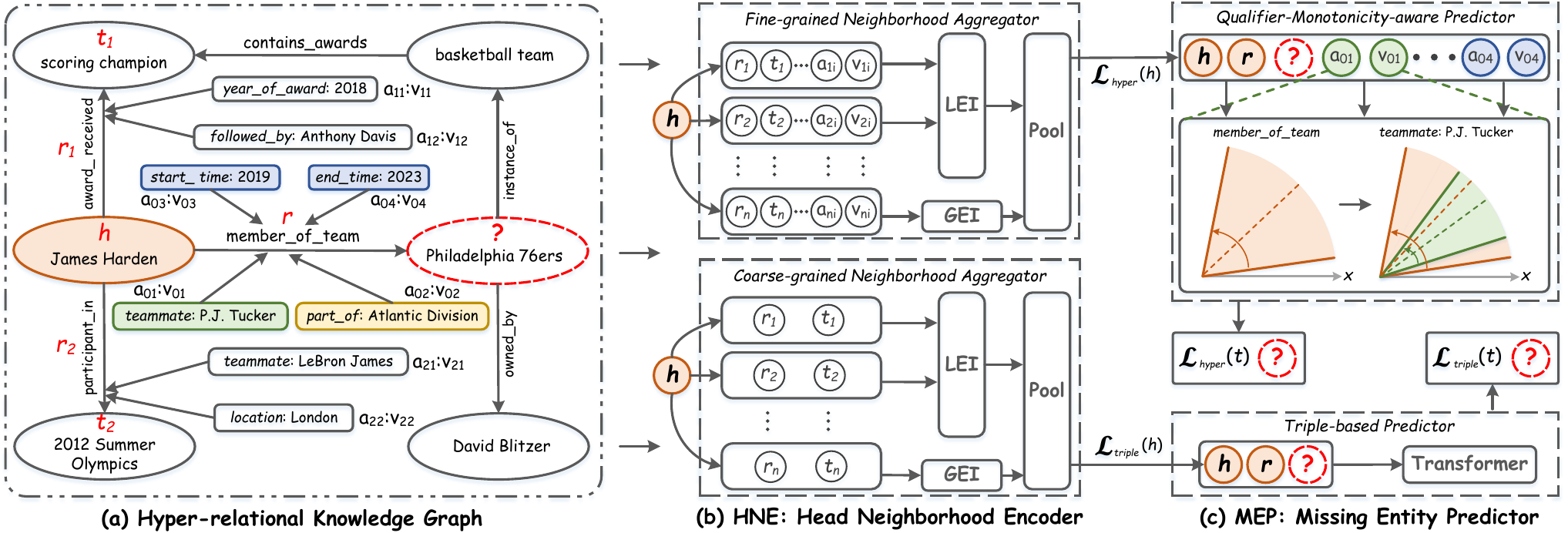}
    \caption{\zhiweihu{\texttt{HyperMono}}'s model structure, containing two modules: Head \zhiweihu{Neighborhood} Encoder (HNE) and Missing Entity \zhiweihu{Predictor} (MEP). \zhiweihu{HNE module contains two sub-modules: Coarse-grained Neighborhood Aggregator (CNA) and Fine-grained Neighborhood Aggregator (FNA). MEP module contains two sub-modules: Triple-based Predictor (TP) and Qualifier-Monotonicity-aware Predictor (QMP).}}
    \label{figure_model}
\end{figure*}

To infer the missing tail entity in the  triple $(h, r, \texttt{?})$, it is important to  not only  consider the input embeddings of $h$ and $r$, but also look at how $h$ relates to its neighboring entities. For instance, in Figure~\ref{figure_model}(a), for the triple (\textit{James Harden}, \textit{member\_of\_team}, \texttt{?}), the relational neighbors (\textit{award\_received}, \textit{scoring champion}) and (\textit{participant\_in}, \textit{2012 Summer Olympics}) of the head entity \textit{James Harden} can help us infer that \textit{James Harden} is an athlete and  not a movie star. The same phenomenon is witnessed for hyper-relational facts $(\textit{h}, \textit{r}, \texttt{?}, \mathcal{Q}$), i.e., we can exploit the information about how $h$ relates to its neighbors (including or not including its qualifiers) to obtain an enhanced head entity embedding representation.
%after multi-perspective aggregation, thereby providing richer knowledge when predicting the tail entity placeholder \texttt{?}. }

%For an incomplete KG triple (\textit{h}, \textit{r}, \texttt{?}) of $\mathcal{G}_{triple}$ to predict the tail entity placeholder \texttt{?}, in addition to only considering the input embedding of \textit{h} and \textit{r}, we also observe that the embedding of \textit{h} have a strong correlation with its relational neighbors. For instance, in Figure~\ref{figure_model}(a), for incomplete triple (\textit{James Harden}, \textit{member\_of\_team}, \texttt{?}), the relational neighbors (\textit{award\_received}, \textit{scoring champion}) and (\textit{participant\_in}, \textit{2012 Summer Olympics}) of head entity \textit{James Harden} can infer that \textit{James Harden} is an athlete, not a movie star. The same thing happens with the hyper-relational fact (\textit{h}, \textit{r}, \texttt{?}, $\mathcal{Q}$) of $\mathcal{G}_{hyper}$, i.e., we can exploit the relational neighbors information with hyper-relational knowledge of the head entity \textit{h} to obtain the enhanced head entity embedding representation after multi-perspective aggregation, thereby providing richer knowledge when predicting the tail entity placeholder \texttt{?}.

To adequately encode the content of the neighbors of a head entity, as a first step, we introduce the \textit{\textbf{H}ead \textbf{N}eighborhood \textbf{E}ncoder} (\texttt{\textbf{HNE}}) module, cf.\  Figure~\ref{figure_model}(b).  The HNE module considers two types of relational contexts, characterized by the use of either main triples or complete hyper-relational facts  to ascertain the neighbors of a given entity: \textit{\textbf{C}oarse-grained \textbf{N}eighbor \textbf{A}ggregator} (\texttt{\textbf{CNA}}), only considering  main triples in  $ \mathcal{G}_{triple}$; \textit{\textbf{F}ine-grained \textbf{N}eighbor \textbf{A}ggregator} (\texttt{\textbf{FNA}}), considering hyper-relational facts in $\mathcal G$. The resulting neighbor-aware embeddings from the CNA and FNA modules will be used in the subsequent module (\S \ref{sec:MEP}) to provide further evidence, based on the neighbors of $h$, for the prediction of the \texttt{?} entity.
%
%To adequately encode the content of the neighbors of head entity, we introduce the \textit{\textbf{H}ead \textbf{N}eighbor \textbf{E}ncoder} (\texttt{\textbf{HNE}}) module as shown in Figure~\ref{figure_model}(b). Specifically, according to whether the neighbors contain qualifier pairs, HNE module considers two types of relational contexts, i.e., \textit{\textbf{C}oarse-grained \textbf{N}eighbors \textbf{A}ggregator} (\texttt{\textbf{CNA}}) from $\mathcal{G}_{triple}$ and \textit{\textbf{F}ine-grained \textbf{N}eighbors \textbf{A}ggregator} (\texttt{\textbf{FNA}}) from $\mathcal{G}_{hyper}$. As we mentioned in qualifier monotonicity, adding qualifiers will shrink the answer space, so the answer obtained without introducing any qualifiers will be more coarse-grained, correspondingly, the answer set obtained after introducing qualifier information will become fine-grained.

\smallskip
\noindent \textbf{Coarse-grained \zhiweihu{Neighborhood} Aggregator.} 
For a main triple $(h,r, \texttt{?})$ from $\mathcal{G}_{triple}$, we use the following three steps to encode information about the neighbors of $h$ into its representation: 
%For a certain incomplete triple (\textit{h}, \textit{r}, \texttt{?}) from $\mathcal{G}_{triple}$, we use the following three steps to encode head entity neighbors into the entity \textit{h} representation, to further predict the tail entity placeholder \texttt{?}:
\begin{enumerate}[itemsep=0.5ex, leftmargin=5mm]
\item \textit{Neighbor-aware Embeddings.} Let $N_h = \{(h,r',t') ~|~$ $ (h,r',t') \in \mathcal G_{\textit{triple}}\}$ be the set of neighbors of $h$. For a triple $\textit{tr}= (h,r',t') \in N_h$, we use the placeholder \texttt{[mask]} in place of $h$ to obtain  the neighbor sequence $\{[\texttt{mask}], r', t'\}$. Then, we apply a standard transformer encoder~\citep{Ashish_2017} to obtain the \texttt{[mask]} token embedding $\textbf{\textit{N}}_{\textit{tr}}^{h} \in \mathbb{R}^{1 \times d}$, $d$ represents the embedding dimension. Clearly, if $n =|N_h|$, \textit{n} different \texttt{[mask]} representations will be obtained. So, we sum up and average all \texttt{[mask]} representations to obtain an aggregated embedding $\textbf{\textit{N}}^{h}$ considering all neighbors of \textit{h}:   $\textbf{\textit{N}}^{h}=mean(\sum_{tr \in N_h}\textbf{\textit{N}}_{\textit{tr}}^{h})$.

%Let $(h, r_i, t_i) \in \mathcal{G}_{triple}$ denotes \textit{i}-th relational neighbors of entity \textit{h}. Firstly, we use the placeholder \texttt{[mask]} in place of \textit{h} to get the \textit{i}-th neighbor sequence $\{[\texttt{mask}], r_i, t_i\}$. Then, we apply a standard Transformer encoder~\citep{Ashish_2017} to obtain \texttt{[mask]} token embedding $\textbf{\textit{N}}_i^{h} \in \mathbb{R}^{1 \times d}$, $d$ represents the embedding dimension. Finally, assuming that \textit{h} contains \textit{n} relational neighbors, therefore \textit{n} different \texttt{[mask]} representations will be obtained, further we sum up and average all \texttt{[mask]} representation to obtain the aggregated embedding of neighbors of \textit{h} as $\textbf{\textit{N}}^{h}=mean(\sum_{i=1}^{n}\textbf{\textit{N}}_i^{h})$.

\item \textit{Local and Global  Entity Inference.} Different neighbors of $h$ might provide information about $h$ from different perspectives. We thus propose the \textit{\textbf{L}ocal \textbf{E}ntity \textbf{I}nference} (\texttt{\textbf{LEI}}) mechanism that independently considers each neighbor to infer the  entity corresponding to  \textit{h}. LEI focuses on a single neighbor during inference, reducing the interference from irrelevant information. In practice, for  $\textit{tr} \in N_h$, we perform a matrix multiplication operation on $\textbf{\textit{N}}_{\textit{tr}}^{h}$ and $\textbf{\textit{V}}_{\!\textit{ent}}$ to obtain the entity score vector $\textbf{\textit{P}}_{\textit{tr}}^{h}\in\mathbb{R}^{1\times N}=\textbf{\textit{N}}_{\textit{tr}}^{h}\cdot\textbf{\textit{V}}_{\! \textit{ent}}^{\top}$, where $\textbf{\textit{V}}_{\! \textit{ent}} \in \mathbb{R}^{N \times d}$ denotes the embedding matrix for all entities in the dataset, $\textbf{\textit{V}}_{\! \textit{ent}}^{\top}$ is the transpose of $\textbf{\textit{V}}_{\! \textit{ent}}$,  and $N = |\mathcal{E}|$. However, in some cases, it is difficult to infer a complex entity from a single neighbor. Therefore, we additionally propose a \textit{\textbf{G}lobal \textbf{E}ntity \textbf{I}nference} (\texttt{\textbf{GEI}}) mechanism which averages the entity score vector predicted by different relational neighbors to aggregate them: $\textbf{\textit{P}}_{G}^{h}=mean(\sum_{\textit{tr} \in N_h}\zhiweihu{\textbf{\textit{P}}_{\textit{tr}}^{h}})$.

%Sometimes, according to our observation, it is difficult to infer some complex entity from a single neighbor. Therefore, we further propose a \textit{\textbf{G}lobal \textbf{I}nference \textbf{E}ntity} (\texttt{\textbf{GIE}}) mechanism which averages the entity score vector predicted by different relational neighbors to aggregate different entities' relational neighbors to infer the possible entity, which can be denoted as $\textbf{\textit{P}}_{G}^{h}=mean(\sum_{i=1}^{n}\textbf{\textit{P}}_i^{h})$.

\item \textit{Aggregating Inference Results.} The LEI and GEI mechanisms will generate multiple entity inference results, to unify them, inspired by ~\citep{Zhiwei_2022, Weiran_2021}, we adopt an exponentially weighted pooling method to generate the final inference result:
\begin{equation}
\label{equation_1}
\textbf{\textit{P}}^h=pool(\{\textbf{\textit{P}}_{\textit{tr}_1}^{h}, \textbf{\textit{P}}_{\textit{tr}_2}^{h},...,\textbf{\textit{P}}_{\textit{tr}_n}^{h}, \textbf{\textit{P}}_G^{h}\})=\sum_{i=1}^{n, G}\omega_i\textbf{\textit{P}}_{\textit{tr}_i}^h
\end{equation}
where $\omega_i=\frac{{\rm exp}\,\gamma\textbf{\textit{P}}_{\textit{tr}_i}^h}{\sum_{j=1}^{n, G}{{\rm exp}\,\gamma\textbf{\textit{P}}_{\textit{tr}_j}^h}}$, $\gamma$ is a hyperparameter that controls the temperature of the pooling process. We further apply a softmax function $\theta$ to $\textbf{\textit{P}}^h$: $\textbf{\textit{p}}^h=\theta(\textbf{\textit{P}}^h)$ to map the scores between 0 and 1. Finally, we calculate the cross-entropy loss between the prediction logit $\textbf{\textit{p}}^h$ and the corresponding label $\textbf{\textit{L}}^h_1$: $\mathcal{L}_{\textit{triple}}(h)=\texttt{CrossEntropy}(\textbf{\textit{p}}^h, \textbf{\textit{L}}^h_1)$.

%where $\omega_i=\frac{{\rm exp}\,\gamma\textbf{\textit{P}}_i^h}{\sum_{j=1}^{n, G}{{\rm exp}\,\gamma\textbf{\textit{P}}_j^h}}$, $\gamma$ is a hyperparameter that controls the temperature of the pooling process. We further apply a softmax function $\theta$ to $\textbf{\textit{P}}^h$, denoted as $\textbf{\textit{p}}^h=\theta(\textbf{\textit{P}}^h)$ to map the scores between 0 and 1. Finally, we calculate the cross-entropy loss between the prediction logit $\textbf{\textit{p}}^h$ and corresponding label $\textbf{\textit{L}}^h_1$ as $\mathcal{L}_{triple}(h)=\texttt{CrossEntropy}(\textbf{\textit{p}}^h, \textbf{\textit{L}}^h_1)$.
\end{enumerate}

\smallskip
\noindent \textbf{Fine-grained \zhiweihu{Neighborhood} Aggregator.} In this module, we follow similar steps as in the CNA module to obtain an aggregated representation of  $h$, but with hyper-relational knowledge. For a  hyper-relational fact $(h, r, \texttt{?}, \mathcal{Q}) \in \mathcal H$, in \textit{the  neighbor-aware embeddings step}, after masking out  $h$ for all its neighbors, we obtain $\textbf{\textit{M}}^{h}$ defined as  $=mean(\sum_{ \textit{hrf} \in N_{h, \mathcal H}}\textbf{\textit{M}}_{\textit{hrf} }^{h})$, where \vi{$N_{h, H} = \{(h,r',t',\mathcal Q') ~|~ (h,r',t',\mathcal Q') \in \mathcal H )\}$ }. In \textit{the local and global entity inference step}, we obtain entity inference results  for all elements in $N_{h, H}$: $\{\textbf{\textit{Q}}_{\textit{hrf}_1}^{h}, \textbf{\textit{Q}}_{\textit{hrf}_2}^{h},...,\textbf{\textit{Q}}_{\textit{hrf}_m}^{h}, \textbf{\textit{Q}}_G^{h}\}$, such that $|N_{h, H}|=m$. In the \textit{aggregating inference results} step, we also employ a pooling strategy to aggregate multiple inference results into a single prediction $\textbf{\textit{Q}}^h$. Finally, we calculate the cross-entropy loss between the $\textbf{\textit{q}}^h=\theta(\textbf{\textit{Q}}^h)$ and the true label $\textbf{\textit{L}}^h_2$ as: $\mathcal{L}_{\textit{hyper}}(h)=\texttt{CrossEntropy}(\textbf{\textit{q}}^h, \textbf{\textit{L}}^h_2)$.

%Given a hyper-relational fact (\textit{h}, \textit{r}, \texttt{?}, $\mathcal{Q}$) from $\mathcal{G}_{hyper}$, we adopt the similar steps as in \texttt{CNA} module to obtain the aggregated representation of the head entity \textit{h} with hyper-relational knowledge. Specially, in \textit{obtain neighbor embeddings} step, we are able to obtain the representation $\textbf{\textit{M}}^{h}=mean(\sum_{j=1}^{m}\textbf{\textit{M}}_j^{h})$ after masking out the \textit{h} position with \textit{m} different neighbor masked. In \textit{local and global inference entity} step, utilizing the results from the previous step, we introduce both \texttt{LIE} and \texttt{GIE} mechanisms to obtain various entity inference results $\{\textbf{\textit{Q}}_1^{h}, \textbf{\textit{Q}}_2^{h},...,\textbf{\textit{Q}}_m^{h}, \textbf{\textit{Q}}_G^{h}\}$. In \textit{aggregate inference results} step, we also employ a pooling strategy to aggregate multiple inference results into a single prediction $\textbf{\textit{Q}}^h$. Furthermore, we calculate the cross-entropy loss between the $\textbf{\textit{q}}^h=\theta(\textbf{\textit{Q}}^h)$ and the true label $\textbf{\textit{L}}^h_2$, denoted as $\mathcal{L}_{hyper}(h)=\texttt{CrossEntropy}(\textbf{\textit{q}}^h, \textbf{\textit{L}}^h_2)$.

\subsubsection{MEP: Missing Entity Predictor}\label{sec:MEP} The CNA and FNA modules  introduced the supervision signals $\mathcal{L}_{\textit{triple}}(h)$ and $\mathcal{L}_{\textit{hyper}}(h)$ for the possible values of $h$. However, in the hyper-relational knowledge graph completion task, the neighborhood information of the head entity $h$ can be used as prior knowledge to infer the missing placeholder \texttt{?} results. To effectively make use of the neighborhood information, while faithfully capturing the two-stage reasoning and qualifier monotonicity properties, we introduce the \textit{\textbf{M}issing \textbf{E}ntity \textbf{P}redictor} (\texttt{\textbf{MEP}})
module. The \textit{\textbf{T}riple-based \textbf{P}redictor} (\texttt{\textbf{TP}}) component of MEP accomplishes coarse-grained inferences by solely considering main triples. Based on the results from the TP sub-module, the \textit{\textbf{Q}ualifier-\textbf{M}onotonicity-aware \textbf{P}redictor} (\texttt{\textbf{QMP}}) subsequently obtains fine-grained inferences by considering hyper-relational facts with qualifiers, while properly representing the qualifier monotonicity property through cone embeddings. We note that the interaction between the coarse-grained and the fine-grained stages is implicitly captured through the loss function calculation. This naturally models the interaction between these stages, without  having to  look for `correct' restrictions to explicitly capture it. We next describe the details of the TP and QMP components of MEP.

\smallskip
\noindent \textbf{Triple-based Predictor.} For an incomplete triple $(h,r,\texttt{?})$, we employ the following three steps to predict  entities at the placeholder \texttt{?} location: (\textit{i}) We replace the placeholder \texttt{?} with the [\texttt{mask}] token to obtain input sequences $\{h,r, \texttt{[mask]}\}$. Further, we use the $\textbf{\textit{N}}^{h}$ embedding obtained at  Step (1) of the CNA  module  as the input embedding of $h$, and randomly initialize the input embedding of \textit{r}. (\textit{ii}) We use a transformer encoder to obtain token embeddings \{$\textbf{\textit{F}}^{h}$, $\textbf{\textit{F}}^{r}$, $\textbf{\textit{F}}^{[\texttt{mask}]}$\}. (\textit{iii}) Finally, we use $\textbf{\textit{p}}^{t}=\theta(\textbf{\textit{F}}^{[\texttt{mask}]}\cdot\textbf{\textit{V}}_{ent}^{\top})$ to infer the most likely  probability  for entities at the mask position. Additionally, we calculate the cross-entropy loss between the prediction logit $\textbf{\textit{p}}^{t}$ and ground truth label $\textbf{\textit{L}}^t_1$ as $\mathcal{L}_{\textit{triple}}(t)=\texttt{CrossEntropy}(\textbf{\textit{p}}^t, \textbf{\textit{L}}^t_1)$.

%For an incomplete triple (\textit{h}, \textit{r}, \texttt{?}), we employ the following three steps to predict potential entities at the placeholder \texttt{?} location: (\textit{i}) firstly, we replace the placeholder \texttt{?} with [\texttt{mask}] token to obtain input sequences \{\textit{h}, \textit{r}, \texttt{[mask]}\}, further we adopt the $\textbf{\textit{N}}^{h}$ obtained at the step \textit{obtain neighbor embeddings} of module \texttt{CNA} as the input embedding of \textit{h}, and randomly initialize the input embedding of \textit{r}. (\textit{ii}) then, we adopt Transformer encoder to obtain token embeddings \{$\textbf{\textit{F}}^{h}$, $\textbf{\textit{F}}^{r}$, $\textbf{\textit{F}}^{[\texttt{mask}]}$\}. (\textit{iii}) finally, we use $\textbf{\textit{p}}^{t}=\theta(\textbf{\textit{F}}^{[\texttt{mask}]}\cdot\textbf{\textit{V}}_{ent}^{\top})$ to infer the most likely entity probability at the mask position, further we calculate the cross-entropy loss between the prediction logit $\textbf{\textit{p}}^{t}$ and ground truth label $\textbf{\textit{L}}^t_1$ as $\mathcal{L}_{triple}(t)=\texttt{CrossEntropy}(\textbf{\textit{p}}^t, \textbf{\textit{L}}^t_1)$.

\smallskip
\noindent \textbf{\zhiweihu{Qualifier-Monotonicity-aware Predictor}.} The previous module only considers the main triple  $(h,r, \texttt{?})$ in $(h,r, \texttt{?}, \mathcal Q= \{(a_1: v_1), \ldots, (a_s: v_s)\})$, disregarding qualifier pairs. To fully leverage the qualifier pairs information in the hyper-relational fact,  building upon the  qualifier monotonicity property, we propose the \textit{Qualifier-Monotonicity-aware Predictor} module, composed of the following three sub-modules:

% Unlike triple direct prediction part which only considers the main triple (\textit{h}, \textit{r}, \texttt{?}) in (\textit{h}, \textit{r}, \texttt{?}, $\mathcal{Q}$), the qualifier monotonicity prediction part simultaneously takes into account the content of qualifier pairs $\mathcal{Q}$. For convenience of presentation, we expand $\mathcal{Q}=\{(a_1, v_1),..., (a_m, v_m)\}$, so the hyper-relational fact (\textit{h}, \textit{r}, \texttt{?}, $\mathcal{Q}$) will become (\textit{h}, \textit{r}, \texttt{?}, $\{(a_1, v_1),..., (a_s, v_s)\}$). To make full use of the qualifier pairs information in the hyper-relational fact built upon the foundation of considering qualifier monotonicity, we propose the \textit{Qualifier Monotonicity Predictor} module, which contains \textit{Obtain Token Embeddings}, \textit{Cone Shrinking}, and \textit{Inference Results} three steps.

\smallskip
\noindent \textit{Token Embeddings.} Like in the \textit{triple-based predictor} module, we first replace the placeholder \texttt{?} with the \texttt{[mask]} token to obtain the input sequence $\{h,r, \texttt{[mask]}$, $\{(a_1, v_1),..., (a_s, v_s)\}$\}. We initialize the input embedding of \textit{h} using the $\textbf{\textit{M}}^{h}$ embedding obtained in the FNA module,  and the remaining input embeddings of \textit{r}, $a_i$, and $v_i$ are  randomly initialized. Then,  we use a transformer encoder to obtain the output embedding representation of each token: $\textbf{\textit{S}}=$\{$\textbf{\textit{E}}^{h}$, $\textbf{\textit{E}}^{r}$, $\textbf{\textit{E}}^{[\texttt{mask}]}$, \{($\textbf{\textit{E}}^{a_1}$, $\textbf{\textit{E}}^{v_1}$),...,($\textbf{\textit{E}}^{a_s}$, $\textbf{\textit{E}}^{v_s}$)\}\}.

% Consistent with \texttt{TDP} module, firstly, we replace the placeholder \texttt{?} with \texttt{[mask]} token to obtain the input sequence \{\textit{h}, \textit{r}, \texttt{[mask]}, $\{(a_1, v_1),..., (a_s, v_s)\}$\}, we initialize the input embedding of \textit{h} using the $\textbf{\textit{M}}^{h}$ obtained by \texttt{FNA} module, and the remaining input embeddings of \textit{r}, $a_i$, and $v_i$ are initialized randomly. Next, we use the Transformer encoder to obtain the output embedding representation of each tokens as $\textbf{\textit{S}}=$\{$\textbf{\textit{E}}^{h}$, $\textbf{\textit{E}}^{r}$, $\textbf{\textit{E}}^{[\texttt{mask}]}$, \{($\textbf{\textit{E}}^{a_1}$, $\textbf{\textit{E}}^{v_1}$),...,($\textbf{\textit{E}}^{a_s}$, $\textbf{\textit{E}}^{v_s}$)\}\}.

\begin{figure}[t!]
    \centering
    \includegraphics[width=0.48\textwidth]{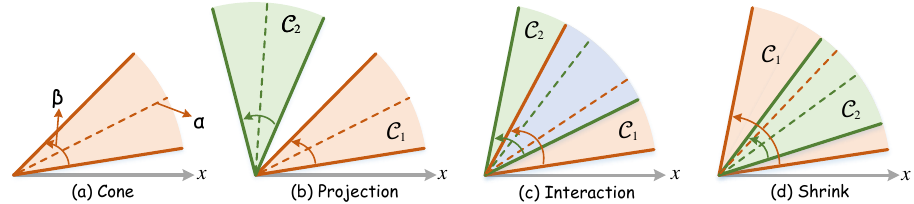}
    \caption{An overview of cone embedding with projection $\mathcal{P}$, intersection $\mathcal{I}$ and shrink $\mathcal{S}$, the embedding dimension $d$=1.}
    \label{figure_shrink}
\end{figure}

\smallskip
\noindent \textit{Cone Shrinking.} 
To faithfully capture the monotonicity property, we will resort to a geometric  representation based on  cone embeddings. We represent main triples  as  cones and each qualifier is modeled  as a cone that shrinks  the cone of the main triple to which the qualifier has  been added. This shrinkage will be implemented  through a reduction of the axis and aperture of  main triples cones.  Thus, our cone-based representation effectively captures that attaching qualifiers to a main triple  might narrow down  the answer set, but never enlarge it. Technically, our goal  is to predict the most likely entity category of $\textbf{\textit{E}}^{[\texttt{mask}]}$ using within the cone space the different types of  token embeddings in $\textbf{\textit{S}}=$\{$\textbf{\textit{E}}^{h}$, $\textbf{\textit{E}}^{r}$, $\textbf{\textit{E}}^{[\texttt{mask}]}$, \{($\textbf{\textit{E}}^{a_1}$, $\textbf{\textit{E}}^{v_1}$),...,($\textbf{\textit{E}}^{a_s}$, $\textbf{\textit{E}}^{v_s}$)\}\}. To this aim, we introduce the \textit{\textbf{C}one \textbf{S}hrink \textbf{B}lock} (\textbf{\texttt{CSB}}) module, comprising the following three steps. We also  prove that cones respect the qualifier monotonicity property  in \textcolor{blue}{\textbf{\hyperlink{proof_of_propositions}{Appendix I}}}.

\begin{enumerate}[itemsep=0.5ex, leftmargin=5mm]
\item We first convert the different types of token embeddings in $\textbf{\textit{S}}$ into the cone space. Specifically, the  head entity token embedding $\textbf{\textit{E}}^{h}$ is converted to the cone $\mathcal{C}_h=\mathcal{C}(\alpha_h, \beta_h)$, and the relation to the cone $\mathcal{C}_r=\mathcal{C}(\alpha_r, \beta_r)$, with  $\alpha_h$, $\beta_h$, $\alpha_r$, and $\beta_r$ defined as follows:
%
%Firstly, we need to convert various role token embeddings into cone embedding space. Specifically, we convert head entity token embedding $\textbf{\textit{E}}^{h}$ to cone $\mathcal{C}_h=\mathcal{C}(\alpha_h, \beta_h)$, and relation token embedding to cone $\mathcal{C}_r=\mathcal{C}(\alpha_r, \beta_r)$, the definitions of $\alpha_h$, $\beta_h$, $\alpha_r$, and $\beta_r$ are as shown in Equation~\ref{equation_2}:
\begin{equation}
\label{equation_2}
\left\{
	\begin{array}{l}
	\alpha_h=f(\texttt{MLP}(\textbf{\textit{E}}^{h})),\,\,\beta_h=g(\texttt{MLP}(\textbf{\textit{E}}^{h}))\vspace{1.2ex}\\
    
    \alpha_r=f(\texttt{MLP}(\textbf{\textit{E}}^{r})),\,\,\beta_r=g(\texttt{MLP}(\textbf{\textit{E}}^{r}))\vspace{1.2ex}\\

    f(x_i)=\pi\texttt{tanh}(\lambda_1x_i), \,\,g(y_i)=\pi\texttt{tanh}(\lambda_2y_i)+\pi\\
    
	\end{array} \right.
\end{equation}
where \texttt{MLP}: $\mathbb{R}^d \rightarrow \mathbb{R}^d$ is a multi-layer perceptron network, the functions $f$ and $g$ are used to scale the semantic center axis and aperture into their normal ranges: $\alpha_h=f(\textbf{x})=[\alpha_h^1,...,\alpha_h^d; \alpha_h^i=f(x_i)]$, $\alpha_h^i$ and $f(x_i)$ denote the \zhiweihu{\textit{i}}-th element of $\alpha_h$ and $f(\textbf{x})$; $\beta_h=g(\textbf{y})=[\beta_h^1,...,\beta_h^d; \beta_h^i=g(y_i)]$, $\beta_h^i$ and $g(\zhiweihu{y}_i)$ denote the \textit{i}-th element of $\beta_h$ and $g(\textbf{y})$, $\lambda_1$ and $\lambda_2$ are two scaling hyper-parameters. $\alpha_r$ and $\beta_r$ are defined similarly.

\vic{We then map the embedding region of the head entity set $\mathcal{C}_h$ to the  entity set $\mathcal{C}_{h,r}$ through a relation-dependent projection operation $\mathcal{P}$: $\mathcal{C}_{h,r}\zhiweihu{=\mathcal{C}(\alpha_{h,r}, \beta_{h,r})}=\mathcal{P}(\mathcal{C}_h, \mathcal{C}_r)$.}
% \nb{V: There i something weird there are two definitions of $\mathcal{C}_{h,r}$ }
%where \texttt{MLP}: $\mathbb{R}^d \rightarrow \mathbb{R}^d$ is a multi-layer perceptron network, function $f$ and $g$ are used to scale the semantic center axis and aperture into their normal ranges, $\alpha_h=f(\textbf{x})=[\alpha_h^1,...,\alpha_h^d; \alpha_h^i=f(x_i)]$, $\alpha_h^i$ and $f(x_i)$ denote the \textit{t}-th element of $\alpha_h$ and $f(\textbf{x})$, $\beta_h=g(\textbf{y})=[\beta_h^1,...,\beta_h^d; \beta_h^i=g(y_i)]$, $\beta_h^i$ and $g(x_i)$ denote the \textit{i}-th element of $\beta_h$ and $g(\textbf{y})$, $\lambda_1$ and $\lambda_2$ are two scaling hyper-parameters. Then, we map the embedding region of a head entity set $\mathcal{C}_h$ to another entity set $\mathcal{C}_{h,r}=\mathcal{C}(\alpha_{h,r}, \beta_{h,r})$ via a relation-dependent projection operation $\mathcal{P}$, denoted as $\mathcal{C}_{h,r}=\mathcal{P}(\mathcal{C}_h, \mathcal{C}_r)$.
\item As discussed, the monotonocity property can be intuitively captured by  reducing the axis and aperture of main triple cones to which qualifiers are added. More precisely, to geometrically represent this property in the cone embedding space, we model each qualifier ($a_i$ : $v_i$) as a shrinking of a cone. The spatial representation is such that the cone obtained after shrinking is a sub-cone of the original one. For example, in Figure~\ref{figure_shrink}(d), the shrink cone $\mathcal{C}_2$ is a subset of the original cone  $\mathcal{C}_1$.  Given a cone $\mathcal{C}(\alpha_{h,r}, \beta_{h,r})$ obtained in the previous step and the  token embedding ($\textbf{\textit{E}}^{a_i}$, $\textbf{\textit{E}}^{v_i}$) of the \textit{i}-th qualifier pair,  we formally define a cone-to-cone shrink transformation from $\mathcal{C}(\alpha_{h,r}, \beta_{h,r})$ to $\mathcal{C}(\alpha_{\mathcal{S}_i}, \beta_{\mathcal{S}_i})$ using Equation~\eqref{equation_3} below. Intuitively, Equation~\eqref{equation_3} adjusts the source angle of axis $\alpha_{h,r}$ and the aperture $\beta_{h,r}$ to shrink them to $\alpha_{\mathcal{S}_i}$ and $\beta_{\mathcal{S}_i}$, respectively, ensuring that the resulting cone is contained in the original cone.
%Attaching any qualifier pairs to a certain query will not enlarge the answer space, which can be intuitively reflected by a reduced axis and aperture of cones incurred by adding qualifiers. To geometrically mimic this property in the cone embedding space, we model each qualifier ($a_i$, $v_i$) as a shrinking of the cone, and the spatial representation is that the cone after shrinking is a sub-cone before shrinking, as shown in Figure~\ref{figure_shrink}(d), the shrink cone $\mathcal{C}_2$ is a subset of the cone embedding $\mathcal{C}_1$ before the shrink operation. Given a cone $\mathcal{C}(\alpha_{h,r}, \beta_{h,r})$ from the (1) step and the \textit{i}-th qualifier pair token embedding ($\textbf{\textit{E}}^{a_i}$, $\textbf{\textit{E}}^{v_i}$), we formally define a cone-to-cone shrink transformation from $\mathcal{C}(\alpha_{h,r}, \beta_{h,r})$ to $\mathcal{C}(\alpha_{\mathcal{S}_i}, \beta_{\mathcal{S}_i})$ as shown in Equation~\ref{equation_3}, which adjust the source angle of axis $\alpha_{h,r}$ and aperture $\beta_{h,r}$ to shrink axis $\alpha_{\mathcal{S}_i}$ and aperture $\beta_{\mathcal{S}_i}$, to ensure the resulting cone inside of the source cone.
\begin{equation}
\label{equation_3}
\left\{
	\begin{array}{l}
	\alpha_{r, a_i, v_i}=f(\texttt{MLP}_1(\Theta_{r, a_i, v_i})),\,\,\beta_{r, a_i, v_i}=g(\texttt{MLP}_1(\Theta_{r, a_i, v_i}))\vspace{1.2ex}\\
    
    {\rm offset}=(\beta_{h, r}-\beta_{\mathcal{S}_i})\odot\sigma(\alpha_{r, a_i, v_i}),\,\,\beta_{\mathcal{S}_i}=\sigma(\beta_{r, a_i, v_i})\odot\beta_{h,r}\vspace{1.2ex}\\
    \alpha_{\mathcal{S}_i}=\alpha_{h,r}-0.5\odot\beta_{h,r}+0.5\odot\beta_{\mathcal{S}_i}+0.5\odot{\rm offset}\\
    
	\end{array} \right.
\end{equation}
where $\Theta_{r, a_i, v_i}=\texttt{MLP}_2(\texttt{Concat}(\textbf{\textit{E}}^{r}, \textbf{\textit{E}}^{a_i}, \textbf{\textit{E}}^{v_i}))$, which leverages an \texttt{MLP} layer which takes the primary relation and attribute-value qualifier as input, and outputs the combined representation of relational and hyper-relational knowledge.  $\texttt{MLP}_1$: $\mathbb{R}^d \rightarrow \mathbb{R}^d$ and $\texttt{MLP}_2$: $\mathbb{R}^{3d} \rightarrow \mathbb{R}^d$ are multi-layer perceptron networks, $\sigma$ is a \textit{sigmoid} function and $\odot$ is the element-wise vector multiplication. From Equation (\ref{equation_3}), we  obtain a  sector-cone $\mathcal{C}_{\mathcal{S}_i}=\mathcal{C}(\alpha_{\mathcal{S}_i}, \beta_{\mathcal{S}_i})$ that has been shrinked by the qualifier  ($a_i$ : $v_i$).
%where $\Theta_{r, a_i, v_i}=\texttt{MLP}_2(\texttt{Concat}(\textbf{\textit{E}}^{r}, \textbf{\textit{E}}^{a_i}, \textbf{\textit{E}}^{v_i}))$, which leverages an \texttt{MLP} layer that takes the primary relation and attribute-value qualifier as input and outputs the combined representation of relational and hyper-relational knowledge,  $\texttt{MLP}_1$: $\mathbb{R}^d \rightarrow \mathbb{R}^d$ and $\texttt{MLP}_2$: $\mathbb{R}^{3d} \rightarrow \mathbb{R}^d$ are multi-layer perceptron networks, $\sigma$ is a \textit{sigmoid} function and $\odot$ is element-wise vector multiplication. After the operation of Equation~\ref{equation_3}, we can obtain a new sector-cone $\mathcal{C}_{\mathcal{S}_i}=\mathcal{C}(\alpha_{\mathcal{S}_i}, \beta_{\mathcal{S}_i})$ that has been shrinked by the qualifier pair ($a_i$, $v_i$).
\item After obtaining a scaled down sector-cone $\mathcal{C}_{\mathcal{S}_i}$, for each $i$-th qualifier, we obtain $s$   independent cones: \{$\mathcal{C}_{\mathcal{S}_1},...,\mathcal{C}_{\mathcal{S}_s}$\}. Finally,  when considering multiple qualifier pairs at the same time, the answer set can be computed by the cone intersection operation $\mathcal{I}$: $\mathcal{C}_{\mathcal{S}}=(\alpha_{\mathcal{S}}, \beta_{\mathcal{S}})=\mathcal{I}(\mathcal{C}_{\mathcal{S}_1},...,\mathcal{C}_{\mathcal{S}_s})$. Note that when $C_S$ is empty, the intersection of all $C_{S_i}$ must form a cone with an aperture of 0. However,  this scenario is highly improbable. The key reason is that each cone exists in a high-dimensional space (with a dimensionality of 200 for our \texttt{HyperMono}), making it virtually impossible for all $\beta$ values in $C_S$ to be 0. Even in extreme cases where the aperture is 0, he cone embedding remains effective in handling this situation, as discussed in ConE~\citep{Zhanqiu_2021}.

% 

%For the hyper-relational fact (\textit{h}, \textit{r}, \texttt{?}, $\{(a_1, v_1),..., (a_s, v_s)\}$) with \textit{s} qualifier pairs, the \textit{i}-th qualifier pair can obtain a scaled sector-cone $\mathcal{C}_{\mathcal{S}_i}$ after the above two steps, then we will be able to obtain \textit{s} mutually independent cones, denoted as \{$\mathcal{C}_{\mathcal{S}_1},...,\mathcal{C}_{\mathcal{S}_s}$\}, each element $\mathcal{C}_{\mathcal{S}_i}$ actually shrink the answer space of \texttt{?}, when considering multiple qualifier pairs at the same time, the answer set can be completed by the cone interaction operation $\mathcal{I}$, i.e., $\mathcal{C}_{\mathcal{S}}=(\alpha_{\mathcal{S}}, \beta_{\mathcal{S}})=\mathcal{I}(\mathcal{C}_{\mathcal{S}_1},...,\mathcal{C}_{\mathcal{S}_s})$.
\end{enumerate}

\smallskip
\noindent \textit{Inference Results.} We apply $\textit{q}^{t}=f(\texttt{MLP}(\texttt{Concat}(\alpha_{\mathcal{S}}, \beta_{\mathcal{S}})),\textbf{\textit{V}}_{ent}^{\top})$, where $f(\cdot)=\mathcal{D}(\textbf{\textit{V}}_{ent}, \mathcal{C}_{\mathcal{S}})$ denotes the score function to calculate the distance between $\textbf{\textit{V}}_{ent}$ and the scaled cone $\mathcal{C}_{\mathcal{S}}$, the distance function is consistent with ConE~\citep{Zhanqiu_2021}, $\textbf{\textit{q}}^{t}=\theta(\textit{q}^{t})$ to change the answer cone $\mathcal{C}_{\mathcal{S}}$ to the entity probability at the \texttt{?} position, $\theta$ is the \textit{softmax} function. Additionally, we calculate the cross-entropy loss between the prediction score $\textbf{\textit{q}}^{t}$ and the label $\textbf{\textit{L}}^t_2$ as $\mathcal{L}_{hyper}(t)=\texttt{CrossEntropy}(\textbf{\textit{q}}^t, \textbf{\textit{L}}^t_2)$.

%We apply the $\textbf{\textit{q}}^{t}=\theta(\texttt{MLP}(\texttt{Concat}(\alpha_{\mathcal{S}}, \beta_{\mathcal{S}}))\cdot\textbf{\textit{V}}_{ent}^{\top})$ to change the answer cone $\mathcal{C}_{\mathcal{S}}$ to the entity probability at the tail placeholder \texttt{?} position, further we calculate the cross-entropy loss between the prediction score $\textbf{\textit{q}}^{t}$ and the label $\textbf{\textit{L}}^t_2$ as $\mathcal{L}_{hyper}(t)=\texttt{CrossEntropy}(\textbf{\textit{q}}^t, \textbf{\textit{L}}^t_2)$.

\begin{table*}[!htp]
\setlength{\abovecaptionskip}{0.05cm}
\renewcommand\arraystretch{1.18}
\setlength{\tabcolsep}{0.23em}
\centering
\small
\caption{Evaluation of different models under the mixed-percentage mixed-qualifier scenario on the WD50K, WikiPeople and JF17K datasets. All baseline results are collected from the original papers. Best scores are highlighted in \colorbox{mycolor2}{\textbf{bold}}, the second best scores are highlighted in \colorbox{mycolor1}{normal}, {the $\Delta$ represents the performance improvement over the state-of-the-art modes,} and ’--’ indicates the results are not reported in previous work. The number in parentheses represents the proportion of triples with hyper-relational knowledge.}
\begin{tabular*}{\linewidth}{@{}c|cccccccccccccccccc@{}}
\hline
\multicolumn{1}{c|}{\multirow{3}{*}{\textbf{Methods}}} & \multicolumn{6}{c|}{\textbf{WD50K (13.6)}} & \multicolumn{6}{c|}{\textbf{WikiPeople (2.6)}} & \multicolumn{6}{c}{\textbf{JF17K (45.9)}}\\
\cline{2-7}\cline{8-13}\cline{14-19}
&\multicolumn{3}{c|}{\textbf{Head / Tail}} &\multicolumn{3}{c|}{\textbf{All Entities}} &\multicolumn{3}{c|}{\textbf{Head / Tail}} &\multicolumn{3}{c|}{\textbf{All Entities}} &\multicolumn{3}{c|}{\textbf{Head / Tail}} &\multicolumn{3}{c}{\textbf{All Entities}}\\
\cline{2-7}\cline{8-13}\cline{14-19}
& \textbf{MRR} & \textbf{H@1} & \multicolumn{1}{c|}{\textbf{H@10}} & \textbf{MRR} & \textbf{H@1} & \multicolumn{1}{c|}{\textbf{H@10}} & \textbf{MRR}  & \textbf{H@1} & \multicolumn{1}{c|}{\textbf{H@10}} & \textbf{MRR} & \textbf{H@1} & \multicolumn{1}{c|}{\textbf{H@10}} & \textbf{MRR} & \textbf{H@1} & \multicolumn{1}{c|}{\textbf{H@10}} & \textbf{MRR}  & \textbf{H@1} & \textbf{H@10}\\
\hline
m-TransH~\cite{Jianfeng_2016}   
&-- &-- &\multicolumn{1}{c|}{--} &--   &--   &\multicolumn{1}{c|}{--}   &0.063  &0.063  &\multicolumn{1}{c|}{0.300}  &--  &--  &\multicolumn{1}{c|}{--}  &0.206  &0.206    &\multicolumn{1}{c|}{0.462} &0.102  &0.069  &0.168 \\
RAE~\cite{Richong_2018} 
&--  &--  &\multicolumn{1}{c|}{--}  &--  &--  &\multicolumn{1}{c|}{--}  &0.058  &0.058  &\multicolumn{1}{c|}{0.306} &0.172  &0.102  &\multicolumn{1}{c|}{0.320}  &0.215  &0.215   &\multicolumn{1}{c|}{0.466} &0.310  &0.219  &0.504\\
NaLP-Fix~\cite{Paolo_2020}   
&0.177  &0.131  &\multicolumn{1}{c|}{0.264}  &0.224  &0.158  &\multicolumn{1}{c|}{0.330}  &0.408  &0.331  &\multicolumn{1}{c|}{0.546}  &0.338  &0.272  &\multicolumn{1}{c|}{0.466}  &0.221  &0.165 &\multicolumn{1}{c|}{0.331} &0.366  &0.290  &0.516  \\
NeuInfer~\cite{Saiping_2020}   
&0.243  &0.176  &\multicolumn{1}{c|}{0.377}  &0.228  &0.162  &\multicolumn{1}{c|}{0.341}  &0.476  &0.415  &\multicolumn{1}{c|}{0.585}  &0.333  &0.259  &\multicolumn{1}{c|}{0.477}  &0.449  &0.361 &\multicolumn{1}{c|}{0.624} &0.473  &0.397  &0.618  \\
HINGE~\cite{Paolo_2020}    
&0.243  &0.176  &\multicolumn{1}{c|}{0.377}  &0.232  &0.164  &\multicolumn{1}{c|}{0.343}  &0.342  &0.272  &\multicolumn{1}{c|}{0.463}  &0.350  &0.282  &\multicolumn{1}{c|}{0.467}  &0.431  &0.342    &\multicolumn{1}{c|}{0.611} &0.517  &0.436  &0.675  \\
ShrinkE~\cite{Bo_2023}    
&0.345 &0.275  &\multicolumn{1}{c|}{0.482}  &--  &--  &\multicolumn{1}{c|}{--}  &0.485  &0.431  &\multicolumn{1}{c|}{0.601}  &--  &--  &\multicolumn{1}{c|}{--}  &0.589  &0.506    &\multicolumn{1}{c|}{0.749} &--  &--  &--  \\
\hline
StarE~\cite{Mikhail_2020}    
&0.349  &0.271  &\multicolumn{1}{c|}{0.496}  &--  &--  &\multicolumn{1}{c|}{--}  &0.491  &0.398  &\multicolumn{1}{c|}{0.592}  &0.378  &0.265  &\multicolumn{1}{c|}{0.542}  &0.574  &0.496    &\multicolumn{1}{c|}{0.725} &0.542  &0.454  &0.685  \\
QUAD~\cite{Harry_2022}    
&0.349  &0.275  &\multicolumn{1}{c|}{0.489}  &--  &--  &\multicolumn{1}{c|}{--}  &0.497  &0.431  &\multicolumn{1}{c|}{0.617}  &--  &--  &\multicolumn{1}{c|}{--}  &0.596  &0.519    &\multicolumn{1}{c|}{0.751} &--  &--  &--  \\
\hline
Hy-Transformer~\cite{Donghan_2021}  
&0.356  &0.281  &\multicolumn{1}{c|}{0.498}  &--  &--  &\multicolumn{1}{c|}{--}  &0.501  &0.426  &\multicolumn{1}{c|}{0.634}  &--  &--  &\multicolumn{1}{c|}{--}  &0.582  &0.501    &\multicolumn{1}{c|}{0.742} &--  &--  &--  \\
GRAN~\cite{Quan_2021}   
&--  &--  &\multicolumn{1}{c|}{--}  &0.309  &0.240  &\multicolumn{1}{c|}{0.441}  &\cellcolor{mycolor1}0.503  &\cellcolor{mycolor1}0.438  &\multicolumn{1}{c|}{0.620}  &0.479  &0.410  &\multicolumn{1}{c|}{0.604}  &0.617  &0.539    &\multicolumn{1}{c|}{0.770} &0.656  &0.582  &0.799  \\
HyNT~\cite{Chanyoung_2023}    
&0.314 &0.241  &\multicolumn{1}{c|}{0.454}  &--  &--  &\multicolumn{1}{c|}{--}  &0.464  &0.380  &\multicolumn{1}{c|}{0.601}  &--  &--  &\multicolumn{1}{c|}{--}  &0.554  &0.475    &\multicolumn{1}{c|}{0.711} &--  &--  &--  \\
HAHE~\cite{Haoran_2023}    
&\cellcolor{mycolor1}0.368 &\cellcolor{mycolor1}0.291  &\multicolumn{1}{c|}{\cellcolor{mycolor1}0.516}  &0.402  &0.327  &\multicolumn{1}{c|}{0.546}  &\cellcolor{mycolor2}\textbf{0.509}  &\cellcolor{mycolor2}\textbf{0.447}  &\multicolumn{1}{c|}{0.639}  &0.495  &\cellcolor{mycolor1}0.420  &\multicolumn{1}{c|}{0.631}  &0.623  &0.554    &\multicolumn{1}{c|}{\cellcolor{mycolor2}\textbf{0.806}} &0.668  &0.597  &0.816  \\
HyperFormer~\cite{Zhiwei_2023}    
&0.366 &0.288  &\multicolumn{1}{c|}{0.514}  &\cellcolor{mycolor1}0.417  &\cellcolor{mycolor1}0.344  &\multicolumn{1}{c|}{\cellcolor{mycolor1}0.555}  &0.491  &0.386  &\multicolumn{1}{c|}{\cellcolor{mycolor1}0.654}  &0.479  &0.373  &\multicolumn{1}{c|}{\cellcolor{mycolor1}0.644}  &\cellcolor{mycolor1}0.664  &\cellcolor{mycolor1}0.601    &\multicolumn{1}{c|}{0.787} &0.662  &0.596  &0.790  \\
HyperCL~\cite{Yuhuan_2024}    
&0.302 &0.231  &\multicolumn{1}{c|}{0.441}  &0.395  &0.321  &\multicolumn{1}{c|}{0.539}  &0.455  &0.365  &\multicolumn{1}{c|}{0.596}  &\cellcolor{mycolor2}\textbf{0.509}  &\cellcolor{mycolor2}\textbf{0.437}  &\multicolumn{1}{c|}{\cellcolor{mycolor1}0.644}  &0.513  &0.429    &\multicolumn{1}{c|}{0.684} &\cellcolor{mycolor1}0.673  &\cellcolor{mycolor1}0.604  &\cellcolor{mycolor1}0.818  \\
\hline
\zhiweihu{\texttt{HyperMono}}    
&\cellcolor{mycolor2}\textbf{0.375} &\cellcolor{mycolor2}\textbf{0.298}  &\multicolumn{1}{c|}{\cellcolor{mycolor2}\textbf{0.522}}  &\cellcolor{mycolor2}\textbf{0.447}  &\cellcolor{mycolor2}\textbf{0.376}  &\multicolumn{1}{c|}{\cellcolor{mycolor2}\textbf{0.582}}  &0.494  &0.390  &\multicolumn{1}{c|}{\cellcolor{mycolor2}\textbf{0.657}}  &\cellcolor{mycolor1}0.498  &0.394  &\multicolumn{1}{c|}{\cellcolor{mycolor2}\textbf{0.659}}  &\cellcolor{mycolor2}\textbf{0.685}  &\cellcolor{mycolor2}\textbf{0.630}    &\multicolumn{1}{c|}{\cellcolor{mycolor1}0.796} &\cellcolor{mycolor2}\textbf{0.766}  &\cellcolor{mycolor2}\textbf{0.717}  &\cellcolor{mycolor2}\textbf{0.861}  \\
\hline
\multicolumn{1}{c|}{$\Delta$ (\%)}   &\textcolor{mygreen}{+0.7}  &\textcolor{mygreen}{+0.7}&\multicolumn{1}{c|}{\textcolor{mygreen}{+0.6}} &\textcolor{mygreen}{+3.0} &\textcolor{mygreen}{+3.2}  &\multicolumn{1}{c|}{\textcolor{mygreen}{+2.7}} &\textcolor{myred}{-1.5} &\textcolor{myred}{-5.7} &\multicolumn{1}{c|}{\textcolor{mygreen}{+0.3}}  &\textcolor{myred}{-1.1} &\textcolor{myred}{-4.3} &\multicolumn{1}{c|}{\textcolor{mygreen}{+1.5}}   &\textcolor{mygreen}{+2.1} &\textcolor{mygreen}{+2.9} &\multicolumn{1}{c|}{\textcolor{myred}{-1.0}} &\textcolor{mygreen}{+9.3}  &\textcolor{mygreen}{+11.3}&\textcolor{mygreen}{+4.3}\\
\hline
\end{tabular*}
\label{table_mixed_percentage_mixed_qualifier}
\end{table*}

\subsubsection{Joint Training} \vic{We combine the loss functions obtained above into a joint loss function:}
% From  the \texttt{CNA} and \texttt{FNA} components of the \texttt{HNE} module, we  obtain the losses $\mathcal{L}_{\textit{triple}}(h)$ and $\mathcal{L}_{\textit{hyper}}(h)$, from the predictors \texttt{TDP} and \texttt{QMP} of the \texttt{MEP} module, we obtain the losses $\mathcal{L}_{\emph{triple}}(t)$ and $\mathcal{L}_{\emph{hyper}}(t)$. We further combine these losses as the joint loss function:
\begin{equation}
\label{equation_4}
\mathcal{L}_{\textit{joint}}=\mathcal{L}_{\textit{triple}}(h) + \mathcal{L}_{\textit{hyper}}(h) + \mathcal{L}_{\textit{triple}}(t) + \mathcal{L}_{\textit{hyper}}(t)
\end{equation}

%% file: sections/experiment.tex
\vic{To evaluate the effectiveness of our model, we aim to explore the following five research questions:}
%To evaluate the effectiveness of, we aim to explore the following research questions:
\begin{itemize}[itemsep=0.5ex, leftmargin=5mm]
\item 
\textbf{RQ1 (Effectiveness):} \vic{How our model performs compared to the state-of-the-art under different conditions?} We also present the Case study in \textcolor{blue}{\textbf{\hyperlink{case_study}{Appendix H}}}.
\item 
\textbf{RQ2 (Ablation studies):} \vic{How  different components  contribute to its performance?}
\item 
\textbf{RQ3 (Parameter sensitivity):} \vic{How  hyper-parameters influence its performance?} Additional results can be
found in \textcolor{blue}{\textbf{\hyperlink{parameter_sensitivity}{Appendix D}}} and \textcolor{blue}{\textbf{\hyperlink{additional_results}{Appendix F}}}.
\item
\textbf{RQ4 (Complexity analysis):} 
What is the amount of computation and parameters used by \texttt{HyperMono}? Results can be found in \textcolor{blue}{\textbf{\hyperlink{complexity_analysis}{Appendix E}}}.
\item 
\textbf{RQ5 (Model transferability):} Can \texttt{HyperMono} be used to complete \textit{hypergraphs}? Results can be found in \textcolor{blue}{\textbf{\hyperlink{model_transferability}{Appendix G}}}.
% \item 
% \zhiweihu{\textbf{RQ4 (Complexity analysis):} 
% What is the amount of computation and parameters used by \texttt{HyperMono}?}
\end{itemize}

\subsection{Experiment Setup}\label{sec:expresults}
\subsubsection{Datasets}
\label{section_datasets}
We conduct experiments on three hyper-relational KGs: WD50K~\citep{Mikhail_2020}, WikiPeople~\citep{Saiping_2019}, and JF17K~\citep{Jianfeng_2016}, where WD50K and WikiPeople have been extracted from Wikidata~\citep{Denny_2014} and JF17K from Freebase~\citep{Kurt_2008}. We also construct three different sub-datasets, namely: {\textit{Mixed-percentage Mixed-qualifier}, \textit{Fixed-percentage Fixed-qualifier}, and \textit{Fixed-percentage Mixed-qualifier}}. Statistics and details on the construction can be found in \textcolor{blue}{\textbf{\hyperlink{datasets}{Appendix A}}}.

\subsubsection{Baselines.} We compare \texttt{HyperMono} with {fourteen} state-of-the-art baselines for HKGC, including \textit{Embedding-based Methods:} m-TransH~\citep{Jianfeng_2016}, RAE~\citep{Richong_2018}, NaLP-Fix~\citep{Paolo_2020}, {NeuInfer~\cite{Saiping_2020}}, HINGE~\citep{Paolo_2020}, and ShrinkE~\citep{Bo_2023}; \textit{Transformer-based with GNNs based Methods:}  StarE~\citep{Mikhail_2020} and QUAD~\citep{Harry_2022}; \textit{Transformer-based without GNNs based Methods:} Hy-Transformer~\citep{Donghan_2021}, GRAN~\citep{Quan_2021}, HyNT~\citep{Chanyoung_2023}, HAHE~\citep{Haoran_2023}, HyperFormer~\citep{Zhiwei_2023}, and {HyperCL~\cite{Yuhuan_2024}}. We note that although HELIOS~\citep{Yuhuan_2023_1}, sHINGE~\citep{Yuhuan_2023} and NYLON~\citep{Weijian_2024} also tackle the HKGC task, there are some variations in terms of experimental design and the selection of evaluation metrics, e.g., HELIOS~\citep{Yuhuan_2023_1} selects MAP and NDCG as evaluation metrics, sHINGE~\citep{Yuhuan_2023} heavily relies on additional schema knowledge and NYLON~\citep{Weijian_2024} employs data augmentation techniques. Thus, we do not consider them as baselines. Detailed baseline descriptions, evaluation metrics and implementation details can respectively be found in \textcolor{blue}{\textbf{\hyperlink{baselines}{Appendix B}}} and \textcolor{blue}{\textbf{\hyperlink{protocol_and_implements}{Appendix C}}}.

\begin{table*}[!htp]
\setlength{\abovecaptionskip}{0.05cm}
\renewcommand\arraystretch{1.05}
\setlength{\tabcolsep}{0.42em}
\centering
\small
\caption{Evaluation of different models under the fixed-percentage fixed-qualifier scenario on the WikiPeople and JF17K datasets. Best scores are highlighted in \colorbox{mycolor2}{\textbf{bold}}, the second best scores are highlighted in \colorbox{mycolor1}{normal}.}
\begin{tabular*}{\linewidth}{@{}ccccccccccccccccc@{}}
\hline
\multicolumn{1}{c}{\multirow{2}{*}{\textbf{Methods}}}   & \multicolumn{4}{c}{\textbf{WikiPeople-3}} & \multicolumn{4}{c}{\textbf{WikiPeople-4}} & \multicolumn{4}{c}{\textbf{JF17K-3}} & \multicolumn{4}{c}{\textbf{JF17K-4}}\\
\cline{2-5}\cline{6-9}\cline{10-13}\cline{14-17}

& \textbf{MRR} & \textbf{H@1}   & \textbf{H@3} & \textbf{H@10} & \textbf{MRR} & \textbf{H@1} & \textbf{H@3} & \textbf{H@10} & \textbf{MRR}  & \textbf{H@1} & \textbf{H@3}   & \textbf{H@10} & \textbf{MRR} & \textbf{H@1} & \textbf{H@3}    & \textbf{H@10}\\
% \hline
% \multicolumn{17}{c}{\textit{Embedding-based Methods}} \\
\hline

ShrinkE~\cite{Bo_2023}    &0.366  &0.291   &0.397  &0.515  &0.222  &0.155  &0.252  &0.353  &0.712  &0.648  &0.745    &0.837  &0.779  &0.726    &0.809  &0.886 \\
% \hline
% \multicolumn{17}{c}{\textit{Transformer-based with GNNs Methods}} \\
\hline
StarE~\cite{Mikhail_2020}    &0.401  &0.310   &0.434  &0.592  &0.243  &0.156  &0.269  &0.430  &0.707  &0.635  &0.744    &0.847  &0.723  &0.669    &0.753  &0.839 \\
QUAD~\cite{Quan_2021}    &0.403  &0.321   &0.438  &0.563  &0.251  &0.167  &0.280  &0.425  &0.730  &0.660  &0.767    &0.870  &0.787  &0.730    &0.823  &0.895 \\
% \hline
% \multicolumn{17}{c}{\textit{Transformer-based without GNNs Methods}} \\
\hline
Hy-Transformer~\cite{Donghan_2021}  &0.403  &0.323   &0.436  &0.569  &0.248  &0.165  &0.275  &0.422  &0.690  &0.617  &0.725    &0.837  &0.773  &0.717    &0.806  &0.875 \\
GRAN~\cite{Quan_2021}    &0.397  &0.328   &0.429  &0.533  &0.239  &0.178  &0.261  &0.364  &0.779  &0.724  &0.811    &0.893  &0.798  &0.744    &0.830  &0.904 \\
HyNT~\cite{Chanyoung_2023}    &0.383  &0.306   &0.415  &0.533  &0.233  &0.166  &0.257  &0.369  &0.733  &0.664  &0.771    &0.873  &0.779  &0.725    &0.809  &0.881 \\
HAHE~\cite{Haoran_2023}    &0.397  &0.323   &0.432  &0.537  &0.251  &0.179  &0.286  &0.392  &0.762  &0.701  &0.791    &0.892  &0.736  &0.666    &0.773  &0.869 \\
HyperFormer~\cite{Zhiwei_2023}   &\cellcolor{mycolor1}0.573  &\cellcolor{mycolor1}0.511  &\cellcolor{mycolor1}0.603  &\cellcolor{mycolor2}\textbf{0.693}  &\cellcolor{mycolor1}0.393 &\cellcolor{mycolor1}0.336  &\cellcolor{mycolor2}\textbf{0.415} &\cellcolor{mycolor2}\textbf{0.496}  &\cellcolor{mycolor1}0.832  &\cellcolor{mycolor1}0.790   &\cellcolor{mycolor1}0.855    &\cellcolor{mycolor1}0.914  &\cellcolor{mycolor1}0.857  &\cellcolor{mycolor1}0.811  &\cellcolor{mycolor1}0.884   &\cellcolor{mycolor1}0.937 \\
HyperCL~\cite{Yuhuan_2024}     &0.397  &0.327   &0.425  &0.538  &0.247  &0.175  &0.279  &0.387  &0.766  &0.707  &0.796    &0.890  &0.733  &0.654    &0.779  &0.874 \\
\hline
\zhiweihu{\texttt{HyperMono}}   &\cellcolor{mycolor2}\textbf{0.586}  &\cellcolor{mycolor2}\textbf{0.531}  &\cellcolor{mycolor2}\textbf{0.611}  &\cellcolor{mycolor1}0.690  &\cellcolor{mycolor2}\textbf{0.398} &\cellcolor{mycolor2}\textbf{0.359}  &\cellcolor{mycolor1}0.408 &\cellcolor{mycolor1}0.474  &\cellcolor{mycolor2}\textbf{0.867}  &\cellcolor{mycolor2}\textbf{0.839}   &\cellcolor{mycolor2}\textbf{0.880}    &\cellcolor{mycolor2}\textbf{0.927}  &\cellcolor{mycolor2}\textbf{0.881}  &\cellcolor{mycolor2}\textbf{0.847}  &\cellcolor{mycolor2}\textbf{0.901}   &\cellcolor{mycolor2}\textbf{0.945} \\

\hline
\end{tabular*}
\label{table_fixed_percentage_fixed_qualifier}
\end{table*}

\subsection{Main Results}
\label{main_results}
\vic{To address \textbf{RQ1}, we conduct experiments on the three types of datasets described above. The corresponding results are shown in Table~\ref{table_mixed_percentage_mixed_qualifier}, Table~\ref{table_fixed_percentage_fixed_qualifier} and Table~\ref{table_fixed_percentage_mixed_qualifier}.} {As shown in Table~\ref{table_mixed_percentage_mixed_qualifier}, the improvement in predicting head and tail entities by \texttt{HyperMono} is less pronounced compared to the improvement  on all entities. Consequently, Tables~\ref{table_fixed_percentage_fixed_qualifier} and ~\ref{table_fixed_percentage_mixed_qualifier} focus more on the prediction of head and tail entities, which pose greater challenges for \texttt{HyperMono}.}

\smallskip
\noindent\textbf{Mixed-percentage Mixed-qualifier.} 
{Table~\ref{table_mixed_percentage_mixed_qualifier} summarizes the results on WD50K, WikiPeople, and JF17K under the mixed-percentage mixed-qualifier conditions. We observe that \texttt{HyperMono} outperforms existing SoTA baselines by a large margin across most metrics on WD50K and JF17K in the  head/tail entities and all entities scenarios. In particular, in the prediction of head and tail entities,  \texttt{HyperMono} respectively achieves 2.1\% and 2.9\% improvements over HyperFormer (the best baseline) in MRR and Hits@1 on the JF17K dataset. A similar phenomenon occurred on the WD50K dataset. However, for the WikiPeople dataset, even though we can achieve optimal performance on the Hits@10 metric, we cannot surpass the best performing baseline on MRR and Hits@1 in the head/tail and all entities settings. The intuitive reason for this is that, in comparison to the WD50K and JF17K datasets where 13.6\% and 45.9\% of the main triples respectively contain hyper-relational knowledge, only 2.6\% of triples in WikiPeople have qualifier pairs. This hinders \texttt{HyperMono}'s ability to exploit stage reasoning and qualifier monotonicity. Specifically, we employ a scaling transformation in the cone space to capture qualifier monotonicity, but the lack of  qualifier pairs constrains this functionality. Furthermore, the proportion of triples with hyper-relational knowledge in JF17K is higher than that in WD50K. We can observe that the performance improvement in JF17K is more significant, which further confirms the importance of the richness of hyper-relational knowledge.}

\noindent\textbf{Fixed-percentage Fixed-qualifier.} To better understand the impact of hyper-knowledge on the performance, we consider the WikiPeople and JF17K datasets with a fixed number of qualifiers, cf.\ Table~\ref{table_fixed_percentage_fixed_qualifier}. We can observe that \zhiweihu{\texttt{HyperMono}} consistently achieves optimal performance on almost all datasets. Specifically, in terms of the MRR metric, \zhiweihu{\texttt{HyperMono}} respectively shows improvements of 3.5\% and 2.4\% on the JF17K-3 and JF17K-4 subsets when compared with HyperFormer. Furthermore, we find that for all models, their performance in the fixed-percentage fixed-qualifier setting is generally higher than in the fixed-percentage mixed-qualifier one. The main reason is that in the fixed-percentage fixed-qualifier scenario, the number of qualifier pairs for each triple is fixed, and there is no imbalance in the distribution of hyper-relational knowledge.

\begin{table*}[!htp]
\setlength{\abovecaptionskip}{0.05cm}
\renewcommand\arraystretch{1.05}
\setlength{\tabcolsep}{0.295em}
\centering
\small
\caption{Evaluation of different models under the fixed-percentage mixed-qualifier scenario on the WD50K, WikiPeople and JF17K datasets. Best scores are highlighted in \colorbox{mycolor2}{\textbf{bold}}, the second best scores are highlighted in \colorbox{mycolor1}{normal}.}
\begin{tabular*}{0.99\linewidth}{@{}ccccccccccccccccccc@{}}
\hline
\multicolumn{1}{c}{\multirow{3}{*}{\textbf{Methods}}} & \multicolumn{6}{c}{\textbf{WD50K}} & \multicolumn{6}{c}{\textbf{WikiPeople}} & \multicolumn{6}{c}{\textbf{JF17K}}\\
\cline{2-7}\cline{8-13}\cline{14-19}
& \multicolumn{2}{c}{\textbf{33\%}} & \multicolumn{2}{c}{\textbf{66\%}} & \multicolumn{2}{c}{\textbf{100\%}} & \multicolumn{2}{c}{\textbf{33\%}} & \multicolumn{2}{c}{\textbf{66\%}} & \multicolumn{2}{c}{\textbf{100\%}} & \multicolumn{2}{c}{\textbf{33\%}}  & \multicolumn{2}{c}{\textbf{66\%}} & \multicolumn{2}{c}{\textbf{100\%}} \\
\cline{2-3}\cline{4-5}\cline{6-7}\cline{8-9}\cline{10-11}\cline{12-13}\cline{14-15}\cline{16-17}\cline{18-19}

& \textbf{MRR} & \textbf{H@1} & \textbf{MRR} & \textbf{H@1} & \textbf{MRR}  & \textbf{H@1} & \textbf{MRR} & \textbf{H@1} & \textbf{MRR} & \textbf{H@1} & \textbf{MRR}  & \textbf{H@1} & \textbf{MRR} & \textbf{H@1} & \textbf{MRR} & \textbf{H@1} & \textbf{MRR}  & \textbf{H@1} \\

% \hline
% \multicolumn{19}{c}{\textit{Embedding-based Methods}} \\
\hline
ShrinkE~\cite{Bo_2023}  &0.246  &0.190  &0.356  &0.294  &0.470  &0.404    &0.176  &0.129 &0.244 &0.198 &0.332 &0.276   &0.259 &0.176   &0.263 &0.185   &0.299 &0.220 \\
% \hline
% \multicolumn{19}{c}{\textit{Transformer-based with GNNs Methods}} \\
\hline
StarE~\cite{Mikhail_2020}   &0.308  &0.247  &0.449  &0.388  &0.610  &0.543    &0.192  &0.143 &0.259 &0.205 &0.343 &0.279   &0.290 &0.197   &0.302 &0.214   &0.321 &0.223\\
QUAD~\cite{Quan_2021}  &0.329  &0.266  &0.479  &0.416  &0.646  &0.572    &0.204  &0.155 &0.282 &0.228 &0.385 &0.318   &0.307 &0.210   &0.334 &0.241   &0.379 &0.277\\
% \hline
% \multicolumn{19}{c}{\textit{Transformer-based without GNNs Methods}} \\
\hline
Hy-Transformer~\cite{Donghan_2021}   &0.313  &0.255  &0.458  &0.397  &0.621  &0.557    &0.192  &0.140 &0.268 &0.215 &0.372 &0.316   &0.298 &0.204   &0.325 &0.234   &0.361 &0.266\\
GRAN~\cite{Quan_2021}   &0.322  &\cellcolor{mycolor1}0.269  &0.472  &0.419  &0.647  &0.593    &0.201  &0.156 &0.287 &0.244 &0.403 &0.349   &0.307 &0.212   &0.326 &0.237   &0.382 &0.290\\
HyNT~\cite{Chanyoung_2023}  &0.295  &0.238  &0.438  &0.377  &0.595  &0.525    &0.175  &0.130 &0.243 &0.192 &0.329 &0.269   &0.285 &0.195   &0.298 &0.216   &0.352 &0.261 \\
HAHE~\cite{Haoran_2023}   &0.301  &0.239  &0.432  &0.368  &0.607  &0.537    &0.188  &0.142 &0.263 &0.213 &0.367 &0.304   &0.311 &0.221   &0.330 &0.248   &0.387 &0.293 \\
HyperFormer~\cite{Zhiwei_2023}   &\cellcolor{mycolor1}0.338  &\cellcolor{mycolor2}\textbf{0.280}  &\cellcolor{mycolor2}\textbf{0.492}  &\cellcolor{mycolor2}\textbf{0.434}  &\cellcolor{mycolor1}0.666  &\cellcolor{mycolor1}0.611    &\cellcolor{mycolor1}0.213  &\cellcolor{mycolor1}0.161 &\cellcolor{mycolor1}0.298 &\cellcolor{mycolor1}0.255 &\cellcolor{mycolor1}0.426 &\cellcolor{mycolor1}0.373   &\cellcolor{mycolor1}0.352 &\cellcolor{mycolor1}0.254   &\cellcolor{mycolor1}0.411 &\cellcolor{mycolor1}0.325   &\cellcolor{mycolor1}0.478 &\cellcolor{mycolor1}0.396 \\
HyperCL~\cite{Yuhuan_2024}    &0.263  &0.208  &0.388  &0.334  &0.632  &0.566    &0.155  &0.109 &0.221 &0.177 &0.363 &0.303   &0.284 &0.191   &0.300 &0.218   &0.386 &0.290 \\
\hline
\zhiweihu{\texttt{HyperMono}}   &\cellcolor{mycolor2}\textbf{0.341}  &\cellcolor{mycolor2}\textbf{0.280}  &\cellcolor{mycolor1}0.490  &\cellcolor{mycolor1}0.433  &\cellcolor{mycolor2}\textbf{0.667}  &\cellcolor{mycolor2}\textbf{0.612}    &\cellcolor{mycolor2}\textbf{0.221}  &\cellcolor{mycolor2}\textbf{0.173} &\cellcolor{mycolor2}\textbf{0.309} &\cellcolor{mycolor2}\textbf{0.264} &\cellcolor{mycolor2}\textbf{0.427} &\cellcolor{mycolor2}\textbf{0.379}   &\cellcolor{mycolor2}\textbf{0.357} &\cellcolor{mycolor2}\textbf{0.267}   &\cellcolor{mycolor2}\textbf{0.423} &\cellcolor{mycolor2}\textbf{0.340}   &\cellcolor{mycolor2}\textbf{0.493} &\cellcolor{mycolor2}\textbf{0.408} \\
\hline
\end{tabular*}
\label{table_fixed_percentage_mixed_qualifier}
\end{table*}

\noindent\textbf{Fixed-percentage Mixed-qualifier.} 
\vic{Table~\ref{table_fixed_percentage_mixed_qualifier} presents the results on WD50K, WikiPeople, and JF17K under different fixed ratios of relational facts with qualifiers. For each dataset, we construct three subsets (cf.\  \S~\ref{section_datasets}, Point 2) containing approximately 33\%, 66\%, and 100\% of facts with qualifiers. We observe that under different proportions of hyper-relational knowledge \texttt{HyperMono} also  achieves optimal performance. Particularly, for the WikiPeople dataset, in which \texttt{HyperMono} performs poorly in the mixed-percentage mixed-qualifier scenario, adding a certain percentage of qualifiers information to triples results in \texttt{HyperMono} outperforming  all baselines. For example, for the Hits@1 metric, \texttt{HyperMono} respectively achieves improvements over the best performing baseline, HyperFormer, of 1.2\% / 0.9\% / 0.6\% in the 33\% / 66\% / 100\% variants. This confirms that the proportion of hyper-relational knowledge is a major factor influencing the performance over WikiPeople under the mixed-percentage mixed-qualifier condition. In contrast, on the WD50K dataset, no significant additional improvement was obtained. The most likely reason is the insufficient richness of hyper-relational knowledge in WD50K. For example, in WD50K\_100    65.23\% of the triples  only contain \emph{one} hyper-relation pair, \zhiweihu{which limits the functionality of qualifier monotonicity module.}
}

\begin{table*}[!htp]
\setlength{\abovecaptionskip}{0.05cm}
\renewcommand\arraystretch{1.2}
\setlength{\tabcolsep}{0.48em}
\centering
\small
\caption{Ablation studies about different components. Best scores are highlighted in \textbf{bold}.}
\begin{tabular*}{\linewidth}{@{}ccccccccccccccccc@{}}
\hline
\multicolumn{1}{c}{\multirow{2}{*}{\textbf{Methods}}}   & \multicolumn{4}{c}{\textbf{WikiPeople-3}} & \multicolumn{4}{c}{\textbf{WikiPeople-4}} & \multicolumn{4}{c}{\textbf{JF17K-3}} & \multicolumn{4}{c}{\textbf{JF17K-4}}\\
\cline{2-5}\cline{6-9}\cline{10-13}\cline{14-17}

& \textbf{MRR} & \textbf{H@1}   & \textbf{H@3} & \textbf{H@10} & \textbf{MRR} & \textbf{H@1} & \textbf{H@3} & \textbf{H@10} & \textbf{MRR}  & \textbf{H@1} & \textbf{H@3}   & \textbf{H@10} & \textbf{MRR} & \textbf{H@1} & \textbf{H@3}    & \textbf{H@10}\\
\hline

\texttt{w/o CNA+TP}    &0.559  &0.493   &0.591  &0.686  &0.367  &0.303  &0.382  &\textbf{0.477}  &0.832  &0.792  &0.855    &0.912  &0.860  &0.820    &0.885  &0.940 \\
\texttt{w/o FNA+QMP}    &0.560  &0.510   &0.578  &0.663  &0.377  &0.340  &0.383  &0.445  &0.839  &0.811  &0.848    &0.900  &0.842  &0.809    &0.859  &0.909 \\
\texttt{w/o LEI}    &0.576  &0.523   &0.599  &0.678  &0.394  &0.359  &0.400  &0.465  &0.860  &0.834  &0.871    &0.918  &0.878  &0.843    &0.898  &0.946 \\
\texttt{w/o GEI}    &0.584  &0.527   &0.606  &0.687  &0.396  &0.355  &0.404  &0.471  &0.866  &0.838 &0.878    &0.926  &0.877  &0.842    &0.898  &0.941 \\
\texttt{w/o CSB}    &0.565  &0.502   &0.595  &0.687  &0.355  &0.296  &0.380  &0.464  &0.831  &0.789  &0.853    &0.915  &0.854  &0.815    &0.881  &0.933 \\
\hline
\zhiweihu{\texttt{HyperMono}}   &\textbf{0.586}  &\textbf{0.531}  &\textbf{0.611}  &\textbf{0.690}  &\textbf{0.398} &\textbf{0.359}  &\textbf{0.408} &0.474  &\textbf{0.867}  &\textbf{0.839}   &\textbf{0.880}    &\textbf{0.927}  &\textbf{0.881}  &\textbf{0.847}  &\textbf{0.901}   &\textbf{0.945} \\

\hline
\end{tabular*}
\label{table_ablation_studies}
\end{table*}

\subsection{Ablation Studies}
To address \textbf{RQ2}, we conduct ablation experiments on WikiPeople-3, WikiPeople-4, JF17K-3, and JF17K-4 subsets to verify the contribution of each component of \zhiweihu{\texttt{HyperMono}}. These include the following three aspects: a) whether the knowledge of qualifiers is considered to infer missing entities, see rows ``\texttt{w/o CNA+\zhiweihu{TP}}'' and ``\texttt{w/o FNA+QMP}'' in Table~\ref{table_ablation_studies}; b) different neighbor aggregation strategies, see rows ``\texttt{w/o LEI}'' and ``\texttt{w/o GEI}'' in Table~\ref{table_ablation_studies}; c) without the cone shrink operation in \texttt{QMP}, see row ``\texttt{w/o CSB}'' in Table~\ref{table_ablation_studies}.

\smallskip
\noindent\textbf{Different Variants of \zhiweihu{\texttt{HyperMono}}.} \vic{In Table~\ref{table_ablation_studies}, we observe that under  most metrics removing any functional module will bring some  performance degradation, which justifies the necessity of each component. Specifically, a) removing either \texttt{CNA+\zhiweihu{TP}}, which are related to coarse-grained answer prediction, or \texttt{FNA+QMP}, which are related to with fine-grained answer prediction, will result in a significant loss of accuracy. The intuitive reason is that the coarse-grained answer set can be used to constrain the decision space of fine-grained answers, preventing the occurrence of answer drift. Meanwhile, the fine-grained answer set allows to refine the answer set, ensuring the reliability of the predicted results. b) The neighborhood information of known entities has a positive effect on predicting missing entities, and the use of \texttt{LEI} for integrating the neighbors of known entities has a greater impact on inference than \texttt{GEI}. c) Removing the \texttt{CSB} module will bring  a significant decrease in performance. The main reason is that the \texttt{CSB} module captures the qualifier monotonicity of hyper-relational KGs. The main characteristic of the HKGC task is the application of qualifier knowledge in  triples  to infer answer sets. Through the natural scaling transformation of the cone space, it can be captured that attaching qualifiers to a  triple can only narrow down the answer set, but never enlarge it.
}

%% file: sections/conclusion.tex
% \vic{In this paper, we propose \zhiweihu{\texttt{HyperMono}}, a  framework for hyper-relational graph completion that properly implements \vi{two-stage reasoning} and  faithfully captures qualifier monotonicity. \zhiweihu{Specifically, we model each qualifier as a shrinking of the main triple cone to a qualifier cone to realize qualifier monotonicity. } Empirical experiments under various settings on the WD50K, WikiPeople, and JF17K datasets demonstrate  \zhiweihu{\texttt{HyperMono}'s}  robust performance, when compared to competitive baselines. Additionally, we present various ablation studies. For future work, we will look at leveraging types and numerical discrete values associated with entities. }

In this paper, we propose \texttt{HyperMono}, a  framework for hyper-relational graph completion that properly implements two-stage reasoning and  faithfully captures qualifier monotonicity. Specifically, we model each qualifier as a shrinking of the main triple cone to a qualifier cone to realize qualifier monotonicity. Empirical experiments under various settings and  datasets demonstrate \texttt{HyperMono}'s robust performance, when compared to competitive baselines. Additionally, we present various ablation studies on parameter sensitivity, complexity analysis and model transferability. For future work, we believe that it is necessary to pay special attention to numerical attribute knowledge in hyper-relational knowledge. How to elegantly integrate schema knowledge into the representation process is also an interesting line of research.

%In this paper, we propose RDQM, a novel framework for hyper-relational knowledge graph completion that respects reasoning diversity and qualifier monotonicity when utilizing qualifiers knowledge. Specifically, RDQM simultaneously considers the with and without qualifiers information to infer missing entities to ensure reasoning diversity, and explicitly encodes each qualifier as a shrinking cone to satisfy qualifier monotonicity. Empirical experiments under mixed-percentage mixed-qualifier, fixed-percentage mixed-qualifier, and fixed-percentage fixed-qualifier scenario conditions on the WD50K, WikiPeople, and JF17K datasets demonstrate that RDQM performs well as compared to competitive baselines. Further experiments validate the important of the various components and parameter sensitivity in our framework. For future work, on the one hand, we observe that most entities in the benchmark contain type knowledge. However, existing methods have not effectively leveraged coarse-grained type knowledge. On the other hand, given the relatively small proportion of discrete numerical values in the hyper-relational knowledge graph, we have not established a dedicated mechanism to handle this particular scenario. In the next step, we will consider hyper-relational knowledge graph completion only for entities with discrete numerical values.

%% file: sections/appendix.tex
\begin{table}[!htp]
\setlength{\abovecaptionskip}{0.03cm} 
\renewcommand\arraystretch{1.1}
\setlength{\tabcolsep}{0.60em}
\centering
\small
\caption{Statistics of datasets under \zhiweihu{three different scenarios}. The values in parentheses indicate  the percentage of triples  in  the corresponding dataset with hyper-relational facts. The hyphen -\textit{n} indicates that each main triple in the corresponding dataset contains \textit{n} qualifier pairs.}
\begin{tabular*}{0.95\linewidth}{@{}cccccc@{}}
\hline
\multicolumn{1}{c}{\textbf{Datasets}} &\multicolumn{1}{c}{\textbf{Train}} &\multicolumn{1}{c}{\textbf{Valid}} &\multicolumn{1}{c}{\textbf{Test}} &\multicolumn{1}{c}{\textbf{Entity}} &\multicolumn{1}{c}{\textbf{Relation}}\\
\hline
WikiPeople-3  &20656  &2582  &2582  &12270  &66 \\
WikiPeople-4  &12150  &1519  &1519  &9528  &50 \\
JF17K-3  &27635  &3454  &3455  &11541  &104 \\
JF17K-4  &7607  &951  &951  &6536  &23 \\
\hline
WD50K (33)  &73406  &10568  &18133  &38123  &474 \\
WD50K (66)  &35968  &5154  &8045  &27346  &403 \\
WD50K (100) &22738  &3279  &5297  &18791  &278 \\
\cdashline{1-6}
WikiPeople (33)  &28280  &3550  &3542  &20921  &145 \\
WikiPeople (66)  &14130  &1782  &1774  &13651  &133 \\
WikiPeople (100)  &9319  &1181  &1173  &8068  &105 \\
\cdashline{1-6}
JF17K (33)  &56959  &8122  &9112  &24081  &490 \\
JF17K (66)  &27280  &4413  &5403  &19288  &469 \\
JF17K (100)  &17190  &3152  &4142  &12656  &307 \\
\hline
WD50K  &166435  &23913  &46159  &47155  &531 \\
WikiPeople  &294439  &37715  &37712  &34825  &178 \\
JF17K   &76379  &-  &24568  &28645  &501 \\
\hline
\end{tabular*}
\label{table_statistics_datasets}
\end{table}

\subsection*{A\,\,\,Datasets}
\hypertarget{datasets}{}
We conduct  experiments on three hyper-relational KGs: WD50K~\citep{Mikhail_2020}, WikiPeople~\citep{Saiping_2019}, and JF17K~\citep{Jianfeng_2016}, with WD50K and WikiPeople being extracted from Wikidata~\citep{Denny_2014} and JF17K  from Freebase~\citep{Kurt_2008}.  These datasets have the following two characteristics. On the one hand, \emph{not} all main triples contain hyper-relational knowledge: only 13.6\% in WD50K, 2.6\% in  WikiPeople and 45.9\% in JF17K. On the other hand, each main triple has a different number of qualifier pairs: ranging from 0 to 20 in WD50K, 0 to 7 in WikiPeople and 0 to 4 in JF17K. To validate the models' robustness across multiple scenarios, these datasets have been refined taking into account the  two described characteristics~\citep{Zhiwei_2023}, i.e., the percentage of triples containing hyper-relational knowledge (mixed-percentage or fixed-percentage) and the number of qualifiers associated to main triples (mixed-qualifier or fixed-qualifier). We can thus construct three different sub-datasets, namely: \textit{Fixed-percentage Fixed-qualifier}, \textit{Fixed-percentage Mixed-qualifier}, and \textit{Mixed-percentage Mixed-qualifier}. Note  that the by definition \textit{Mixed-percentage Fixed-qualifier scenario does not  exist}. The statistics of the considered datasets are as shown in the Table~\ref{table_statistics_datasets}.

\begin{enumerate}[itemsep=0.5ex, leftmargin=5mm]
\item \textit{\textbf{Mixed-percentage Mixed-qualifier.}} This is the original setting introduced in~\citep{Mikhail_2020}, used to test the performance under variable number of main triples with qualifiers and with different number of qualifiers associated to those triples.

\item \textit{\textbf{Fixed-percentage Fixed-qualifier.}} 
Considering the monotonicity of this representional regime, we filter out the triples with 3 and 4 associated qualifiers to obtain WikiPeople-3, WikiPeople-4, JF17K-3, and JF17K-4.

\item \textit{\textbf{Fixed-percentage Mixed-qualifier.}} Following~\citep{Mikhail_2020, Zhiwei_2023}, we construct subsets where the number of qualifiers for each triple varies, but the proportion of triples with hyper-relational knowledge is fixed. For example, in the subsets WD50K (33), WD50K (66), and WD50K (100), the number in parentheses indicates the corresponding proportion of main triples with hyper-relational knowledge. We consider similar subsets for the WikiPeople and JF17K datasets.

\end{enumerate}

{It is noteworthy that the construction process of the dataset inherently satisfies the qualifier monotonicity property, \textit{i.e.,} for a given main triple, when new qualifier pairs are added to it, the corresponding answer space will not expand. Specifically, the qualifier monotonicity includes two aspects: \textit{from nothing to something} and \textit{from little to more}. Taking the WD50K dataset as an example (the same applies to any other dataset), the content prefix with \textit{P} and \textit{Q} below corresponds to the ID value from the WikiData knowledge graph.}
{
\begin{itemize}[itemsep=0.5ex, leftmargin=5mm]
\item \textbf{\textit{From nothing to something}}: This phrase encapsulates the process of extending a main triple into a hyper-relational fact by incorporating qualifier pairs. For instance, given the main triple (\textit{Q200405}, \textit{P1411}, \texttt{?}), the possible answers to \texttt{?} are \textit{Q4834546}, \textit{Q593098}, and \textit{Q103916}. By introducing a qualifier pair $\mathcal{Q}_1$ = (\textit{P1868}: \textit{Q272860}), the resulting hyper-relational fact is defined as $q_1$ = \{(\textit{Q200405}, \textit{P1411}, \texttt{?}), (\textit{P1868}: \textit{Q272860})\}. Consequently, the set of valid answers to \texttt{?} is refined to \textit{Q593098} and \textit{Q103916}, effectively eliminating \textit{Q4834546} as a possibility. This demonstrates how the incorporation of qualifiers progressively constrains the answer space. During dataset construction, we begin by extracting the main triple and iteratively introduce relevant qualifier pairs, gradually refining the possible answer set. The concept of \textit{from nothing to something} signifies a fundamental transition from conventional triple-based representations to hyper-relational facts enriched with hyper-relational knowledge, marking a significant conceptual shift from 0 to 1.
\item \textbf{From little to more} refers to the progressive enrichment of hyper-relational facts through the incorporation of additional qualifier pairs. For instance, by appending the qualifier pair $\mathcal{Q}_2$=(\textit{P805}: \textit{Q767355}) to $q_1$, we obtain $q_2$=\{(\textit{Q200405}, \textit{P1411}, \texttt{?}), (\textit{P1868}: \textit{Q272860}), (\textit{P805}: \textit{Q767355})\}. Consequently, the corresponding answer to \texttt{?} is refined to \textit{Q103916}, thereby further narrowing the answer set by excluding \textit{Q593098}. This process exemplifies the transition from an incomplete hyper-relational knowledge representation to a more comprehensive one, enhancing the specificity and informativeness of the captured knowledge.
\end{itemize}
}

\subsection*{B\,\,\,Baselines}
\hypertarget{baselines}{}
\zwh{We compare \texttt{HyperMono} with fourteen state-of-the-art baselines for HKGC, including \textbf{Embedding-based Methods}, \textbf{Transformer-based with GNNs Methods}, and \textbf{Transformer-based without GNNs Methods}: \\
The \textbf{Embedding-based Methods} include:
\begin{itemize}[itemsep=0.5ex, leftmargin=5mm]
\item 
\textit{m-TransH}~\cite{Jianfeng_2016} builds on TransH by transforming each hyper-relational fact using a star-to-clique conversion. 
\item
\textit{RAE}~\cite{Richong_2018} integrates m-TransH with a multi-layer perceptron by considering the relatedness of entities. However, since the distance-based scoring function of TransH enforces constraints on relations, it fails to represent some binary relations in KGs. 
\item 
\textit{NaLP-Fix}~\cite{Paolo_2020} uses a convolutional-based framework to compute a relatedness vector for a triple. However, the model does not distinguish between the main triple and relation-specific qualifiers.
\item 
{\textit{NeuInfer}~\cite{Saiping_2020} distinguishes the information in the same n-ary fact and represented each n-ary fact as a primary triple coupled with a set of its auxiliary descriptions.}
\item 
\textit{HINGE}~\cite{Paolo_2020} adopts a convolutional framework for modeling hyper-relational facts. However, it only operates on a triple-quintuple level that lacks  the necessary granularity for  representing a relation instance with its qualifiers.
\item 
\textit{ShrinkE}~\cite{Bo_2023} models the primary triplets as a spatial-functional transformation from the head into a relation-specific box.
\end{itemize}
The \textbf{Transformer-based with GNNs Methods} include:
\begin{itemize}[itemsep=0.5ex, leftmargin=5mm]
\item 
\textit{StarE}~\cite{Mikhail_2020} employs a message passing based graph encoder to obtain entity/relation embedings, and feeds these embedings to a transformer decoder to score hyper-relational facts.
\item 
\textit{QUAD}~\cite{Harry_2022} introduces a graph encoder that aggregates information from the perspective of the qualifier entities. 
% There are two variants of QUAD: QUAD and QUAD (Parallel). If there is no special suffix, QUAD denotes QUAD (Parallel). 
\end{itemize}
The \textbf{Transformer-based without GNNs Methods} include:
\begin{itemize}[itemsep=0.5ex, leftmargin=5mm]
\item 
\textit{Hy-Transformer}~\cite{Donghan_2021} replaces the graph encoder with light-weight embedding  modules, achieving higher efficiency without sacrificing effectiveness.
\item 
\textit{GRAN}~\cite{Quan_2021} represents a fact as a heterogeneous graph, and employs edge-specific attentive bias to capture both local and global dependencies within the given fact. 
% Note that GRAN contains three variants, i.e., GRAN{\small{-hete}} is the full model, GRAN{\small{-homo}} retains the graph structure but ignores heterogeneity and GRAN{\small{-complete}} considers neither the graph structure nor heterogeneity. If there is no special suffix, GRAN denotes GRAN{\small{-hete}}.
\item 
\textit{HyNT}~\cite{Chanyoung_2023} proposes a unified framework that learns representations of a hyper-relational knowledge graph containing numeric literals in either triples or qualifiers. 
\item 
\textit{HAHE}~\cite{Haoran_2023} presents a model with hierarchical attention in global and local level to encode the graphical structure and sequential structure via self-attention layers.
\item 
\textit{HyperFormer}~\cite{Zhiwei_2023} considers local-level sequential information to encode the content of entities, relations and qualifiers. It  includes an entity neighbor aggregator, relation qualifier aggregator and convolution-based bidirectional interaction modules.
\item 
{\textit{HyperCL}~\cite{Yuhuan_2024} designs relation-aware graph attention networks to capture the hierarchical ontology and a concept-aware contrastive loss to alleviate the dominance issue.}
\end{itemize}
}

\begin{figure*}[htbp]
  \centering
  \includegraphics[width=0.23\linewidth]{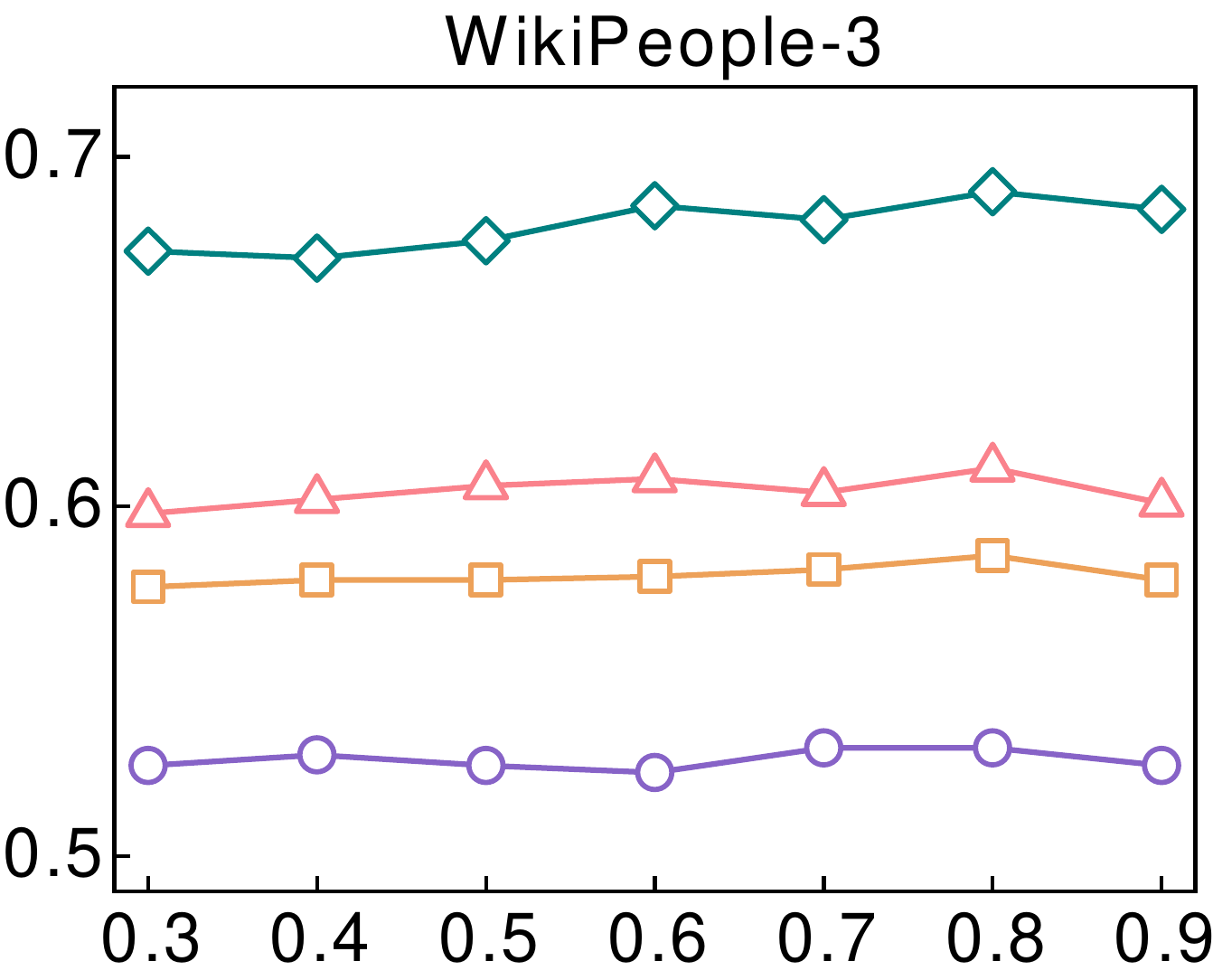}\hfill
  \includegraphics[width=0.23\linewidth]{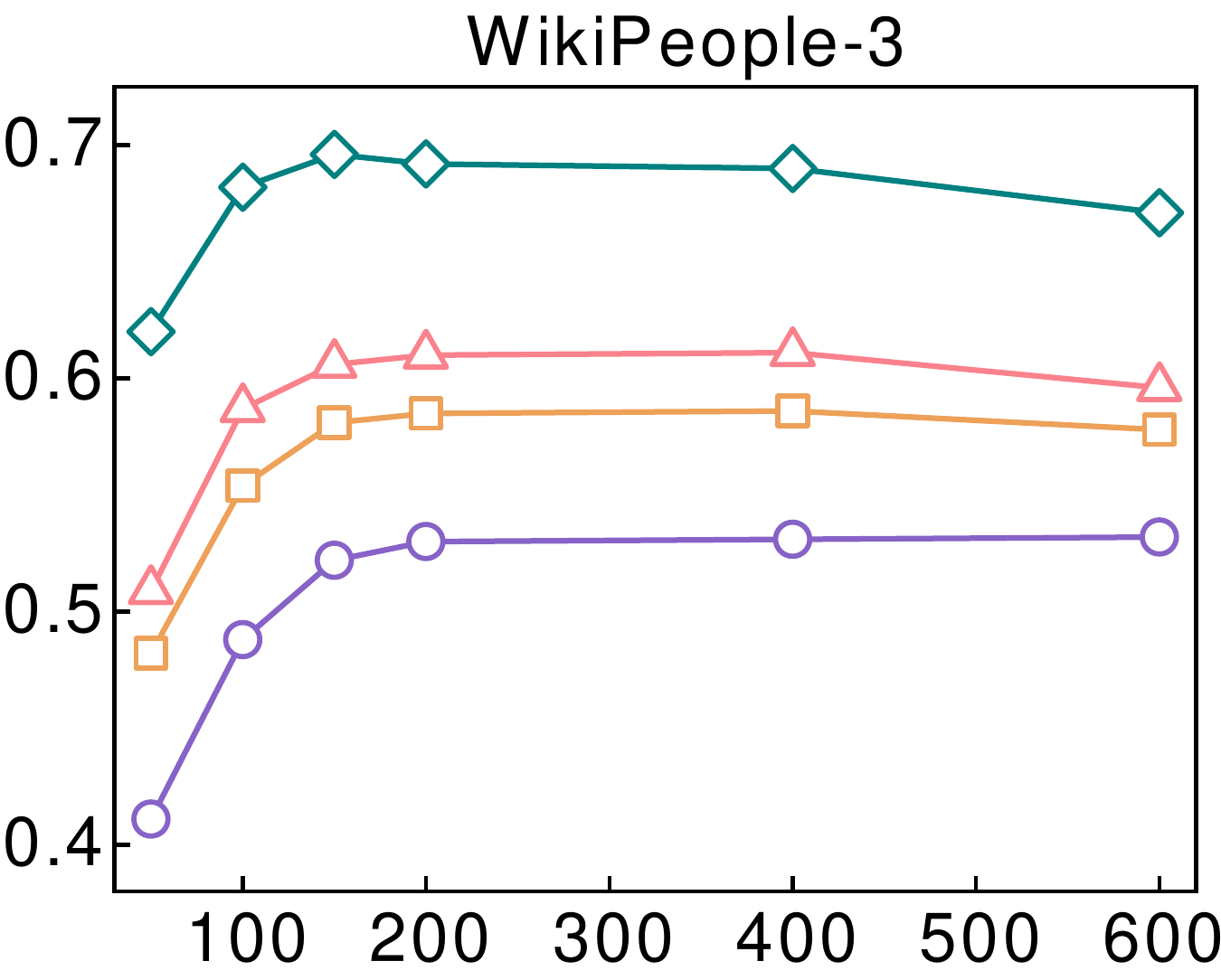}\hfill
  \includegraphics[width=0.23\linewidth]{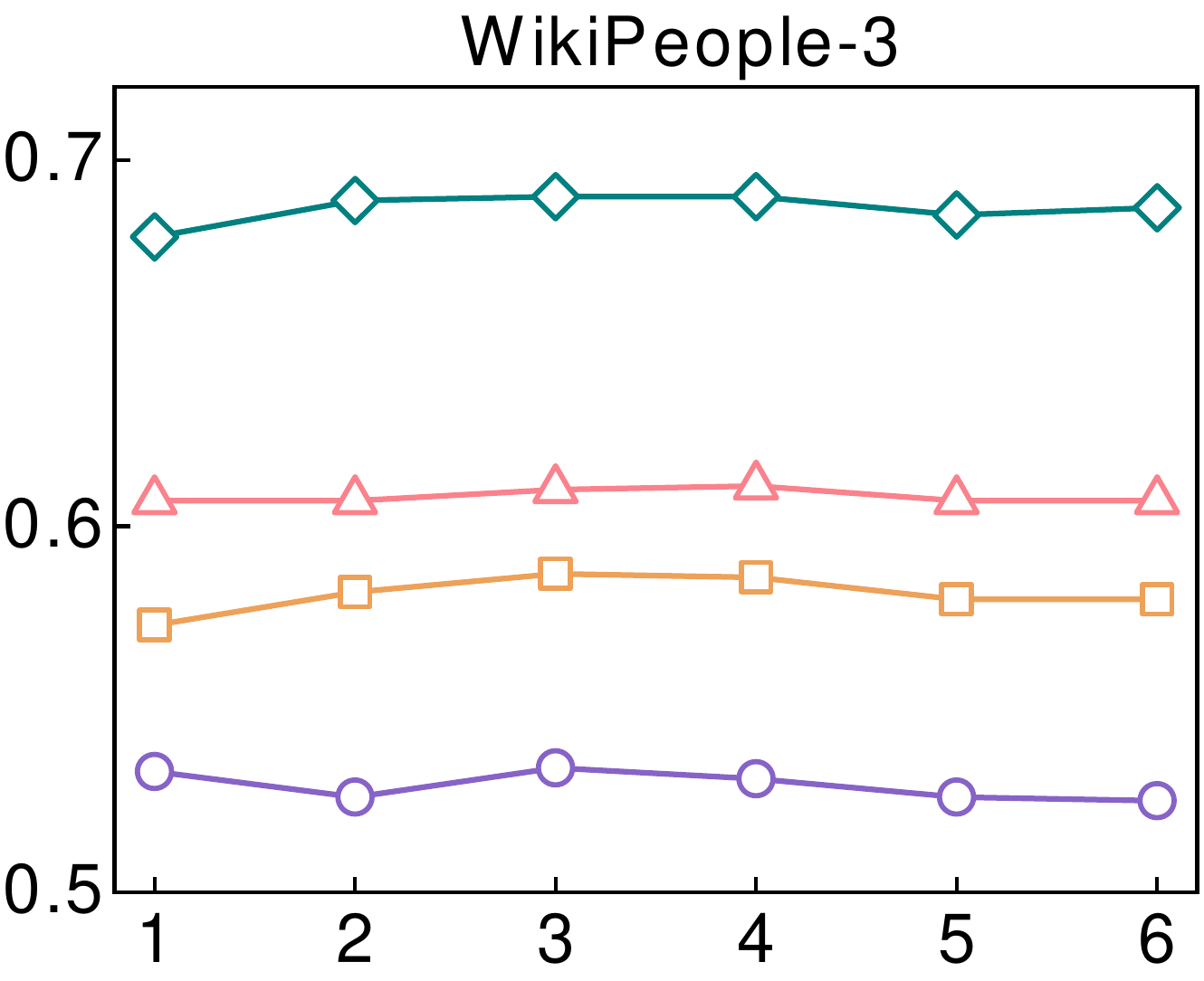}\hfill
  \includegraphics[width=0.24\linewidth]{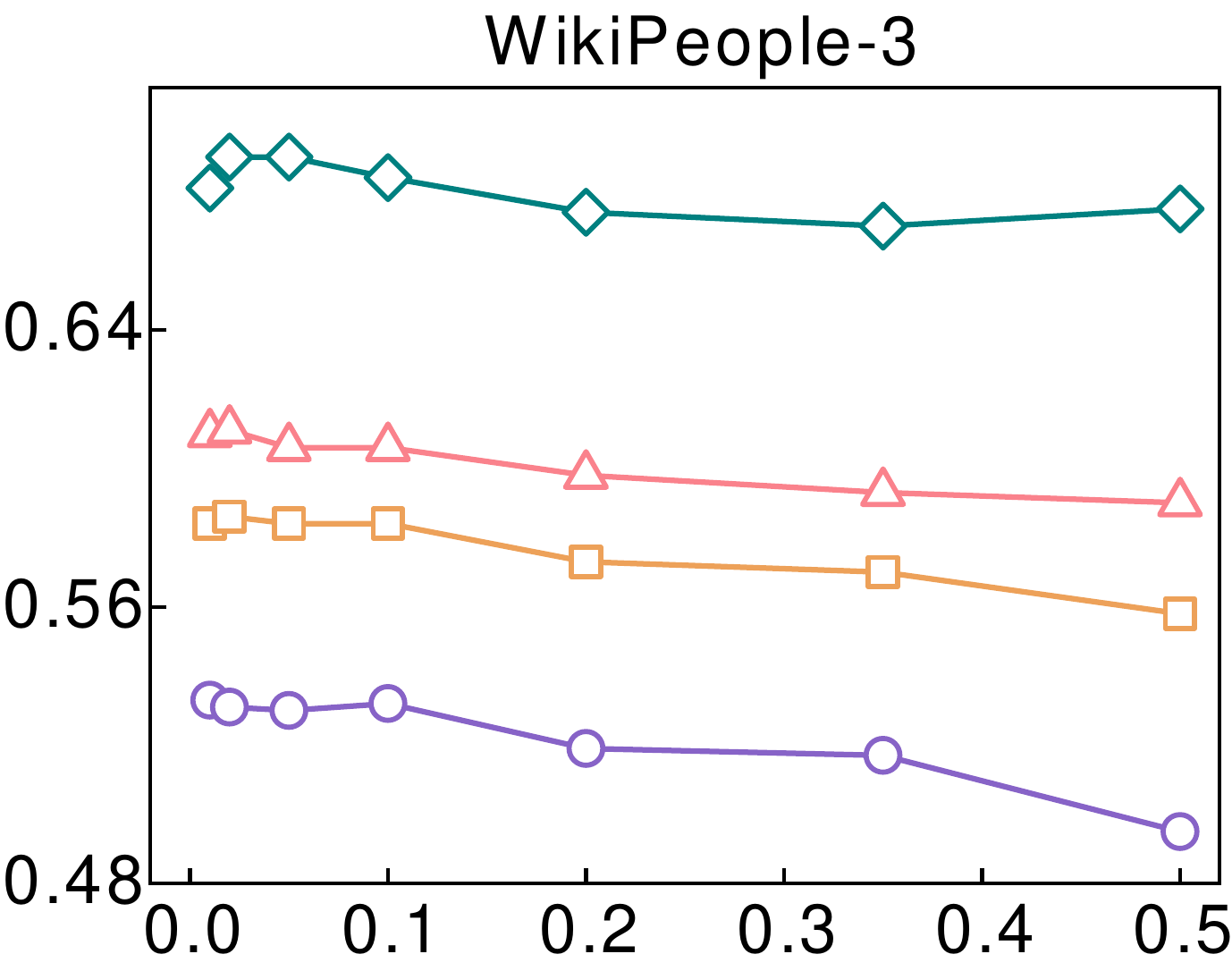}
  
  \includegraphics[width=0.235\linewidth]{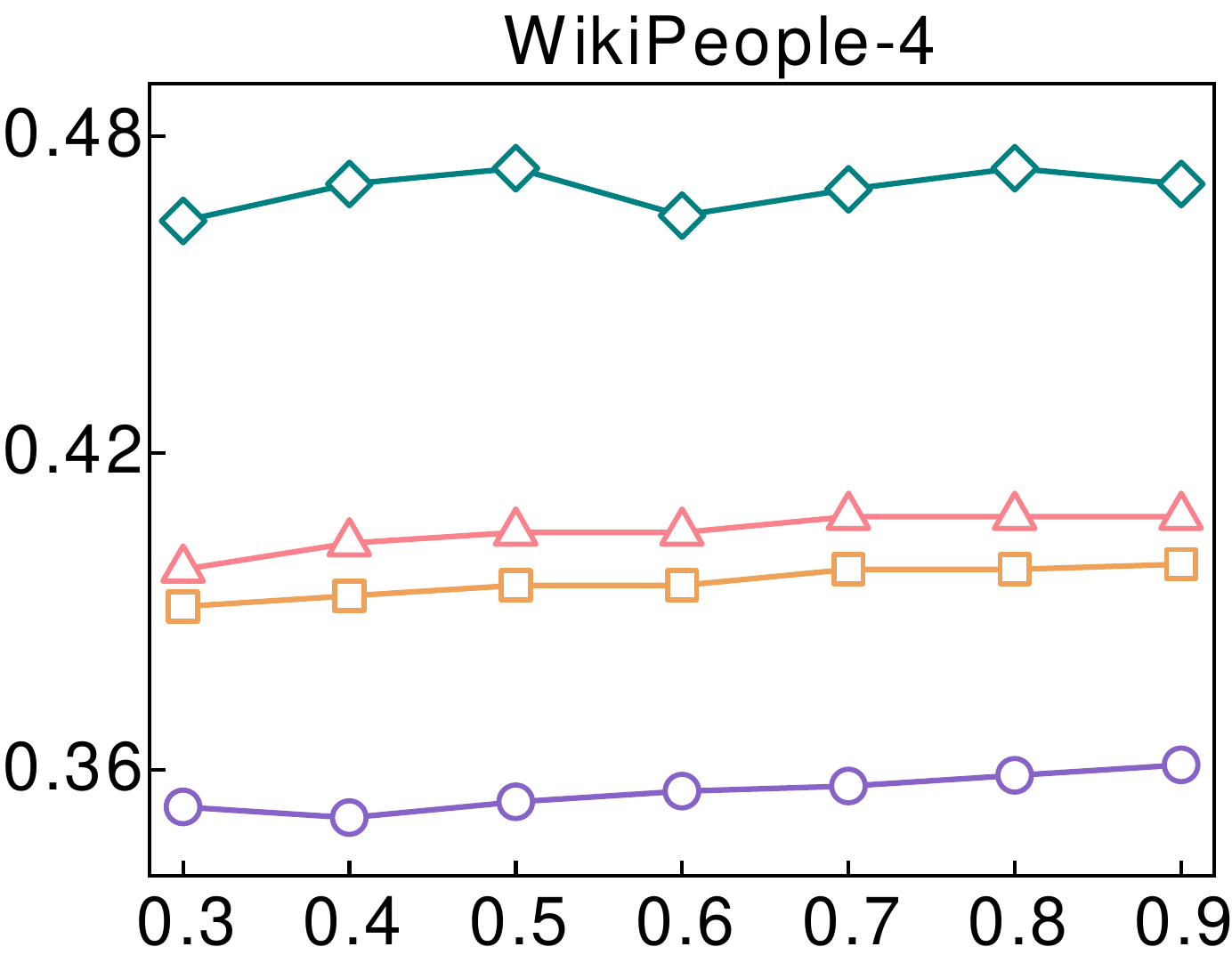}\hfill
  \includegraphics[width=0.242\linewidth]{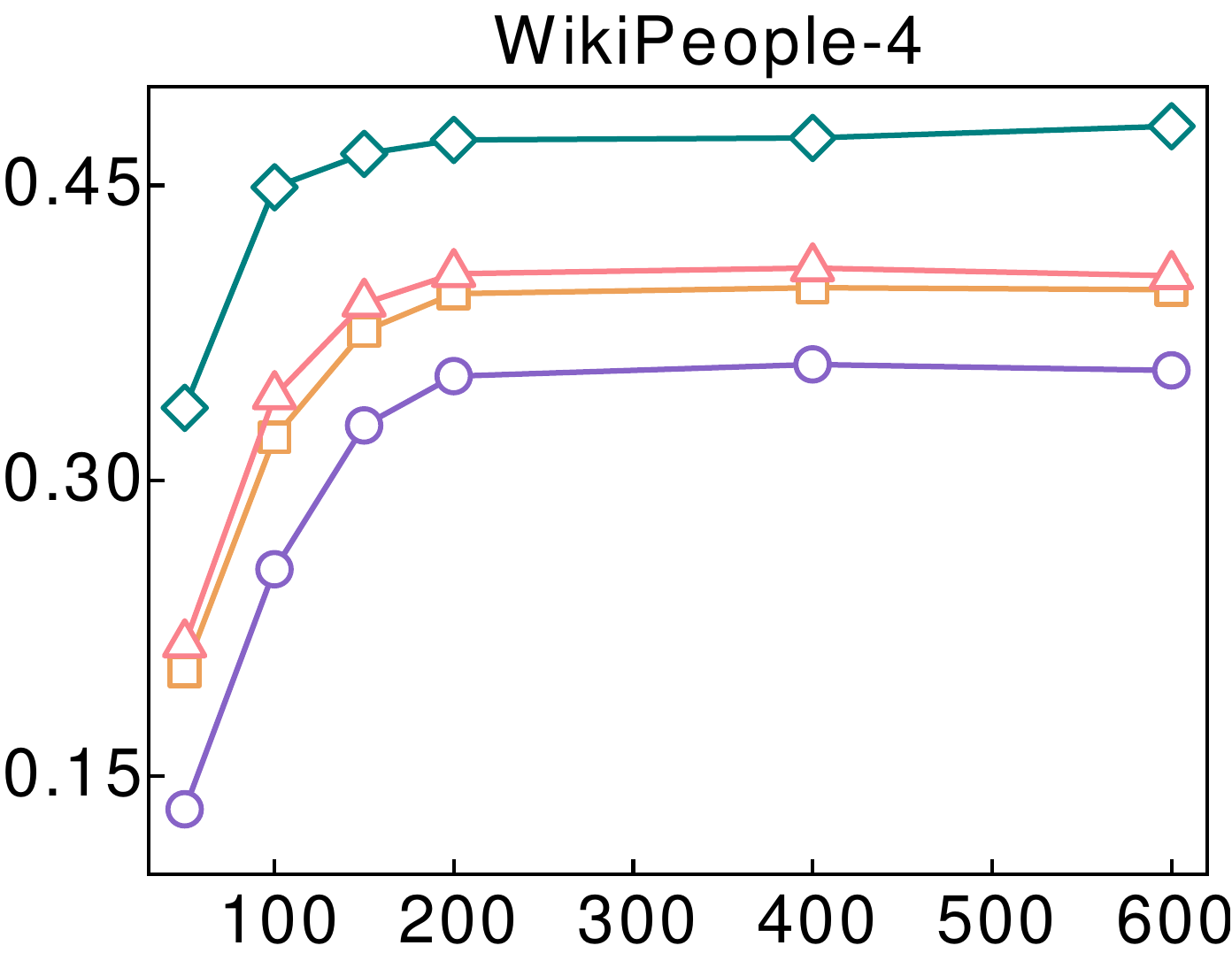}\hfill
  \includegraphics[width=0.24\linewidth]{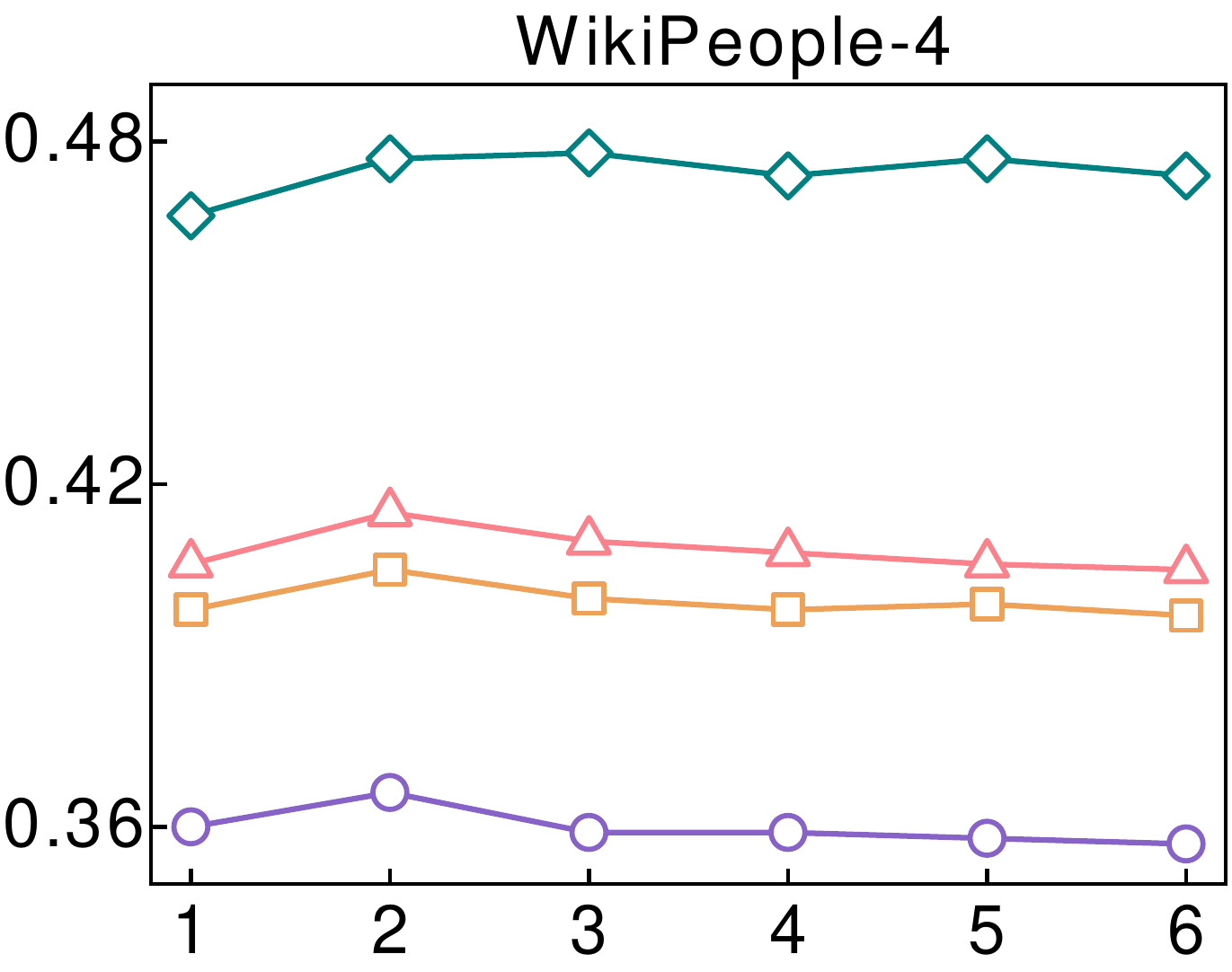}\hfill
  \includegraphics[width=0.24\linewidth]{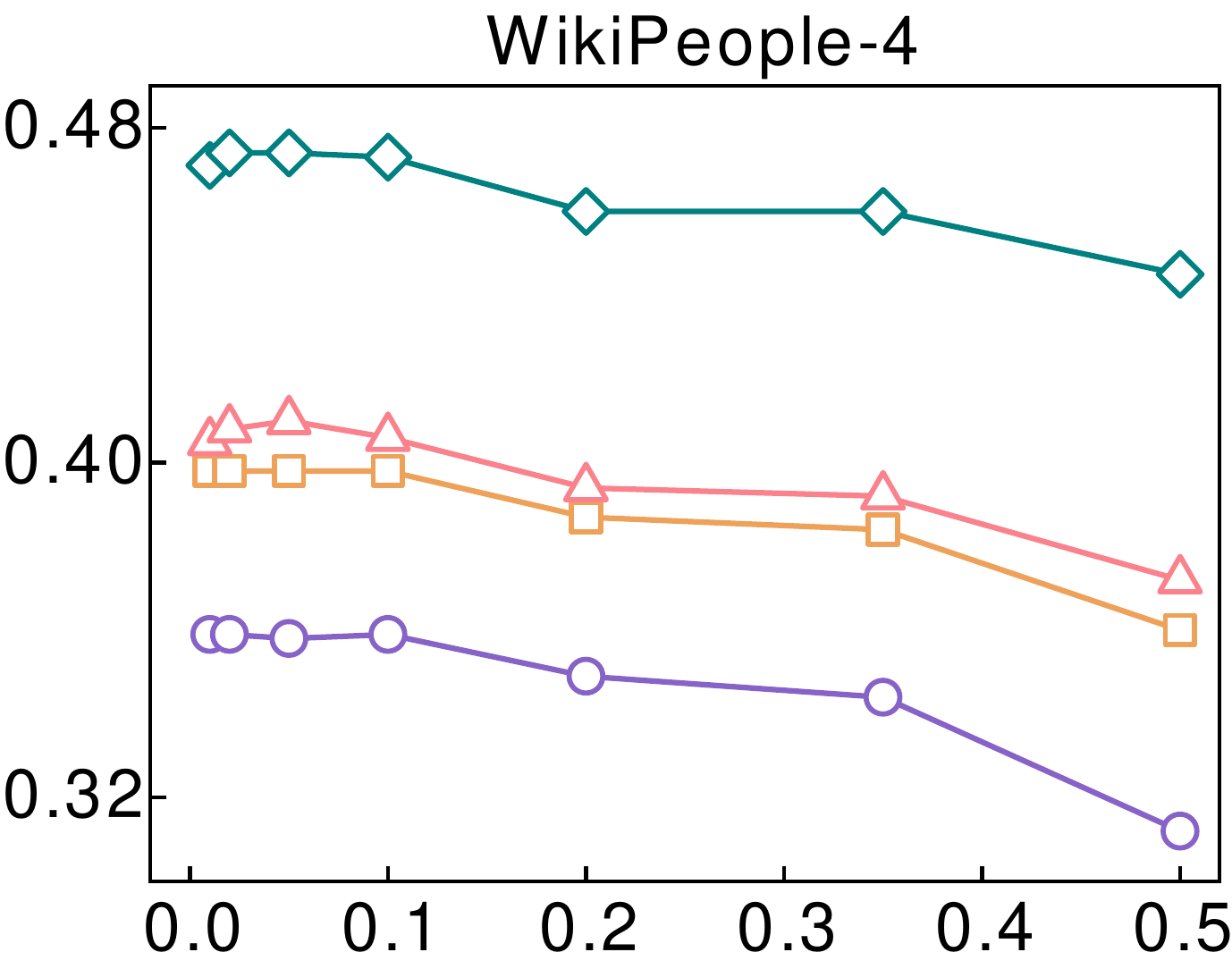}

  \includegraphics[width=0.238\linewidth]{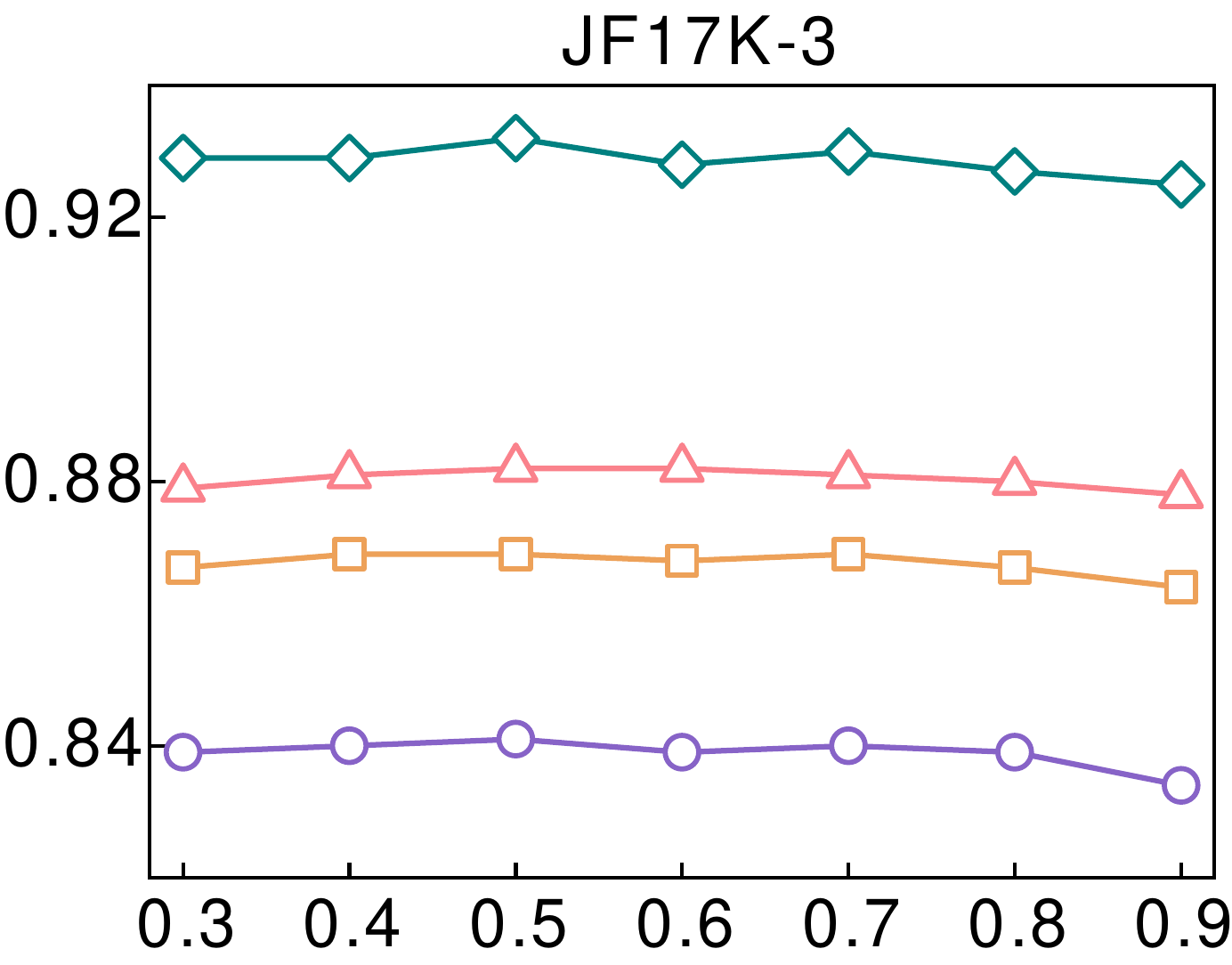}\hfill
  \includegraphics[width=0.236\linewidth]{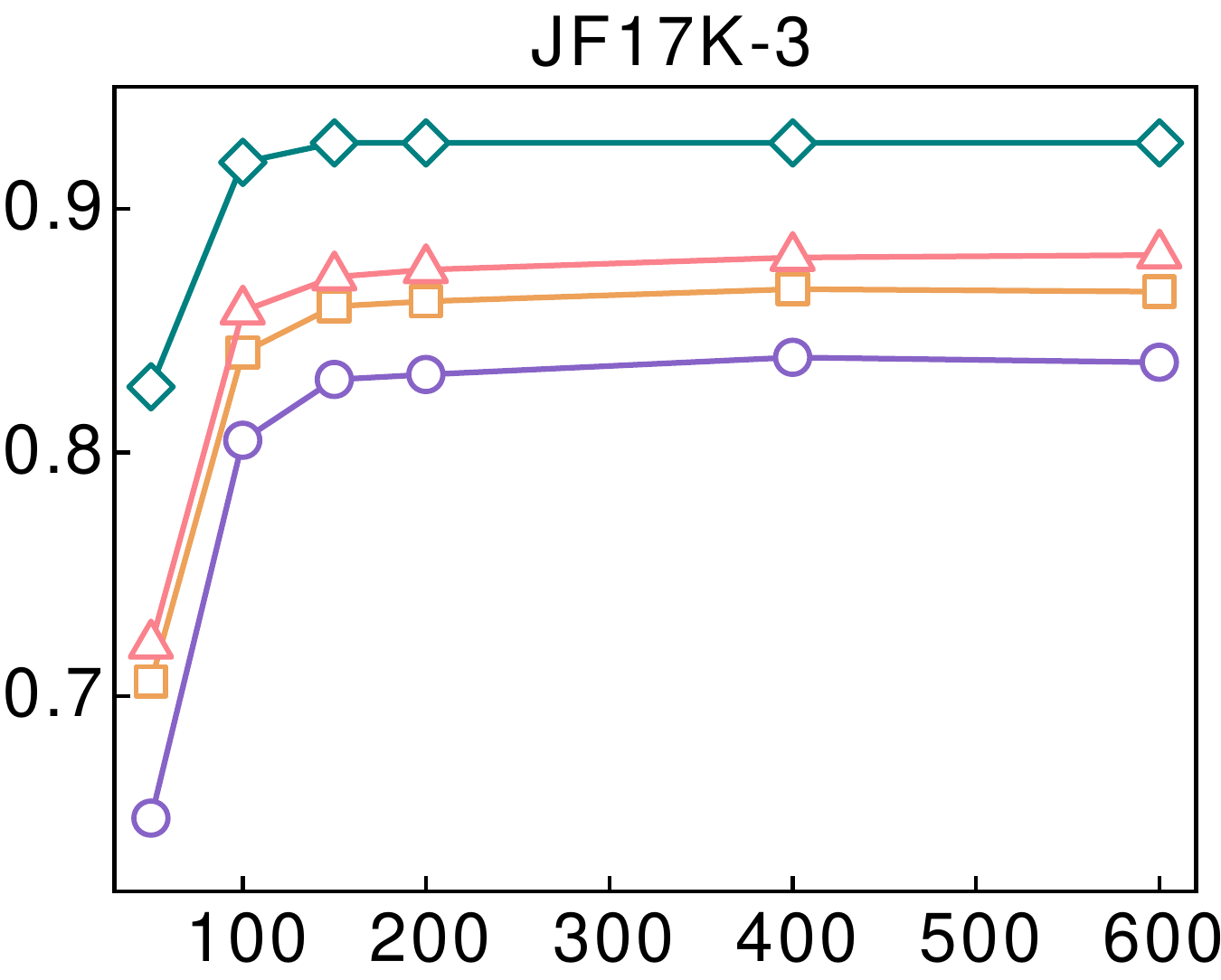}\hfill
  \includegraphics[width=0.24\linewidth]{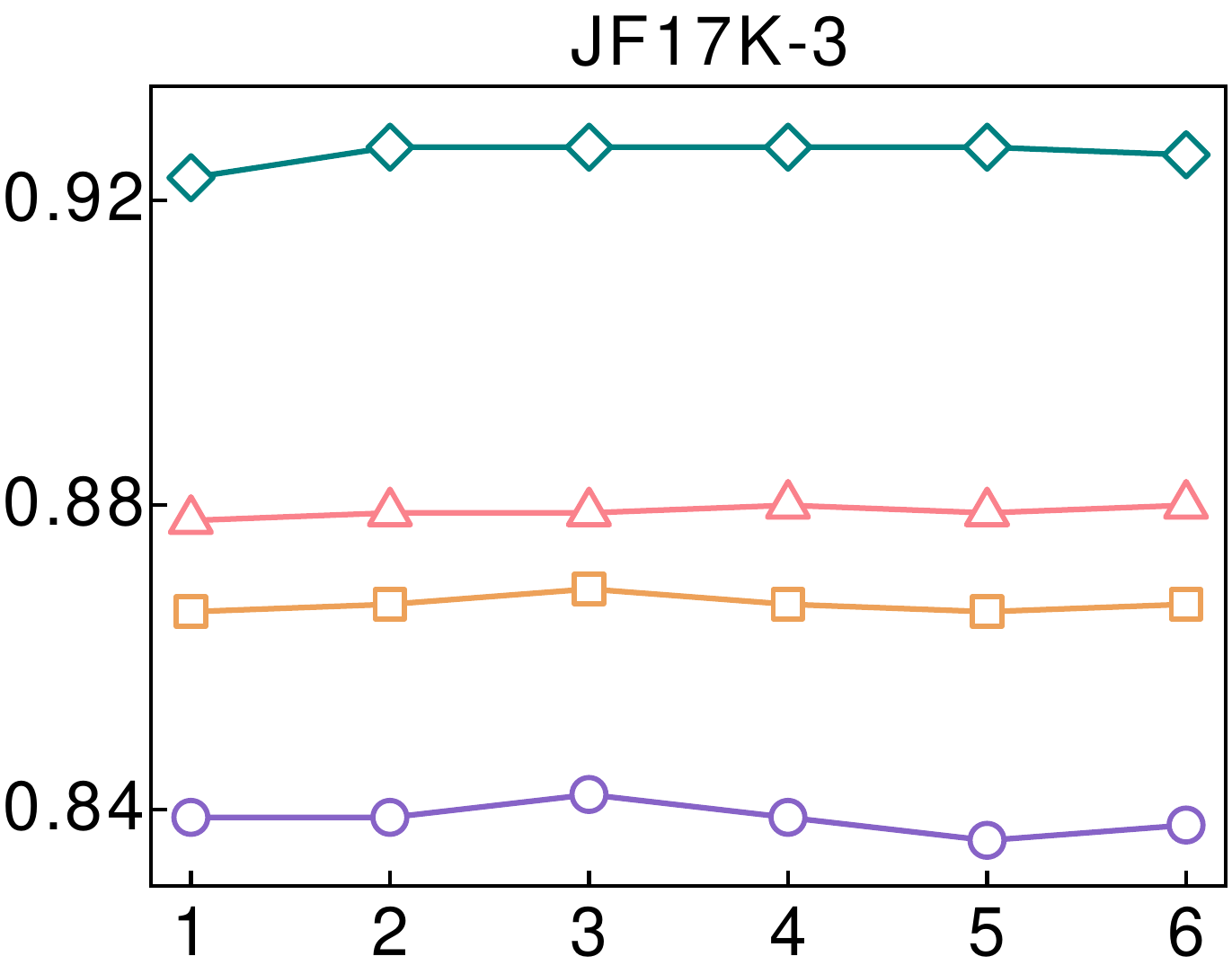}\hfill
  \includegraphics[width=0.24\linewidth]{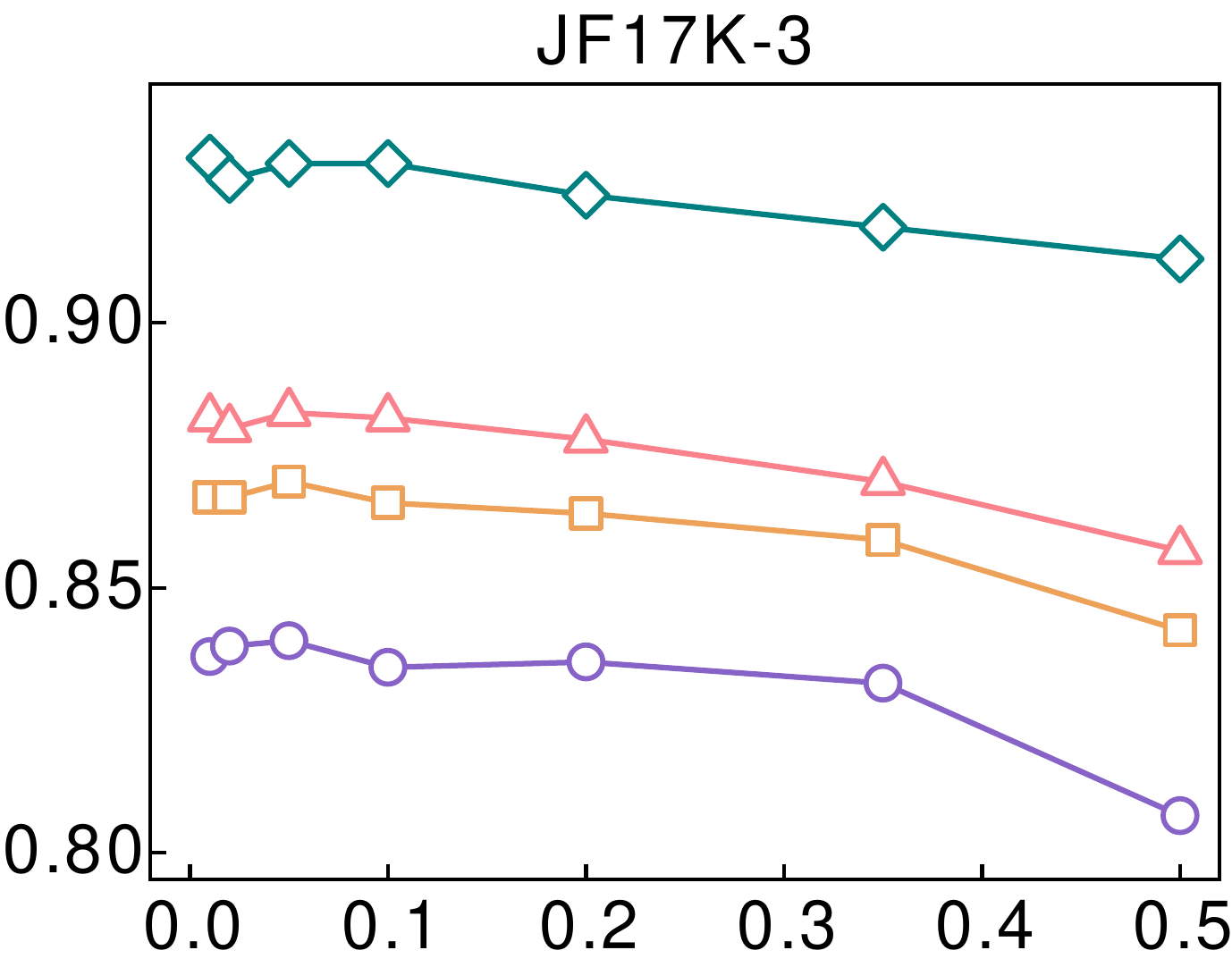}

  \subcaptionbox{Label Smoothing}{\includegraphics[width=0.238\linewidth]{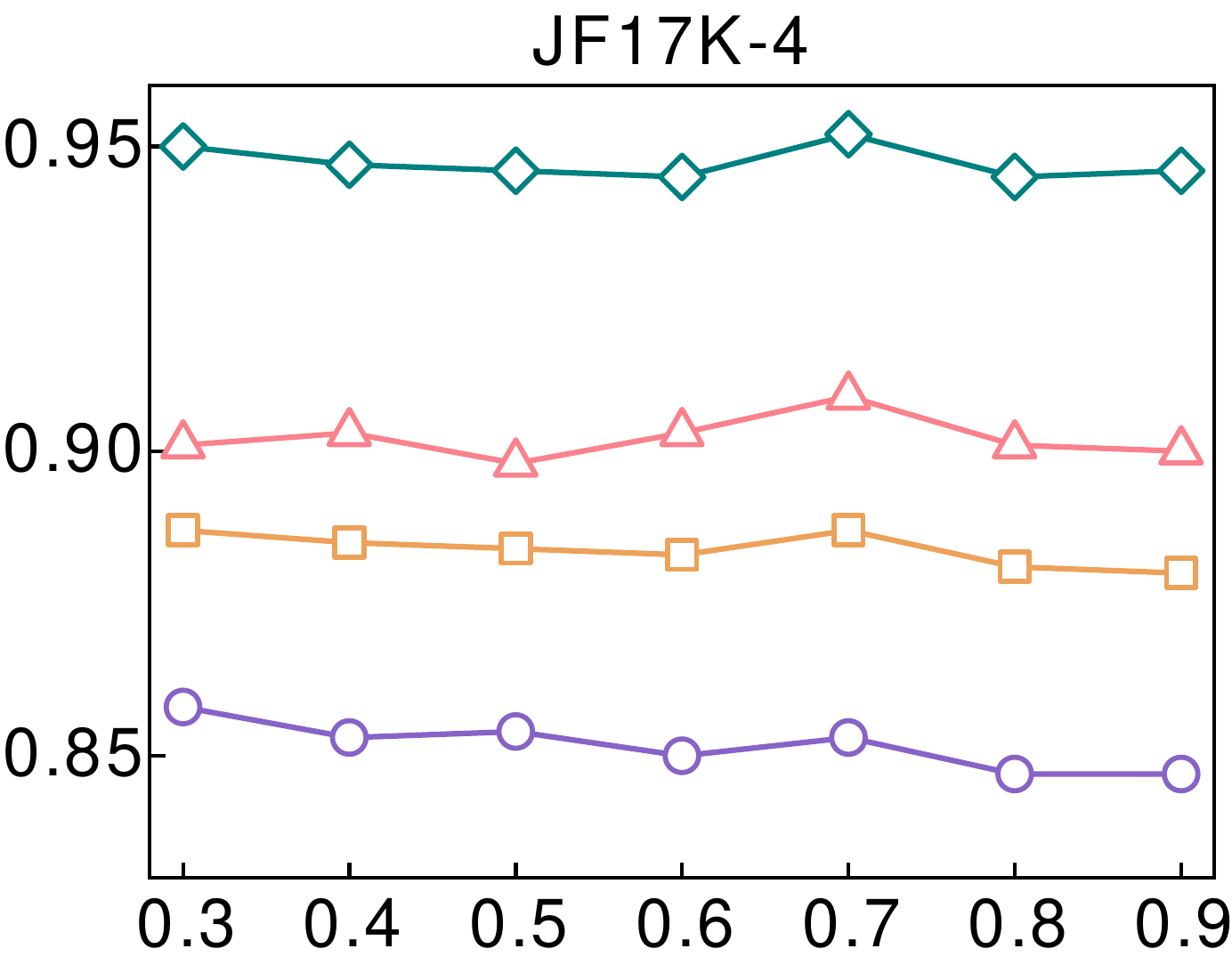}}\hfill
  \subcaptionbox{Embedding Dimension $d$}{
  \includegraphics[width=0.236\linewidth]{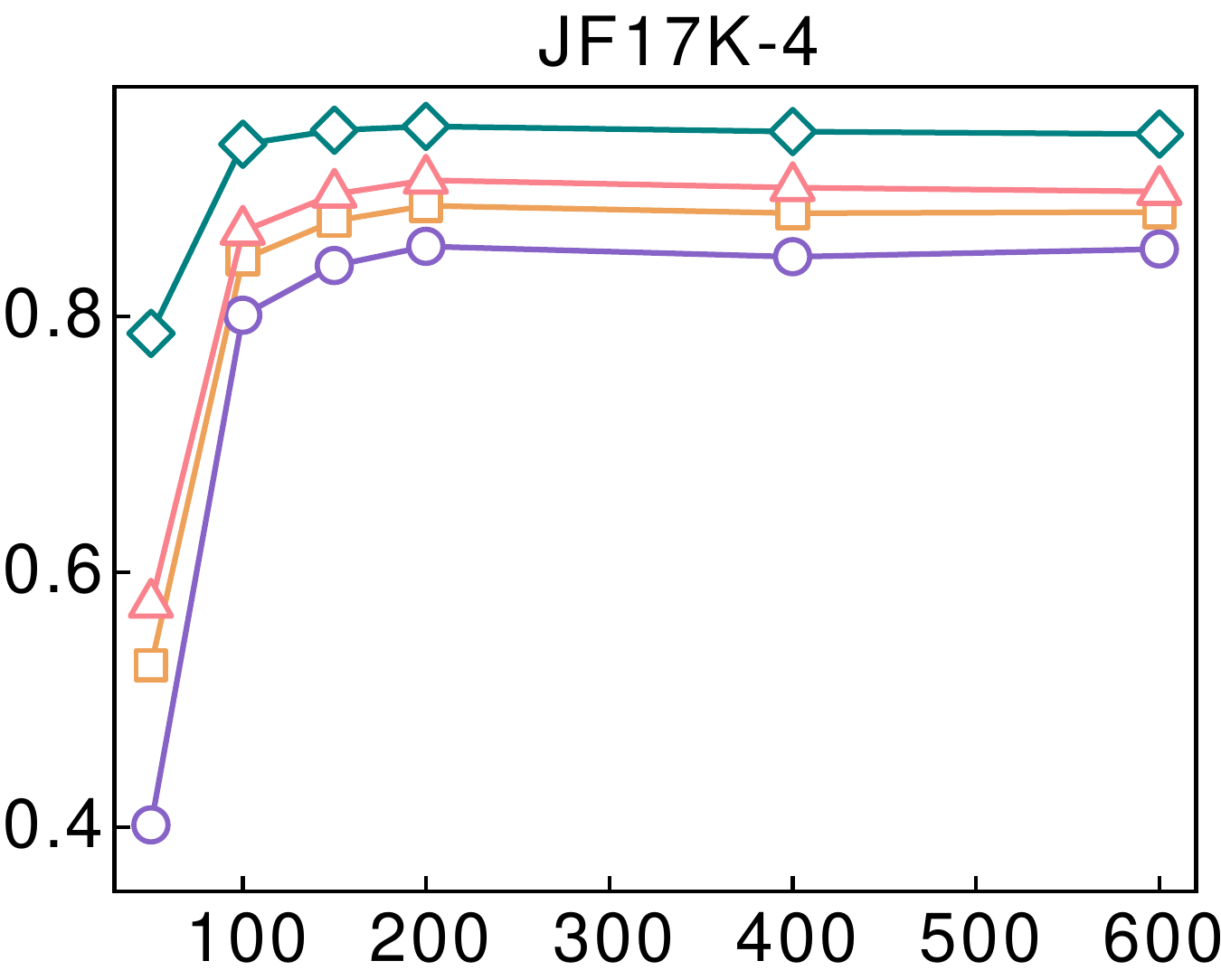}}\hfill
  \subcaptionbox{Temperature $\gamma$}{
  \includegraphics[width=0.24\linewidth]{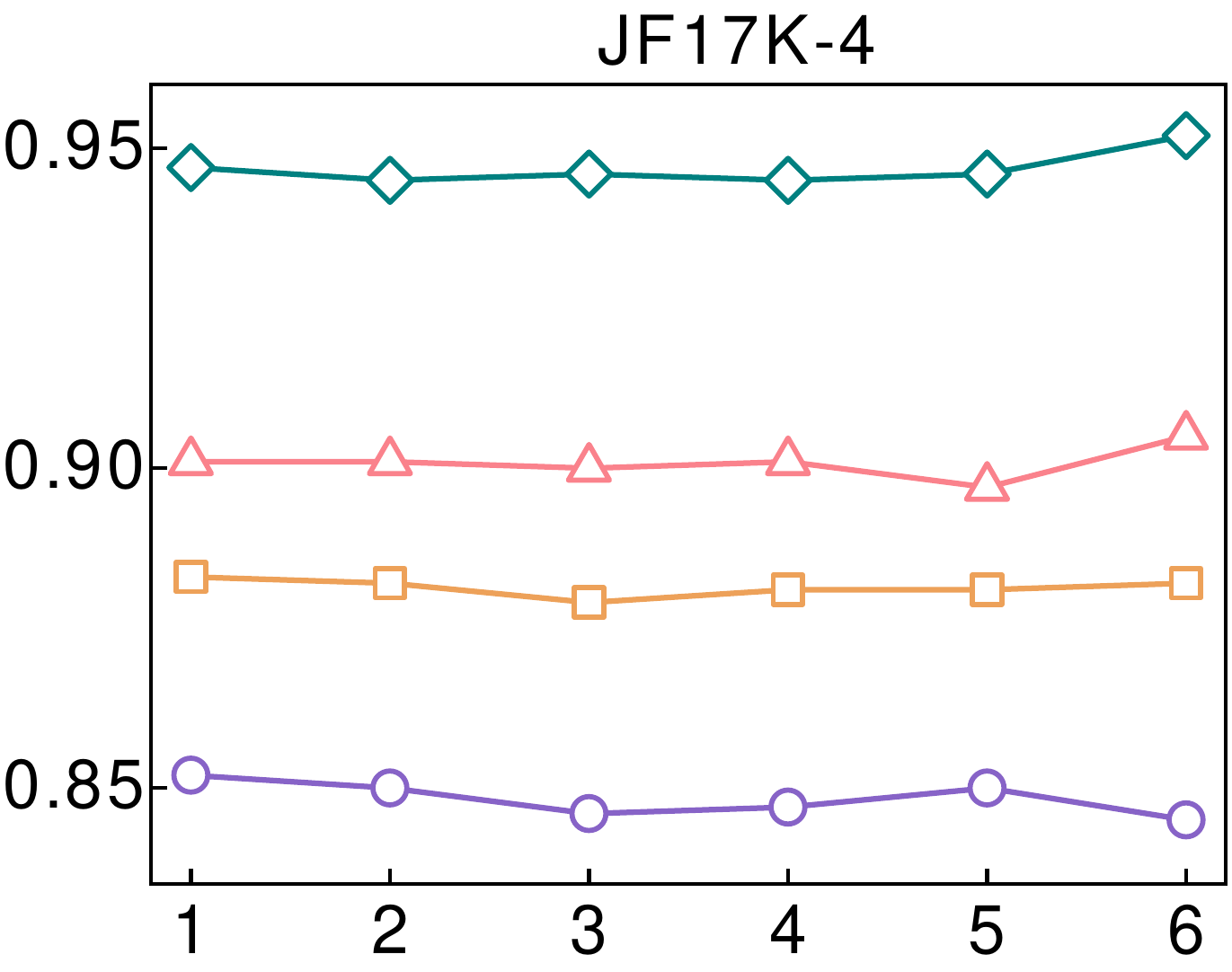}}\hfill
  \subcaptionbox{Initialize Range}{
  \includegraphics[width=0.24\linewidth]{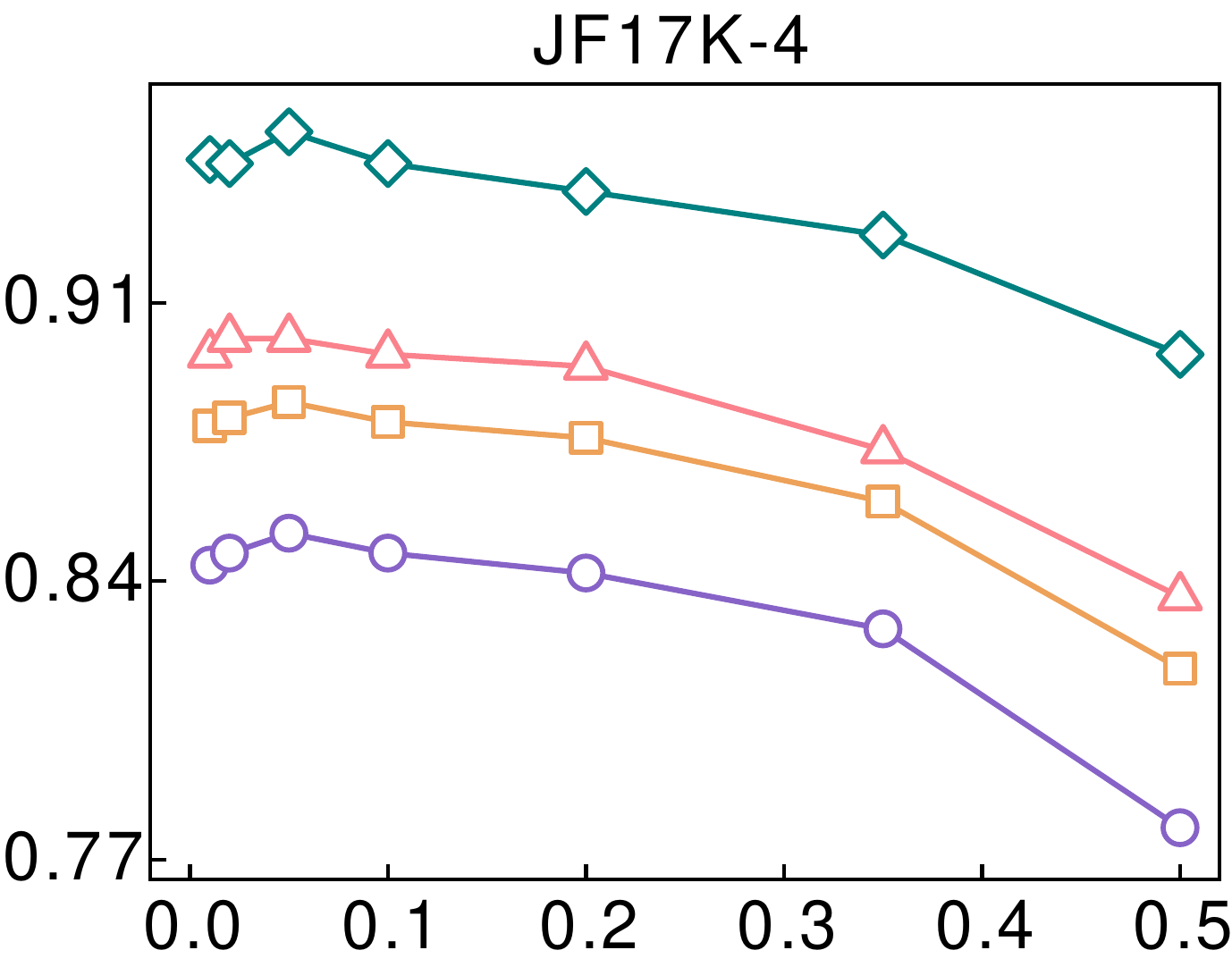}}

  \includegraphics[width=0.4\linewidth]{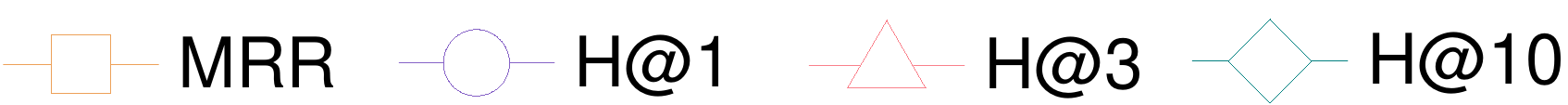}
  \caption{Parameter sensitivity experiments under different conditions on Wikipeople-3, Wikipeople-4, JF17K-3 and JF17K-4 datasets.}
\label{figure_hyper_parameters}
\end{figure*}

\subsection*{C\,\,\,Evaluation Protocol and Implements Details}
\hypertarget{protocol_and_implements}{}

\smallskip
\noindent \zwh{\textit{\textbf{Evaluation Protocol.}}} 
Following existing work~\cite{Chanyoung_2023, Bo_2023, Haoran_2023, Zhiwei_2023}, we rank all entities to predict the head entity in query $(\texttt{?}, r, t, \mathcal{Q})$ or the tail entity in query $(h, r, \texttt{?}, \mathcal{Q})$. Specifically, we employ MRR and Hits@$k$ (abbreviated sometimes as H@\textit{k}, $k \in \{1, 3, 10\}$) as evaluation metrics. MRR defines the inverse of the rank for the first correct answer, while Hits@k calculates the percentage of correct types ranked among the top-\textit{k} answers. For all metrics, the larger the value, the better. Following the standard evaluation protocol in the filtered setting~\cite{Antoine_2013}, all true triples in the HKGs are filtered out during evaluation, since predicting a low rank for these triples should not be penalized. All metrics are computed by averageing over two directions: head entity prediction and tail entity prediction.

\smallskip
\noindent \zwh{\textit{\textbf{Implementation Details.}}} 
\zwh{We conduct all experiments on six 32G Tesla V100 GPUs, {use the Pytorch 1.8.1 framework (it should be mentioned that using Pytorch 2.1.2 will result in incorrect running results)}, and employ the AdamW~\cite{adamw_2019} optimizer and a cosine decay scheduler with a linear warm-up for optimization. The hyperparameters are tuned according to the MRR on the validation set. We use  grid search  to select the optimal hyperparameters, mainly including: the learning rate $lr \in$ \{3e-4, 4e-4, 5e-4, \textbf{6e-4}, 7e-4\}, the label smooth values $ls \in$ \{0.3, 0.4, 0.5, 0.6, 0.7, \textbf{0.8}, 0.9\}, the embedding dimension $d \in$ \{50, 100, 150, \textbf{200}, 400, 600\}, the temperature of Equation (\ref{equation_1})  $\gamma \in$ \{1, 2, 3, \textbf{4}, 5, 6\}, the initialization range of embeddings  $ir \in$ \{0.01, \textbf{0.02}, 0.05, 0.1, 0.2, 0.35, 0.5\}, the neighbors of the head entity  \{2, \textbf{3}, 4\}, the number of qualifiers for a relation  \{5, \textbf{6}, 7\}, the number of layers in the transformer \{2, 4, \textbf{8}\}, the number of heads in transformer  \{1, 2, 4, \textbf{8}\}, the input dropout rate in the transformer  \{0.6, \textbf{0.7}, 0.8\}.} {For the reproducibility of the model, we also provide the running logs of \texttt{HyperMono} in different scenarios in the anonymous code link: \url{https://anonymous.4open.science/r/HKGC-HyperMono-AE61}.}

\subsection*{D\,\,\,Parameter Sensitivity}
\hypertarget{parameter_sensitivity}{}
\zwh{To address \textbf{RQ3}, we carry out parameter sensitivity experiments on the WikiPeople-3, WikiPeople-4, JF17K-3, and JF17K-4 datasets, including: a) effect of label smoothing; b) effect of embedding dimension $d$; c) effect of temperature $\gamma$, and d) effect of initialization range. The corresponding results are shown in Figure~\ref{figure_hyper_parameters}.}

\smallskip
\noindent\zwh{\textbf{Effect of Label Smoothing.}} 
\zwh{Label smoothing is widely used in tasks related to knowledge graphs~\citep{Xiao_2022, Yang_2022}. It introduces a degree of uncertainty, making the model more adaptable to different data distributions during the training process. From Figure~\ref{figure_hyper_parameters}(a), we observe that the values of label smoothing have an impact on the performance, but there is no evident pattern. We need to make appropriate adjustments to its values based on the dataset. For instance, we select 0.8 and 0.5 as the most suitable label smoothing values for the WikiPeople-4 and JF17K-3 datasets, respectively.}

\smallskip
\noindent\zwh{\textbf{Effect of Embedding Dimension $d$.}}
\zwh{In principle,  more semantic knowledge could be captured by having a higher number of dimensions in  entity, relation, and qualifiers embeddings. However, as shown in Figure~\ref{figure_hyper_parameters}(b), there is not a strict proportional relationship between embedding dimensions and the actual performance. After reaching a certain value of embedding dimensions, the accuracy stabilizes. In practice, one chooses the minimum value that achieves stability. For example, for the WikiPeople-4 and JF17K-3 datasets, the embedding dimension $d$ is set at 200.}

\smallskip
\noindent\zwh{\textbf{Effect of Temperature $\gamma$.}} 
\zwh{The temperature coefficient $\gamma$ in Equation (\ref{equation_1}) can be used to adjust the contribution of each neighbor of an  entity. Figure~\ref{figure_hyper_parameters}(c) illustrates the performance difference with values $\gamma$ ranging from 1 to 6. One can observe that the choice of the $\gamma$ value has a minimal impact on the final performance. The performance curve only exhibits local oscillations, indicating that the \texttt{HyperMono} model is robust to the choice of the $\gamma$ value.}

\smallskip
\noindent\zwh{\textbf{Effect of the Initialization Range.}}
\zwh{We observe that the initialization value range of embeddings has a significant impact on \texttt{HyperMono}'s performance. As shown in Figure~\ref{figure_hyper_parameters}(d), when the  initialization values are relatively small, the performance of \texttt{HyperMono} is more stable. However, the model becomes sensitive to larger initialization values. In practice, it is important to carefully set the initialization range to prevent the model from experiencing  sharp  performance declines.}

\subsection*{E\,\,\,Complexity Analysis}
\hypertarget{complexity_analysis}{}
{To address \textbf{RQ4}, 
we conduct a complexity analysis of HyperMono and HyperFormer (the best performing baseline in most cases) across the WD50K, WikiPeople, and JF17K datasets from the following five perspectives: a) \textit{time complexity}, which is measured by the number of floating-point operations (\#FLOPs) required during training phase; b) \textit{space complexity}, which refers to the amount of a model's parameters (\#Params); c) \textit{model training time} (\#TrainTime), measured in hours (\textit{h}); d) \textit{inference time per sample} (\#InferTime), measured in milliseconds (\textit{ms}); e) \textit{GPU memory consumption} (\#GPU), measured in gigabytes (G). All the aforementioned metrics are desired to be as minimal as possible, as lower values indicate reduced resource expenditure and greater efficiency. The corresponding results are shown in Table~\ref{table_flops_and_params}. We can observe that HyperMono requires significantly less overhead, specifically, on  the WD50K dataset, \texttt{HyperMono} only requires nearly half the \#FLOPs and one-sixth the \#Params of HyperFormer, which demonstrates that \texttt{HyperMono} is more efficient and can achieve better results while occupying less time and space, a similar trend is observed in \#TrainTime, \#InferTime, and \#GPU.}

% we analyze \texttt{HyperMono}'s complexity on the WD50K, WikiPeople, and JF17K datasets from two perspectives: time complexity and space complexity. The corresponding results are shown in Table~\ref{table_flops_and_params}.  The time complexity is measured by the number of floating-point operations (\#FLOPs) required during training phase, while the space complexity refers to the amount of a model's parameters (\#Params). The larger the values of \#FLOPs and \#Params are, the more computing power and higher memory usage are required during the training process. We observe that compared with HyperFormer, \texttt{HyperMono} requires less time and space overhead in each dataset. Specifically, on  the WD50K dataset, \texttt{HyperMono} only requires nearly half the \#FLOPs and one-sixth the \#Params of HyperFormer, which demonstrates that \texttt{HyperMono} is more efficient and can achieve better results while occupying less time and space.
\begin{table}[!htp]
\setlength{\abovecaptionskip}{0.03cm}
\renewcommand\arraystretch{1.2}
\setlength{\tabcolsep}{0.25em}
\centering
\small
\caption{{The quantitative results of \texttt{HyperFormer} and \texttt{HyperMono} under different complexity evaluation metrics.}}
\begin{tabular*}{0.88\linewidth}{@{}ccccc@{}}
\hline
\multicolumn{1}{c}{\textbf{Metrics}}&\multicolumn{1}{c}{\textbf{Model}}&\multicolumn{1}{c}{\textbf{WD50K}}&\multicolumn{1}{c}{\textbf{WikiPeople}}&\multicolumn{1}{c}{\textbf{JF17K}} \\
\hline
\multirow{2}{*}{\#FLOPs} &\texttt{HyperFormer} &118.397G &118.167G &118.070G \\
&\texttt{HyperMono} &69.600G &69.371G &69.273G \\
\hline
\multirow{2}{*}{\#Params} &\texttt{HyperFormer} &66.956M &66.956M &66.956M \\
&\texttt{HyperMono} &10.645M &10.645M &10.645M \\
\hline
\multirow{2}{*}{\#TrainTime} &\texttt{HyperFormer} &39\textit{h} &68.7\textit{h} &15.9\textit{h} \\
&\texttt{HyperMono} &31\textit{h} &50.7\textit{h} &12.1\textit{h} \\
\hline
\multirow{2}{*}{\#InferTime} &\texttt{HyperFormer} &6.0\textit{ms} &6.5\textit{ms} &6.0\textit{ms} \\
&\texttt{HyperMono} &4.7\textit{ms} &5.0\textit{ms} &4.2\textit{ms} \\
\hline
\multirow{2}{*}{\#GPU} &\texttt{HyperFormer} &23.9G &23.5G &23.6G \\
&\texttt{HyperMono} &9.5G &8.5G &8.0G \\
\hline
\end{tabular*}
\label{table_flops_and_params}
\end{table}

\subsection*{F\,\,\,Additional Results}
\hypertarget{additional_results}{}
{To demonstrate the robustness of \texttt{HyperMono}, we further conducted incremental experiments, including: a) effect of pooling operation; b) effect of aggregation operation; c) effect of embedding space; d) effect of learning rate; e) effect of neighbor numbers; f) effect of sparse ratio. The corresponding results are shown in Table~\ref{table_additional_results}, Table~\ref{table_additional_parameters_results}, and Table~\ref{table_removing_neighbors}.}

\smallskip
\noindent{\textbf{Effect of Pooling Operation.}}
{In Equation~\ref{equation_1}, we adopt the weighted pooling method to integrate multiple inference results. To investigate the impact of different pooling strategies, we further replace the weighted pooling with mean pooling, max pooling, and the sum operation. The experimental results on the WikiPeople-3, WikiPeople-4, JF17K-3, and JF17K-4 datasets are presented at the top of Table~\ref{table_additional_results}. We can observe that the weighted pooling operation generally yields the best performance, which proves the rationality of our choice of weighted pooling operation.}

\smallskip
\noindent{\textbf{Effect of Aggregation Operation.}}
{For $\textbf{\textit{P}}_{G}^{h}$ in the Coarse-grained Neighborhood Aggregator and $\textbf{\textit{Q}}_{G}^{h}$ in the Fine-grained Neighborhood Aggregator, we use the mean operation to average the entity vector predicted by different relational neighbors to aggregate them. To understand the impact of different operations, we replace mean with max, sum, and attention operations, and conduct experiments on WikiPeople-3, WikiPeople-4, JF17K-3 and JF17K-4, the corresponding results are shown in the middle of Table~\ref{table_additional_results}. We can observe that replace mean operation in GEI with other methods does not result in significant performance fluctuations for \texttt{HyperMono} across the four datasets. Therefore, this component is not the key factors affecting performance.}

\smallskip
\noindent{\textbf{Effect of Embedding Space.}}
{Our model employs the cone space to encode qualifier monotonicity. To assess the rationale behind the selection of the cone space, we substitute it with Beta~\citep{Hongyu_2020_1} and Gamma~\citep{Dong_2022} probability distribution spaces. It is essential to note that the chosen embedding space must support three operations: projection, intersection, and shrinking. While hyperbolic~\citep{Ines_2020} embeddings support projection and shrinking operations, the intersection of two hyperbolic embeddings does not yield a hyperbolic space, thus hyperbolic space does not meet the intersection condition. In contrast, Beta and Gamma spaces accommodate all three aforementioned operations. The experimental results of replacing cone with Beta and Gamma on the WikiPeople-3, WikiPeople-4, JF17K-3, and JF17K-4 datasets are shown at the bottom of Table~\ref{table_additional_results}. We also replace the cone operation with an operation similar to the \texttt{TP} module, and the corresponding results are shown in the \texttt{None} row in Table~\ref{table_additional_results}. We observe that the cone space demonstrates superior performance, particularly on the JF17K dataset, where it achieves a 1.8\% and 0.9\% improvement in the MRR metric on the JF17K-3 and JF17K-4 datasets, respectively, compared to Beta embeddings. The primary reason for this lies in the fact that the cone space exhibits greater closure under scaling operations relative to Beta and Gamma spaces. In contrast, the scaling outcomes of Beta and Gamma operations often necessitate simulation through attention mechanisms.}

\smallskip
\noindent{\textbf{Effect of Learning Rate.}}
{A larger learning rate can accelerate the convergence of the model but may risk bypassing the optimal solution during the optimization process. Conversely, a smaller learning rate facilitates more stable convergence to the optimal solution, albeit at a slower pace. Consequently, the selection of an appropriate learning rate entails a trade-off between the speed of convergence and the attainment of optimal performance. \texttt{HyperMono} employs a learning rate of \texttt{6e-4}. We further conduct comparative experiments with learning rates of \texttt{5e-4} and \texttt{7e-4} on the WikiPeople-3, WikiPeople-4, JF17K-3, and JF17K-4 datasets, the corresponding results are shown in the upper section of Table~\ref{table_additional_parameters_results}, which indicate that the choice of learning rate has minimal impact on \texttt{HyperMono}, thereby demonstrating the robustness and stability of the \texttt{HyperMono} model.}

\smallskip
\noindent{\textbf{Effect of Neighbor Numbers.}}
{For the CNA and FNA modules, we aim to enhance the embedding representation of the central entity by leveraging its surrounding neighbors. To investigate the impact of the number of neighboring entities on entity representation, \texttt{HyperMono} sets the number of neighbors to 3. On the WikiPeople-3, WikiPeople-4, JF17K-3, and JF17K-4 datasets, we also conduct comparative analyses using 2 and 4 neighbors, with the results presented at the bottom of Table~\ref{table_additional_parameters_results}. Our findings indicate that selecting 2, 3, or 4 neighbors for a given entity does not significantly affect the final performance of \texttt{HyperMono}. Further analysis of the datasets reveals a pronounced long-tail distribution. For instance, in the WikiPeople-3 dataset, entities with only 1 neighbor account for 45.99\%, while those with 5 or more neighbors constitute 22.00\%. Consequently, the most substantial performance degradation occurs when no neighboring entities are available (as shown in rows \texttt{w/o CNA+TP} and \texttt{w/o FNA+QMP} of Table~\ref{table_ablation_studies}). The comparable performance observed with 2, 3, and 4 neighbors may stem from the fact that a certain number of neighbors already suffice to effectively characterize the entity's knowledge. Given that the number of neighbors is merely a hyperparameter, it is essential to select an appropriate value based on the distribution characteristics of the dataset in practical applications.}

\smallskip
\noindent{\textbf{Effect of Sparse Ratio.}}
{In Section~\ref{entity_neighbor_encoder}, the modules CNA and FNA utilize entity neighbor information to enhance the entity representation. To simulate sparse neighbors, we delete a certain proportion neighbors of entity, specifically we perform neighbor pruning on entities with over five neighbors at ratios of 25\%, 50\%, and 75\%. Comparative experimental results for \texttt{HyperFormer} and \texttt{HyperMono} using Wikipeople-3 and JF17K-3 are shown in Table~\ref{table_removing_neighbors}. We can observe that removing entity samples with a higher number of neighbors significantly impacts the performance of both \texttt{HyperFormer} and \texttt{HyperMono}, with performance fluctuations becoming more pronounced as the removal ratio increases. This phenomenon primarily stems from the reduced number of neighbors diminishing the effectiveness of the CNA and FNA modules, as these modules rely on the neighbors of entities to enhance the representation of the central entity. For \texttt{HyperFormer}, which only incorporates the CNA module, and \texttt{HyperMono}, which includes both CNA and FNA modules, we find that \texttt{HyperMono} consistently outperforms \texttt{HyperFormer} across all removal ratios, underscoring the necessity of the FNA module. Furthermore, in the WikiPeople-3 and JF17K-3 datasets, the proportions of samples with more than five entity neighbors are 22.00\% and 29.06\%, respectively. Combined with the analysis in \textit{effect of neighbor numbers}, it becomes evident that existing HKGC datasets exhibit a pronounced long-tail distribution, where some entities have sparse neighbors while others possess an abundance. We plan to address the performance improvement challenges associated with long-tail distribution in the next phase of our work.}

\begin{table*}[!htp]
\setlength{\abovecaptionskip}{0.05cm}
\renewcommand\arraystretch{1.2}
\setlength{\tabcolsep}{0.515em}
\centering
\small
\caption{Additional ablation study experimental results with different settings.}
\begin{tabular*}{\linewidth}{@{}ccccccccccccccccc@{}}
\hline
\multicolumn{1}{c}{\multirow{2}{*}{\textbf{Settings}}}   & \multicolumn{4}{c}{\textbf{WikiPeople-3}} & \multicolumn{4}{c}{\textbf{WikiPeople-4}} & \multicolumn{4}{c}{\textbf{JF17K-3}} & \multicolumn{4}{c}{\textbf{JF17K-4}}\\
\cline{2-5}\cline{6-9}\cline{10-13}\cline{14-17}

& \textbf{MRR} & \textbf{H@1}   & \textbf{H@3} & \textbf{H@10} & \textbf{MRR} & \textbf{H@1} & \textbf{H@3} & \textbf{H@10} & \textbf{MRR}  & \textbf{H@1} & \textbf{H@3}   & \textbf{H@10} & \textbf{MRR} & \textbf{H@1} & \textbf{H@3}    & \textbf{H@10}\\
\hline
\texttt{Mean}    &0.579  &0.529   &0.598  &0.675  &0.391  &0.356  &0.404  &0.462  &0.860  &0.831  &0.874    &0.919  &0.871  &0.846    &0.894  &0.941 \\
\texttt{Max}    &0.580  &0.529   &0.607  &0.681  &0.394  &0.354  &0.405  &0.470  &0.863  &0.834  &0.878    &0.923  &0.878  &0.846    &0.900  &0.945 \\
\texttt{Sum}    &0.578  &0.531   &0.600  &0.670  &0.392  &0.356  &0.398  &0.464  &0.863  &0.834  &0.875    &0.921  &0.869  &0.839    &0.891  &0.941 \\
\hline
\hline
\texttt{Max}    &0.581  &0.528   &0.606  &0.688  &0.396  &0.356  &0.406  &0.470  &0.864  &0.837  &0.876    &0.919  &0.875  &0.849    &0.898  &0.943 \\
\texttt{Sum}    &0.583  &0.529   &0.603  &0.685  &0.395  &0.356  &0.402  &0.469  &0.866  &0.836  &0.881    &0.925  &0.872  &0.843    &0.898  &0.942 \\
\texttt{Attention}    &0.579  &0.526   &0.603  &0.680  &0.395  &0.354  &0.405  &0.469  &0.867  &0.838  &0.878    &0.928  &0.880  &0.849    &0.900  &0.944 \\
\hline
\hline
\texttt{None}    &0.569  &0.518   &0.584  &0.671  &0.382  &0.346  &0.392  &0.458  &0.847  &0.816  &0.852    &0.900  &0.868  &0.823    &0.869  &0.918 \\
\texttt{Beta}    &0.576  &0.526   &0.603  &0.686  &0.392  &0.350  &0.401  &0.465  &0.849  &0.817  &0.860    &0.905  &0.872  &0.838    &0.892  &0.932 \\
\texttt{Gamma}    &0.578  &0.524   &0.601  &0.683  &0.390  &0.351  &0.398  &0.463  &0.855  &0.825  &0.868    &0.916  &0.870  &0.837    &0.884  &0.931 \\
\hline
\zhiweihu{\texttt{HyperMono}}   &0.586  &0.531  &0.611  &0.690  &0.398 &0.359  &0.408 &0.474  &0.867  &0.839   &0.880    &0.927  &0.881  &0.847  &0.901   &0.945 \\
\hline
\end{tabular*}
\label{table_additional_results}
\end{table*}

\begin{table*}[!htp]
\setlength{\abovecaptionskip}{0.05cm}
\renewcommand\arraystretch{1.2}
\setlength{\tabcolsep}{0.515em}
\centering
\small
\caption{Additional parameter sensitivity experimental results with different settings.}
\begin{tabular*}{\linewidth}{@{}ccccccccccccccccc@{}}
\hline
\multicolumn{1}{c}{\multirow{2}{*}{\textbf{Settings}}}   & \multicolumn{4}{c}{\textbf{WikiPeople-3}} & \multicolumn{4}{c}{\textbf{WikiPeople-4}} & \multicolumn{4}{c}{\textbf{JF17K-3}} & \multicolumn{4}{c}{\textbf{JF17K-4}}\\
\cline{2-5}\cline{6-9}\cline{10-13}\cline{14-17}

& \textbf{MRR} & \textbf{H@1}   & \textbf{H@3} & \textbf{H@10} & \textbf{MRR} & \textbf{H@1} & \textbf{H@3} & \textbf{H@10} & \textbf{MRR}  & \textbf{H@1} & \textbf{H@3}   & \textbf{H@10} & \textbf{MRR} & \textbf{H@1} & \textbf{H@3}    & \textbf{H@10}\\
\hline
\texttt{5e-4}    &0.585  &0.531   &0.608  &0.688  &0.400  &0.359  &0.410  &0.480  &0.867  &0.840  &0.883    &0.928  &0.884  &0.851    &0.903  &0.951 \\
\texttt{7e-4}    &0.582  &0.530   &0.605  &0.681  &0.398  &0.359  &0.409  &0.473  &0.868  &0.841  &0.880    &0.926  &0.883  &0.851    &0.902  &0.946 \\
\hline
\hline
\texttt{num=2}    &0.582  &0.531   &0.606  &0.683  &0.397  &0.357  &0.407  &0.473  &0.867  &0.838  &0.882    &0.925  &0.881  &0.849    &0.902  &0.947 \\
\texttt{num=4}    &0.583  &0.531   &0.606  &0.684  &0.398  &0.358  &0.405  &0.471  &0.865  &0.837  &0.879    &0.928  &0.880  &0.846    &0.898  &0.946 \\
\hline
\zhiweihu{\texttt{HyperMono}}   &0.586  &0.531  &0.611  &0.690  &0.398 &0.359  &0.408 &0.474  &0.867  &0.839   &0.880    &0.927  &0.881  &0.847  &0.901   &0.945 \\
\hline
\end{tabular*}
\label{table_additional_parameters_results}
\end{table*}

\begin{table*}[!htp]
\setlength{\abovecaptionskip}{0.05cm}
\renewcommand\arraystretch{1.05}
\setlength{\tabcolsep}{0.275em}
\centering
\small
\caption{Experimental results after removing entity neighbors at different proportions.}
\begin{tabular*}{0.99\linewidth}{@{}ccccccccccccccccccc@{}}
\hline
\multicolumn{1}{c}{\multirow{3}{*}{\textbf{Settings}}} & \multicolumn{9}{c}{\textbf{WikiPeople-3}} & \multicolumn{9}{c}{\textbf{JF17K-3}}\\
\cline{2-10}\cline{11-19}
& \multicolumn{3}{c}{\textbf{25\%}} & \multicolumn{3}{c}{\textbf{50\%}} & \multicolumn{3}{c}{\textbf{75\%}} & \multicolumn{3}{c}{\textbf{25\%}} & \multicolumn{3}{c}{\textbf{50\%}} & \multicolumn{3}{c}{\textbf{75\%}} \\
\cline{2-19}
&\textbf{MRR} & \textbf{H@1} & \textbf{H@10} &\textbf{MRR} & \textbf{H@1} & \textbf{H@10} &\textbf{MRR} & \textbf{H@1} & \textbf{H@10} &\textbf{MRR} & \textbf{H@1} & \textbf{H@10} &\textbf{MRR} & \textbf{H@1} & \textbf{H@10} &\textbf{MRR} & \textbf{H@1} & \textbf{H@10}  \\
\hline
\texttt{HyperFormer}    &0.417  &0.346   &0.543  &0.365  &0.297  &0.493  &0.282  &0.224  &0.387  &0.423  &0.360    &0.543  &0.399  &0.336    &0.516  &0.349 &0.283  &0.467 \\
\hline
\texttt{HyperMono}    &0.414  &0.344   &0.547  &0.380  &0.305  &0.500  &0.304  &0.247  &0.413  &0.424  &0.365    &0.571  &0.410  &0.341    &0.534  &0.379 &0.311  &0.494 \\
\hline
\end{tabular*}
\label{table_removing_neighbors}
\end{table*}

\subsection*{G\,\,\,Model Transferability}
\hypertarget{model_transferability}{}
{To address \textbf{RQ5}, we further validate the transfer ability of our model. To this aim, we also conduct  experiments on the \textit{knowledge hypergraph link prediction} task. We start by introducing the task definition, dataset descriptions, and baselines. It should be emphasized that our objective here is solely to evaluate the transferability of \texttt{HyperMono}, rather than to conduct an in-depth investigation into knowledge hypergraph link prediction task, therefore, we only introduce necessary background, for a detailed discussion, interested readers are directed to the relevant literature~\cite{Shimin_2021, Chenxu_2023, Shimin_2023, Zhao_2024_1, Shiyao_2022, Zhao_2024}.}

\smallskip
\noindent\zwh{\textbf{Task Definition.}} 
{A \textit{knowledge hypergraph  graph} $\mathcal G$ is defined as $\{\mathcal{E}, \mathcal R, \mathcal F\}$, where $\mathcal E$ is a set of entities, $\mathcal R$ is a set of relation types and $\mathcal F$ is a set of n-ary facts of the form $r(e_1, e_2, \ldots, e_n)$ with $r \in \mathcal R$, $e_i \in \mathcal E,$ for all $ i \in [1,n]$ and $n$ is the non-negative arity of $r$, representing the number of entities involved within each relation. The \textit{knowledge hypergraph link prediction} task looks at predicting a missing entity in the $i$-th position of the tuple $r(e_1,e_2, \ldots, ?, \ldots e_{n-1}, e_n)$. Consistent with PolygonE~\cite{Shiyao_2022}, an n-ary fact $\mathcal{F}$ can essentially be represented in the form of triples + pairs, \textit{i.e.,} $\mathcal{F}=r(e_1, e_2, \ldots, e_n)\rightarrow\mathcal{F}'=(e_1, r, e_2, r_1, e_3, \ldots, r_{n-2}, e_n)\rightarrow\mathcal{F}''=\{(e_1, r, e_2),\{(r_1: e_3), \ldots,(r_{n-2}: e_n)\}\}$. In case $n=2$, $\mathcal{F}''$ is a binary fact, if $n>2$, $\mathcal{F}''$ is a beyond-binary fact, then $(e_1, r, e_2)$ is taken as primary triple. Regarding the methodology for transforming $\mathcal{F}$ into $\mathcal{F}''$, we shall herein provide a concrete example to elucidate the process. Given $\mathcal{F}$=\textit{member\_of\_team}(\textit{James Harden}, \textit{Philadelphia 76ers}, \textit{P.J. Tucker}), by assigning numerical identifiers to the relation \textit{member\_of\_team}, can effectively transform $\mathcal{F}$ into $\mathcal{F}''$=\{(\textit{James Harden}, \textit{member\_of\_team}, \textit{Philadelphia 76ers}), \{(\textit{member\_of\_team\_1},  \textit{P.J. Tucker})\}\}. More details about this data transformation and the rationale behind it can be found in the  PolygonE~\cite{Shiyao_2022} work. Consequently, knowledge hypergraph link prediction task will be transformed into predicting the head and tail entities, $e_1$ and $e_2$, along with the other entities $\{e_3, \ldots, e_n\}$, which is equivalent to all entities prediction within HKGC task.}

\smallskip
\noindent {\textbf{Datasets.}} 
{We test the performance of \texttt{HyperMono} on the knowledge hypergraph link prediction task and conduct experiments on hypergraphs with  \textit{mixed arity}: N-WikiPeople, N-JF17K, and N-FB-AUTO and  with \textit{fixed arity}: N-WikiPeople-3, N-WikiPeople-4, N-JF17K-3, and N-JF17K-4. The statistics  of the corresponding datasets are  consistent with existing literature~\cite{Shimin_2021, Chenxu_2023, Shimin_2023, Zhao_2024_1, Shiyao_2022, Zhao_2024}. In order to differentiate from the datasets for the HKGC task, we have prefixed the dataset for knowledge hypergraph link prediction task with an "N-" designation.}

\smallskip
\noindent\zwh{\textbf{Baselines.}} 
{For knowledge hypergraph link prediction task, we consider 14 state-of-the-art baselines that are universally acknowledged, including RAE~\cite{Richong_2018}, NaLP~\cite{Saiping_2019}, HINGE~\cite{Paolo_2020}, NeuInfer~\cite{Saiping_2020}, HypE~\cite{Bahare_2020}, RAM~\cite{Yu_2021}, HyperMLN~\cite{Zirui_2022}, tNaLP+~\cite{Saiping_2023}, S2S~\cite{Shimin_2021}, HyConvE~\cite{Chenxu_2023}, MSeaKG~\cite{Shimin_2023}, HJE~\cite{Zhao_2024_1}, PolygonE~\cite{Shiyao_2022}, and HyCubE~\cite{Zhao_2024}.}

\begin{table*}[!htp]
\setlength{\abovecaptionskip}{0.05cm}
\renewcommand\arraystretch{1.1}
\setlength{\tabcolsep}{0.65em}
\centering
\small
\caption{\vic{Evaluation of the \emph{knowledge hypergraph completion task} in the mixed arity setting   on the N-WikiPeople, N-JF17K and N-FB-AUTO datasets. Best scores are highlighted in \colorbox{mycolor2}{\textbf{bold}}, the second best scores are highlighted in \colorbox{mycolor1}{normal}.}}
\begin{tabular*}{0.85\linewidth}{@{}ccccccccccccc@{}}
\hline
\multicolumn{1}{c}{\multirow{2}{*}{\textbf{Methods}}}   & \multicolumn{4}{c}{\textbf{N-WikiPeople}} & \multicolumn{4}{c}{\textbf{N-JF17K}} & \multicolumn{4}{c}{\textbf{N-FB-AUTO}}\\
\cline{2-5}\cline{6-9}\cline{10-13}

& \textbf{MRR} & \textbf{H@1}   & \textbf{H@3} & \textbf{H@10} & \textbf{MRR} & \textbf{H@1} & \textbf{H@3} & \textbf{H@10} & \textbf{MRR}  & \textbf{H@1} & \textbf{H@3}   & \textbf{H@10}\\
\hline
RAE~\cite{Richong_2018}    &0.253  &0.118   &0.343  &0.463  &0.396  &0.312  &0.433  &0.561  &0.703  &0.614  &0.764    &0.854  \\
NaLP~\cite{Saiping_2019}    &0.338  &0.272   &0.364  &0.466  &0.310  &0.239  &0.334  &0.450  &0.672  &0.611  &0.712    &0.774  \\
HINGE~\cite{Paolo_2020}    &0.333  &0.259   &0.361  &0.477  &0.473  &0.397  &0.490  &0.618  &0.678  &0.630  &0.706    &0.765 \\
NeuInfer~\cite{Saiping_2020}    &0.350  &0.282   &0.381  &0.467  &0.451  &0.373  &0.484  &0.604  &0.737  &0.700  &0.755    &0.805 \\
HypE~\cite{Bahare_2020}    &0.263  &0.127   &0.355  &0.486  &0.494  &0.399  &0.532  &0.650  &0.804  &0.774  &0.824    &0.856 \\
RAM~\cite{Yu_2021}    &0.380  &0.279   &0.455  &0.539  &0.539  &0.463  &0.573  &0.690  &0.830  &0.803  &0.851    &0.876 \\
HyperMLN~\cite{Zirui_2022}    &0.351  &0.270   &0.394  &0.497  &0.556  &0.482  &0.597  &0.717  &0.831  &0.803  &0.851    &0.877 \\
tNaLP+~\cite{Saiping_2023}    &0.339  &0.269   &0.369  &0.473  &0.449  &0.370  &0.484  &0.598  &0.729  &0.645  &0.748    &0.826 \\
S2S~\cite{Shimin_2021}    &0.372  &0.277   &0.439  &0.533  &0.528  &0.457  &0.570  &0.690  &-  &-  &-    &- \\
HyConvE~\cite{Chenxu_2023}    &0.362  &0.275   &0.388  &0.501  &\cellcolor{mycolor1}0.590  &0.478  &0.610  &0.729  &0.847  &0.820  &0.872    &0.901 \\
MSeaKG~\cite{Shimin_2023}    &0.392  &0.290   &0.468  &0.553  &0.561  &0.475  &0.591  &0.705  &-  &-  &-    &- \\
HJE~\cite{Zhao_2024_1}    &\cellcolor{mycolor1}0.450  &\cellcolor{mycolor1}0.375   &\cellcolor{mycolor1}0.487  &\cellcolor{mycolor1}0.582  &\cellcolor{mycolor1}0.590  &\cellcolor{mycolor1}0.507  &0.613  &0.729  &\cellcolor{mycolor1}0.872  &0.848  &0.886    &0.903 \\
PolygonE~\cite{Shiyao_2022}    &0.431  &0.334   &0.454  &0.568  &0.565  &0.485  &0.602  &0.708  &0.858  &0.826  &0.871    &\cellcolor{mycolor1}0.921 \\
HyCubE~\cite{Zhao_2024}    &0.438  &0.358   &0.482  &0.575  &0.581  &\cellcolor{mycolor1}0.507  &\cellcolor{mycolor1}0.615  &\cellcolor{mycolor1}0.731  &\cellcolor{mycolor2}\textbf{0.884}  &\cellcolor{mycolor2}\textbf{0.863}  &\cellcolor{mycolor1}0.897    &0.920 \\
\hline
\zhiweihu{\texttt{HyperMono}}   &\cellcolor{mycolor2}\textbf{0.481}  &\cellcolor{mycolor2}\textbf{0.381}  &\cellcolor{mycolor2}\textbf{0.545}  &\cellcolor{mycolor2}\textbf{0.637}  &\cellcolor{mycolor2}\textbf{0.676} &\cellcolor{mycolor2}\textbf{0.612}  &\cellcolor{mycolor2}\textbf{0.708} &\cellcolor{mycolor2}\textbf{0.801}  &\cellcolor{mycolor2}\textbf{0.884}  &\cellcolor{mycolor1}0.855   &\cellcolor{mycolor2}\textbf{0.907}    &\cellcolor{mycolor2}\textbf{0.937} \\
\hline
\end{tabular*}
\label{table_mixed_arity}
\end{table*}

\begin{table*}[!htp]
\setlength{\abovecaptionskip}{0.05cm}
\renewcommand\arraystretch{1.1}
\setlength{\tabcolsep}{0.485em}
\centering
\small
\caption{\vic{Evaluation of the \emph{knowledge hypergraph completion task} in the fixed arity setting  on the N-WikiPeople-3, N-WikiPeople-4, N-JF17K-3 and N-JF17K-4 datasets. Best scores are highlighted in \colorbox{mycolor2}{\textbf{bold}}, the second best scores are highlighted in \colorbox{mycolor1}{normal}.}}
\begin{tabular*}{\linewidth}{@{}ccccccccccccccccc@{}}
\hline
\multicolumn{1}{c}{\multirow{2}{*}{\textbf{Methods}}}   & \multicolumn{4}{c}{\textbf{N-WikiPeople-3}} & \multicolumn{4}{c}{\textbf{N-WikiPeople-4}} & \multicolumn{4}{c}{\textbf{N-JF17K-3}} & \multicolumn{4}{c}{\textbf{N-JF17K-4}}\\
\cline{2-5}\cline{6-9}\cline{10-13}\cline{14-17}

& \textbf{MRR} & \textbf{H@1}   & \textbf{H@3} & \textbf{H@10} & \textbf{MRR} & \textbf{H@1} & \textbf{H@3} & \textbf{H@10} & \textbf{MRR}  & \textbf{H@1} & \textbf{H@3}   & \textbf{H@10} & \textbf{MRR} & \textbf{H@1} & \textbf{H@3}    & \textbf{H@10}\\
\hline
RAE~\cite{Richong_2018}    &0.239  &0.168   &0.252  &0.379  &0.150  &0.080  &0.149  &0.273  &0.505  &0.430  &0.532    &0.644  &0.707  &0.636    &0.751  &0.835 \\
NaLP~\cite{Saiping_2019}    &0.301  &0.226   &0.327  &0.445  &0.342  &0.237  &0.400  &0.540  &0.515  &0.431  &0.552    &0.679  &0.719  &0.673    &0.742  &0.805 \\
HINGE~\cite{Paolo_2020}    &0.338  &0.255   &0.360  &0.508  &0.352  &0.241  &0.419  &0.557  &0.587  &0.509  &0.621    &0.738  &0.745  &0.700    &0.775  &0.842 \\
NeuInfer~\cite{Saiping_2020}    &0.355  &0.262   &0.388  &0.521  &0.361  &0.255  &0.424  &0.566  &0.622  &0.533  &0.658    &0.770  &0.765  &0.722    &0.808  &0.871 \\
HypE~\cite{Bahare_2020}    &0.266  &0.183   &-  &0.443  &0.304  &0.191  &-  &0.527  &0.364  &0.255  &-    &0.573  &0.408  &0.300    &-  &0.627 \\
RAM~\cite{Yu_2021}    &0.254  &0.190   &-  &0.383  &0.226  &0.161  &-  &0.367  &0.578  &0.505  &-    &0.722  &0.743  &0.701    &-  &0.845 \\
HyperMLN~\cite{Zirui_2022}    &0.252  &0.193   &-  &0.385  &0.224  &0.167  &-  &0.370  &0.574  &0.501  &-    &0.711  &0.734  &0.687    &-  &0.831 \\
tNaLP+~\cite{Saiping_2023}    &0.270  &0.185   &-  &0.444  &0.344  &0.223  &-  &0.578  &0.411  &0.325  &-    &0.617  &0.630  &0.531    &-  &0.722 \\
S2S~\cite{Shimin_2021}    &0.386  &0.299   &\cellcolor{mycolor1}0.421  &0.559  &0.391  &0.270  &\cellcolor{mycolor1}0.470  &0.600  &0.740  &\cellcolor{mycolor1}0.676  &\cellcolor{mycolor1}0.770    &0.860  &0.822  &0.761    &\cellcolor{mycolor1}0.853  &0.924 \\
HyConvE~\cite{Chenxu_2023}    &0.318  &0.240   &-  &0.482  &0.386  &0.271  &-  &0.607  &0.729  &0.670  &-    &0.861  &0.827  &0.770    &-  &0.931 \\
MSeaKG~\cite{Shimin_2023}    &\cellcolor{mycolor1}0.403  &-   &-  &0.579  &\cellcolor{mycolor1}0.409  &-  &-  &\cellcolor{mycolor1}0.624  &\cellcolor{mycolor1}0.754  &-  &-    &\cellcolor{mycolor1}0.889  &\cellcolor{mycolor1}0.833  &-    &-  &\cellcolor{mycolor1}0.938 \\
HJE~\cite{Zhao_2024_1}    &0.381  &\cellcolor{mycolor1}0.303   &-  &\cellcolor{mycolor1}0.596  &0.388  &\cellcolor{mycolor1}0.276  &-  &0.607  &0.707  &0.637  &-    &0.851  &0.827  &\cellcolor{mycolor1}0.775    &-  &0.928 \\
HyCubE~\cite{Zhao_2024}    &0.322  &0.247   &-  &0.469  &0.370  &0.252  &-  &0.583  &0.615  &0.544  &-    &0.752  &0.792  &0.744    &-  &0.879 \\
\hline
\zhiweihu{\texttt{HyperMono}}   &\cellcolor{mycolor2}\textbf{0.454}  &\cellcolor{mycolor2}\textbf{0.375}  &\cellcolor{mycolor2}\textbf{0.486}  &\cellcolor{mycolor2}\textbf{0.611}  &\cellcolor{mycolor2}\textbf{0.486} &\cellcolor{mycolor2}\textbf{0.395}  &\cellcolor{mycolor2}\textbf{0.545} &\cellcolor{mycolor2}\textbf{0.650}  &\cellcolor{mycolor2}\textbf{0.793}  &\cellcolor{mycolor2}\textbf{0.739}   &\cellcolor{mycolor2}\textbf{0.824}    &\cellcolor{mycolor2}\textbf{0.901}  &\cellcolor{mycolor2}\textbf{0.888}  &\cellcolor{mycolor2}\textbf{0.855}  &\cellcolor{mycolor2}\textbf{0.910}   &\cellcolor{mycolor2}\textbf{0.951} \\
\hline
\end{tabular*}
\label{table_fixed_arity}
\end{table*}

\smallskip
\noindent\zwh{\textbf{Performance Analysis.}} 
\zwh{The results of the knowledge hypergraph completion experiments in the  mixed arity and fixed arity settings are respectively shown in Table~\ref{table_mixed_arity} and Table~\ref{table_fixed_arity}. We observe that our \texttt{HyperMono} consistently achieves state-of-the-art performance on all benchmark datasets.  For the mixed arity scenario, compared with the best-performing baseline HyCubE, \texttt{HyperMono} obtains improvements of  4.3\%  on  N-WikiPeople and   9.5\%  on N-JF17K in the MRR metric.  On the N-FB-AUTO dataset, \texttt{HyperMono} virtually matches the results of the best performing baseline. These results demonstrate the capability and effectiveness of \texttt{HyperMono} for capturing and extracting various types of information and knowledge. The same phenomenon occurs in the fixed arity scenario,   \texttt{HyperMono} respectively obtains performance improvements of 13.2\%, 11.6\%, 17.8\%, and 9.6\% in the MRR metric on the N-WikiPeople-3, N-WikiPeople-4, N-JF17K-3, and N-JF17K-4 datasets, when compared with the  best baseline HyCubE. Such a notable  performance improvement is mainly due to our explicit  treatment  of qualifier monotonicity during the encoding process. This feature is more suitable for scenarios where arity is fixed, because introducing qualifier monotonicity can greatly reduce the scope of the optional answer space to obtain the final answer.}

\subsection*{H\,\,\,Case Study}
\hypertarget{case_study}{}
{To demonstrate \texttt{HyperMono}'s encoding capability for qualifier monotonicity, we conduct a case study, with the corresponding results present in Table~\ref{table_case_study}. For the main triple (\textit{Leonardo DiCaprio}, \textit{cast member}, \texttt{?}), \texttt{HyperMono}'s predictions for potential entities at the \texttt{?} position including multiple entities, with the probability values for each prediction showing minimal variance. Upon introducing the qualifier pair (\textit{genre}: \textit{drama film}), we can observe that entities previously associated with higher probability values, such as \textit{Titanic} and \textit{Inception} are excluded. Conversely, the prediction probabilities for entities \textit{The Revenant}, \textit{The Wolf of Wall Street}, and \textit{Don’t Look Up} significantly increased. Further addition of the qualifier pair (\textit{director}: \textit{Martin Scorsese}) led to a further narrowing of the answer space. This process exhibits a phenomenon of answer space expansion and contraction as the number of qualifier pairs increased. The cone embedding within \texttt{HyperMono} elegantly accommodates this property through the shrinking operations of the cone.}
\begin{table}[!htp]
\setlength{\abovecaptionskip}{0.05cm}
\renewcommand\arraystretch{1.05}
\setlength{\tabcolsep}{0.25em}
\centering
\small
\caption{Case study on the importance of the qualifier monotonicity. The probability of -- is close to 0, so we do not show it here.}
\begin{tabular*}{\linewidth}{@{}ccc@{}}
\hline
\multicolumn{1}{c}{\textbf{Hyper-relational Fact}} &\multicolumn{1}{c}{\textbf{Answer}} &\multicolumn{1}{c}{\textbf{Probability}}\\
\hline
\multicolumn{1}{c}{\multirow{6}{*}{(\textit{Leonardo DiCaprio}, \textit{cast member}, \texttt{?})}}  &\textit{Titanic}  &8.0\% \\
&\textit{Inception}  &7.5\% \\
&\textit{The Revenant}   &6.8\% \\
&\textit{The Wolf of Wall Street}  &6.2\% \\
&\textit{Catch Me If You Can}    &5.9\%\\
&\textit{Don't Look Up}  &5.3\%\\
\hline
\multicolumn{1}{c}{\multirow{6}{*}{\makecell{(\textit{Leonardo DiCaprio}, \textit{cast member}, \texttt{?})\\\textcolor{myred}{(\textit{genre}: \textit{drama film})}}}}  
&\textcolor{myred}{\sout{\textit{Titanic}}}  &-- \\
&\textcolor{myred}{\sout{\textit{Inception}}}  &-- \\
&\textit{The Revenant}   &25.6\% \\
&\textit{The Wolf of Wall Street}  &23.4\% \\
&\textcolor{myred}{\sout{\textit{Catch Me If You Can}}}    &-- \\
&\textit{Don't Look Up}  &20.3\% \\
\hline
\multicolumn{1}{c}{\multirow{6}{*}{\makecell{(\textit{Leonardo DiCaprio}, \textit{cast member}, \texttt{?})\\\textcolor{myred}{(\textit{genre}: \textit{drama film})}\\\textcolor{mygreen}{(\textit{director}: \textit{Martin Scorsese})}}}}  
&\textcolor{myred}{\sout{\textit{Titanic}}}  &-- \\
&\textcolor{myred}{\sout{\textit{Inception}}}  &-- \\
&\textcolor{mygreen}{\sout{\textit{The Revenant}}}   &-- \\
&\textit{The Wolf of Wall Street}  &75.2\% \\
&\textcolor{myred}{\sout{\textit{Catch Me If You Can}}}    &-- \\
&\textcolor{mygreen}{\sout{\textit{Don't Look Up}}}  &-- \\
\hline
\end{tabular*}
\label{table_case_study}
\end{table}

\subsection*{I\,\,\ Faithful Encoding of Qualifier Monotonicity}
\hypertarget{proof_of_propositions}{}
{\textbf{Proposition 1.} \textit{Given two hyper-relational facts $q_1=(h, r, \texttt{?}, Q_1)$ and $q_2=(h, r, \texttt{?}, Q_2)$ where $Q_1 \subseteq Q_2$, i.e., $q_1$ is a partial fact of $q_2$, the output of the score function $f(\cdot)$ of the Shrink operation (denoted here as $S(\cdot)$ for simplicity) defined in Equation~\ref{equation_3} adheres to the constraint $f(S(q_1)) \geq f(S(q_2))$.}}

\noindent {\textit{Proof.} We begin by demonstrating that the resulting cone of $q_1$ subsumes that of $q_2$. Since both $q_1$ and $q_2$ contain the same primary triple $(h, r, \texttt{?})$ and satisfy $Q_1 \subseteq Q_2$, it follows that the final shrunken cone of $q_2$ must be a subset of the shrunken cone of $q_1$. That is, we obtain $S({q_1}) \supseteq S({q_2})$. Given the tail entity \texttt{?}, whose embedding is represented as \textbf{E} (which effectively consists only of the symmetry axis with an aperture size of 0), we primarily consider three possible cases regarding its position.}
{
\begin{enumerate}[itemsep=0.5ex, leftmargin=5mm]
\item \textbf{E} is located inside the smaller cone $S({q_2})$. Since $S({q_1}) \supseteq S({q_2})$, it follows that \textbf{E} must also be inside $S({q_1})$. The distance function $\mathcal{D}(\cdot)$ for measuring the point-to-cone distance, as proposed in \textbf{Section 4.3} of ConE~\citep{Zhanqiu_2021}, is monotonically increasing \textit{w.r.t.} the distance from the tail point to the symmetry axis of the cone. Consequently, we obtain $\mathcal{D}($\textbf{E}, $S({q_1})$) $\geq$ $\mathcal{D}($\textbf{E}, $S({q_2})$), which implies that $f(S(q_1)) \geq f(S(q_2))$.
\item \textbf{E} is located outside the smaller cone $S({q_2})$ but inside the larger cone $S({q_1})$. According to the definition of the point-to-cone distance $\mathcal{D}(\cdot)$, it follows that $\mathcal{D}($\textbf{E}, $S({q_1})$) $\geq$ $\mathcal{D}($\textbf{E}, $S({q_2})$), which implies $f(S(q_1)) \geq f(S(q_2))$.
\item \textbf{E} is located outside the larger cone $S({q_1})$. Since $S({q_1}) \supseteq S({q_2})$, this implies that \textbf{E} is also outside the smaller cone $S({q_2})$. Note that the point-to-cone distance $\mathcal{D}(\cdot)$ is monotonically decreasing \textit{w.r.t.} the increase in the sector-cone area. Therefore, we obtain $\mathcal{D}($\textbf{E}, $S({q_1}$)) $\leq$ $\mathcal{D}($\textbf{E}, $S({q_2}$)), which implies $f(S(q_1)) \geq f(S(q_2))$.
\end{enumerate}
}